\documentclass[table]{article}
\usepackage[final]{graphicx}
\usepackage{template/frExamplee}
\usepackage{apalike}
\usepackage{setspace}
\usepackage{enumitem}
\usepackage[T1]{fontenc}     
\usepackage[utf8]{inputenc}  
\usepackage[english]{babel}  
\usepackage{url}             
\usepackage{upgreek}         
\usepackage{amsmath}         
\usepackage{amsfonts}
\usepackage{mathtools}
\usepackage{geometry}
\usepackage{listings}
\usepackage{tikz}
\usepackage{placeins}
\usetikzlibrary{decorations.text,shapes,arrows,positioning,calc,arrows.meta,patterns}
\usepackage{pgfplots} 
\pgfplotsset{compat=1.18} 
\usepackage{hyperref}
\addto\extrasenglish{
}
\usepackage{bm} 
\usepackage{caption}
\usepackage{cleveref}
\usepackage{subcaption}
\usepackage{csvsimple}
\usepackage{pgf-pie}
\usepackage{tabularx}
\usepackage{multirow} 

\usepackage{soul}
\definecolor{MyOrange}{rgb}{0.92,0.57,0.2}

\definecolor{DarkGreen}{rgb}{0.0,0.5,0.0}

\usepackage{fancyhdr}
\newcommand{\figref}[1]{Figure\;\ref{#1}}

\usepackage{comment}

\title{er.autopilot 1.0: The Full Autonomous Stack for Oval Racing at High Speeds}

\author{
Ayoub Raji\thanks{Correspondence: ayoub.raji@unimore.it} \thanks{Authors are with University of Modena and Reggio Emilia, Italy} \thanks{Authors are with University of Parma, Italy},
\And
Danilo Caporale\thanks{Authors are with Technology Innovation Institute - Autonomous Robotics Research Center, UAE} ,
\And
Francesco Gatti\thanks{Authors are with HIPERT srl} ,
\And
Andrea Giove\thanks{Authors are with University of Pisa, Italy} ,
\And
Micaela Verucchi\footnotemark[25] ,
\And
Davide Malatesta\footnotemark[4] ,
\And
Nicola Musiu\footnotemark[2] ,
\And
Alessandro Toschi\footnotemark[2] ,
\And
Silviu Roberto Popitanu\footnotemark[6] ,
\And
Fabio Bagni\footnotemark[25] ,
\And
Massimiliano Bosi\footnotemark[25] ,
\And
Alexander Liniger\thanks{Authors are with Computer Vision Lab, ETH Zurich, Switzerland} ,
\And
Marko Bertogna\footnotemark[25] ,
\And
Daniele Morra\footnotemark[6] ,
\And
Francesco Amerotti\footnotemark[5] ,
\And
Luca Bartoli\footnotemark[5] ,
\And
Federico Martello\footnotemark[2] ,
\And
Riccardo Porta\footnotemark[2]
}

\begin{document}

\maketitle
\begin{abstract}
The Indy Autonomous Challenge (IAC) brought together for the first time in history nine autonomous racing teams competing at unprecedented speed and in head-to-head scenario, using independently developed software on open-wheel racecars.
This paper presents the complete software architecture used by team TII EuroRacing (TII-ER), covering all the modules needed to avoid static obstacles, perform active overtakes and reach speeds above 75 m/s (270 km/h). In addition to the most common modules related to perception, planning, and control, we discuss the approaches used for vehicle dynamics modelling, simulation, telemetry, and safety. Overall results and the performance of each module are described, as well as the lessons learned during the first two events of the competition on oval tracks, where the team placed respectively second and third.
\end{abstract}
\newpage
\section{Introduction}
\label{sec:introduction}
The introduction of Advanced Driver Assistance Systems (ADAS) and partially automated systems on commercial cars has reduced the number of motor vehicle crashes and deaths in the majority of high-income countries \cite{yellman2022motor}. This trend could become even more effective in the next decades thanks to speed limit regulations and the obligation for car manufacturers to include advanced safety systems, such as Driver Alcohol Detection System for Safety and Driver Drowsiness Detection System, on all their vehicles \cite{ecola}\cite{2018PotentialRI}. Nevertheless, an important number of nowadays crashes and deaths are caused by harsh weather conditions, poor visibility, and loss of control, which are unlikely preventable by the current ADAS \cite{2018PotentialRI}. This should push the research on accelerating fully autonomous driving on highways, high speed scenarios, and harsh road conditions.

Motorsport has always produced innovative technologies which, in many cases, have been transferred later on to road cars improving safety and enhancing performance. Examples are rear-view mirrors, seat belts, active suspensions, and engine recovery systems. Similarly to the integration of Motorsport technological innovations on the human-driven urban cars, Autonomous Racing could help in developing and testing self-driving capabilities in extreme cases on race tracks to be ported on future urban autonomous vehicles \cite{betz1}.

Autonomous driving competitions have historically been very effective in fostering research and industrial interest to push self-driving technology beyond its limits. A first milestone was set in 2005 with the DARPA Grand Challenge, where multiple teams competed to autonomously drive off-road vehicles along a 132 miles path in the desert near the California/Nevada state line. The Stanford Racing Team won the 2 million prize, completing the path in slightly less than 7 hours. Researchers from the five teams that completed the challenge have been involved as founders and chief researchers of companies that only a decade later would render urban autonomous vehicles a commercial reality. 

In the autonomous racing domain, two notable initiatives have been proposed in 2016. The f1tenth\footnote{\href{https://f1tenth.org/}{https://f1tenth.org/}} initiative is an open source platform for the development and testing of autonomous driving software, consisting of 1:10 scale RC cars equipped with a LiDAR scanner, stereo camera, and Nvidia computational boards. Annual international race events of the f1tenth are organized during the most important conferences in Robotics. Instead, Roborace\footnote{\href{https://roborace.com/}{https://roborace.com/}} provides full scale electric racecars able to achieve speeds around $69.4$ m/s ($250$ km/h). The competition is based on a championship formed by several real and virtual races. In 2017, the Formula SAE\footnote{\href{https://www.fsaeonline.com/}{https://www.fsaeonline.com/}} created the new Formula Student Driverless (FSD) class where teams formed by students have to design and develop both the mechanics and software of a prototype capable of autonomously running in a closed loop track created by cones. In 2020, a new competition, the Indy Autonomous Challenge (IAC), was launched aiming to showcase multi-vehicle head to head races at the limits of handling in high speed racetracks. 

In this paper, we present \texttt{er.autopilot 1.0}, the complete software stack used during the IAC by the TII EuroRacing team, which accomplished the second and third position in the first two events. The system demonstrated to be able to avoid static obstacles, perform active overtakes on other vehicles, and achieve speeds above $75$ m/s. 
The aim of this work is not only to be a reference for the Autonomous Racing domain but also for other autonomous systems in edge-case scenarios for road vehicles and sport-cars. Among the major contributions, we present a vehicle model identification approach, where a model-based controller is tuned using simulation tools without prior dynamic data of the vehicle. We exhaustively characterize the performance of each software and control module, and of the overall system, deriving the main lessons learned by analyzing the pros and cons of each solution.

In the remainder of \autoref{sec:introduction}, we give a brief introduction to the competition and the racecar. Related research in the Autonomous Racing domain is discussed in \autoref{sec:related}. The full stack of \texttt{er.autopilot 1.0}, the underlying design principles, and the technological solutions adopted are presented in \autoref{sec:sw_stack}. In particular, the modules related to localization and perception are presented, including a LiDAR-based solution, and the description of different clustering and detection approaches.
The software modules related to motion forecasting, planning, and control have already been presented in \cite{raji}. We thus give here a brief description of their implementation and focus on their effects on the overall system and the final results obtained. \autoref{sec:simulation} presents the simulation platforms used for testing. Telemetry and Visualization tools are illustrated in \autoref{sec:telemetry}. The results of each module and of the overall system during the competition are summarized in \autoref{sec:results}. \autoref{sec:lessons} gathers an overview of the lessons learned by the team. The paper is concluded in \autoref{sec:conclusion}, where potential improvements and future research directions are also presented.

\subsection{Indy Autonomous Challenge}

The IAC\footnote{\href{https://www.indyautonomouschallenge.com/}{https://www.indyautonomouschallenge.com/}} is an international competition that brings together public-private partnerships and academic institutions to challenge university students around the world to imagine, invent and prove a new generation of automated vehicle software to run fully autonomous race cars.

The challenge was carried out in two steps: a simulation race, and a real race. Of the 30 teams from universities all around the world that participated in this competition, only 9 passed the simulation step.
The first race, the Indy Autonomous Challenge powered by Cisco, has been held on October 21st 2021 at the  Indianapolis Motor Speedway (IMS), and the second one, the Autonomous Challenge @ CES, on January 7th 2022 at the Las Vegas Motor Speedway (LVMS). The car shakedown and a considerable part of the development before the IMS race was conducted at Lucas Oil Raceway (LOR).

The race at IMS was a solo time trial competition, that consisted of a semi-final and final event.
In order to get access to the race, the teams had to demonstrate a set of requirements during testing.
This event, besides the time trial, included an obstacle avoidance challenge: two static obstacles were placed in the front stretch, to prove that the car was capable of actively avoiding static obstacles. The final leg was limited to the three teams that achieved the best score in the semi-finals. The entire run was formed of four warm-up laps and two performance laps. The winner was determined based on the highest average speed achieved during the two consecutive performance laps.

The race at LVMS consisted in a Passing Competition, where multiple rounds of head-to-head matches were conducted by two cars that had to take turns playing the role of defender and attacker, attempting to overtake at increasing speeds, until one or both cars were unable to complete a pass. In each round, the attacker had to follow the following four steps:

\begin{enumerate}[noitemsep]
\item Reduce the gap with the defender.
\item Keep a longitudinal safety distance.
\item Overtake the defender once reached a passing zone.
\item Switch the role to defender and reduce the velocity to a pre-determined constant value.
\end{enumerate}

A time trial event was created to determine the teams' seeding in the brackets of the Passing Competition.

It is worth noting that the original plan for the IAC was to have a multi-vehicle race with 10 cars on track already at IMS. Due to several challenges (from weather to logistics to teams facing new difficulties on track over a short period of time), the race rules were modified to deliver a successful show where most teams could participate with their current level of readiness.

\subsection{Dallara AV-21}
\label{sec:dallaraav21}

\begin{figure}[htb!]
  \centering
  \begin{subfigure}{0.49\textwidth}
    \includegraphics[
        width=0.99\textwidth,
    ]{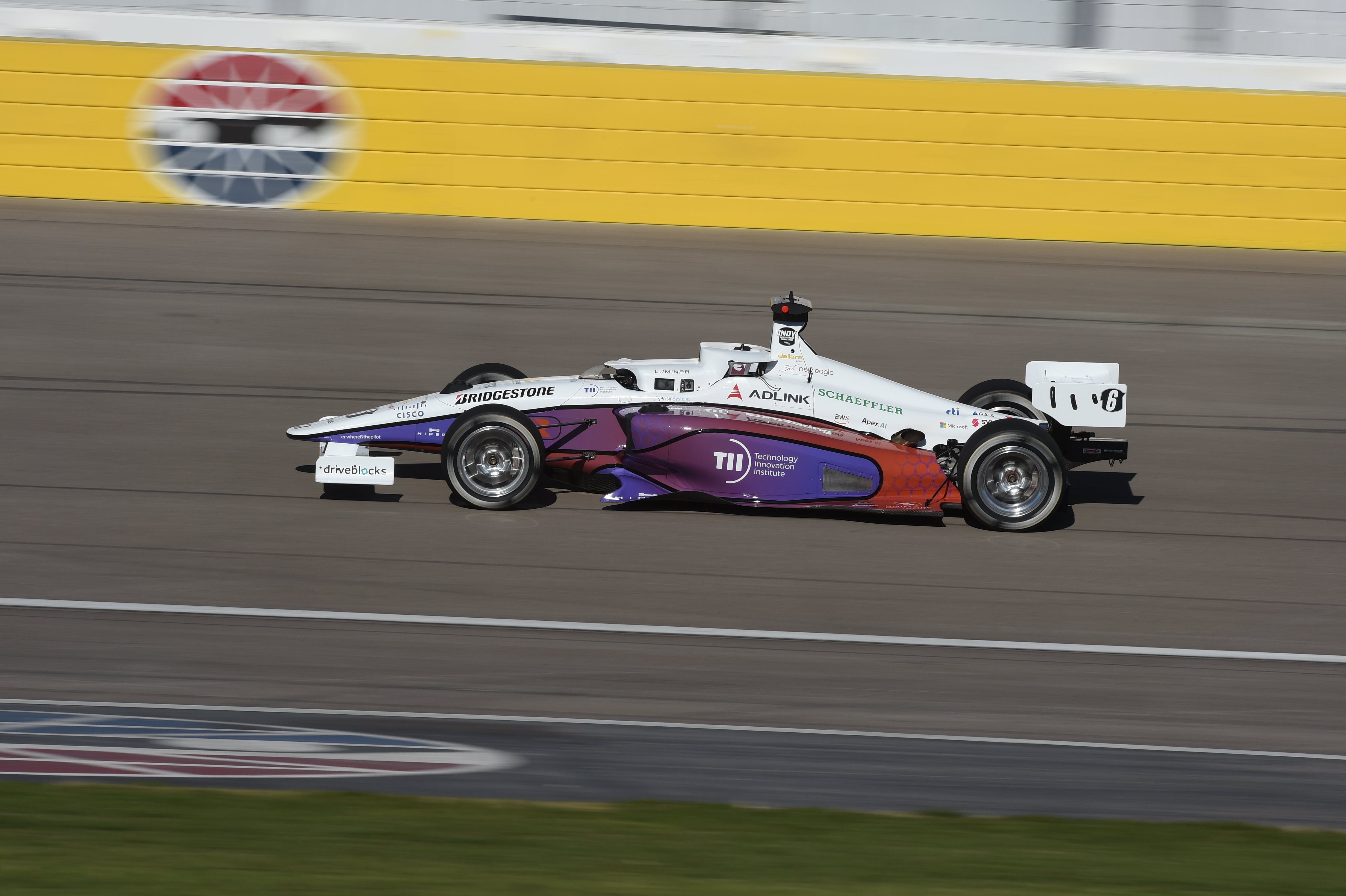}
    \caption{AV-21 with the TII EuroRacing livery.}
  \end{subfigure}
  \begin{subfigure}{0.49\textwidth}
    \includegraphics[
        width=0.99\textwidth,
    ]{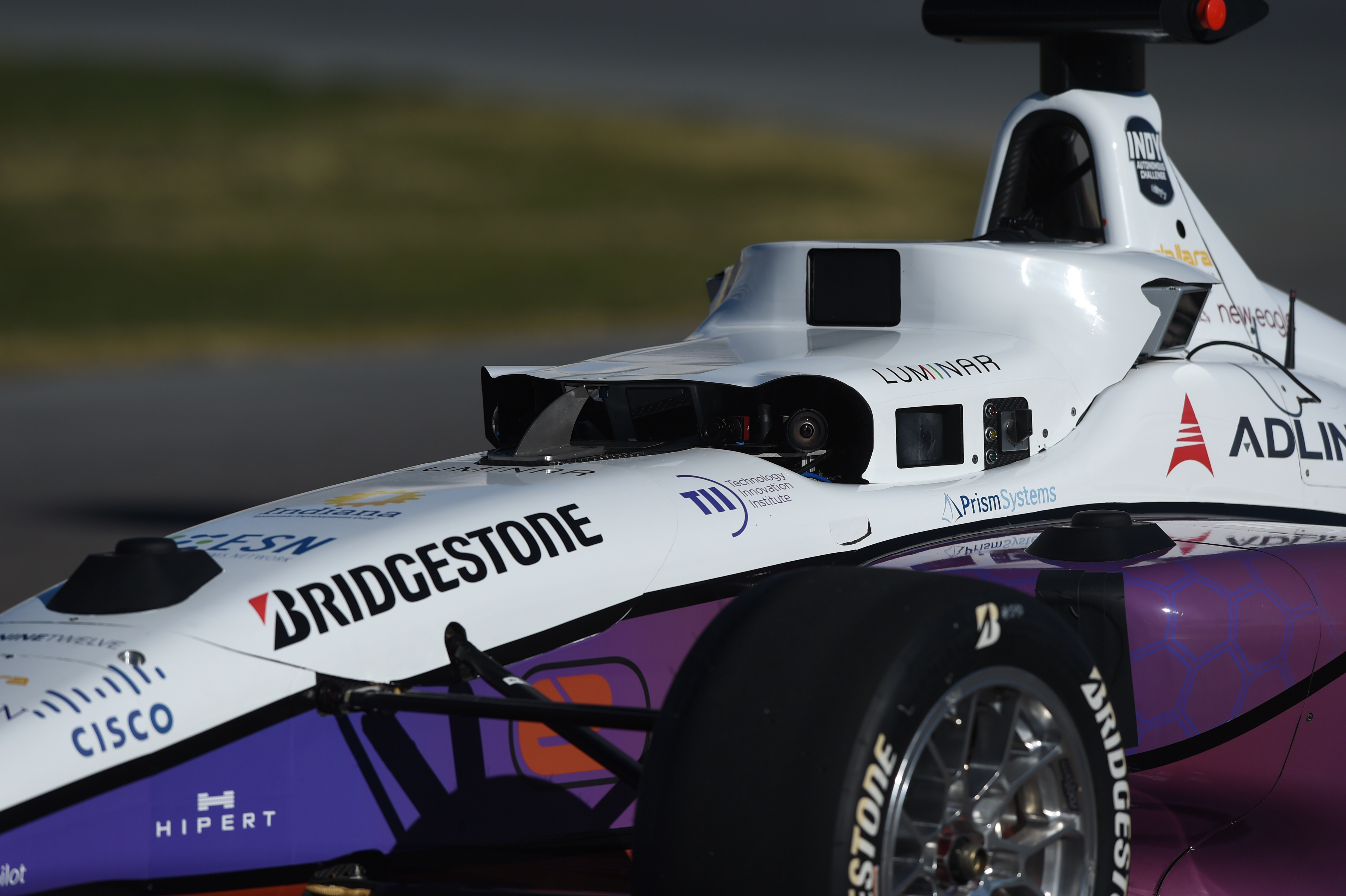}
    \caption{Close view on the onboard sensors.}
  \end{subfigure}
  \caption{Dallara AV-21}
  \label{fig:AV21}
\end{figure}

Each team of the IAC participates with a Dallara AV-21, shown in \autoref{fig:AV21}, a fully autonomous open-wheel race car based on the official Indy Lights IL-15. Unlike the original race car, the engine mounted is a turbo-charged Honda K20 with $390$ horsepower. The mechanics, suspensions, and aerodynamics are adjusted for oval racing and high banked tracks with an asymmetrical setup.

On the perception side, the car is equipped with two GNSS modules, three LiDARs, six cameras (two cameras in a stereo setup and four to cover the 360 range), and three RADARs.
The Novatel GNSS Pwrpak 7d receivers provide a centimetre precision localization of the car thanks to four antennas and RTK correction. 
The three solid-state LUMINAR H3 LiDARs have a range of 200m and they operate at 20 Hz.
The Aptiv ESR 2.5 frontal radar and the MRR side radars have a range of around 160m and they provide the detected obstacles at 10 Hz.
The six high-resolution RGB Mako G319C cameras from Allied Vision are mounted to have a view of almost 360 degrees around the car.
The computing platform used on the vehicle is an ADLINK AVA-3501 consisting of an 8 core Intel Xeon E 2278 GE CPU, an NVIDIA RTX Quadro 8000 GPU, and 64 GB DDR4 RAM. For external communication, Cisco FM-4500 radio transceivers are used on the car, and around the tracks in order to make  telemetry data available to the crew team in the pitlane. Race Control signals can be exchanged by means of a MYLAPS RaceLink system mounted on the racetracks and a transponder mounted on the vehicle.

On the actuation side, the car has a Drive-by-Wire (DBW) system realized by Schaeffler to actuate the steering, the throttle pedal, the brake pedal, and the gearbox. The New Eagle GCM 196 Raptor control module is used as an interface between the DBW system, the computing platform in which the algorithms are executed, and the other units related to the engine and minor subsystems.
\section{Related Work}
\label{sec:related}

Thanks to the availability of low-cost research racecar prototypes and the mediatic impact of autonomous driving challenges, the number of published works in this domain is progressively increasing. In \cite{betz2}, the authors presented a survey on autonomous racing cars reviewing the most relevant publications detailing autonomous driving modules, vehicle modelling, simulation, and complete software architectures.

One of the first problems each team faces when working on an autonomous (race)car is that of obtaining an accurate localization and state estimation of the ego-vehicle within the track.
To solve the localization problem, Extended Kalman Filter (EKF) solutions based on a vehicle model are usually adopted to fuse the measurements from different on-board sensors, like GNSS, LiDAR, Inertial Measurement Unit (IMU) and wheel speed odometry \cite{WISCHNEWSKI2019154}. LiDAR only localization solutions are proposed in \cite{massa} and \cite{schratter}, demonstrating an accuracy suitable for driving the Roborace DevBot 2.0 racecar within $100$ km/h.

On the perception side, the object detection problem in the Autonomous Racing field focused on camera-based solutions. In particular, cones detection using Convolutional Neural Networks (CNN) methods on cameras frame has been used by several teams in the FSD competition \cite{derita}\cite{V_disch_2022}\cite{puchtler}. 
A more robust and redundant solution is presented in \cite{kabzan}, including cone detection with feature-based and CNN-based approaches using both mono and stereo cameras, cone detection and color classification on LiDAR, and a sensor fusion solution based on the projection of the LiDAR in the camera reference system \cite{andresen2020accurate}.
Finally, \cite{strobel2020accurate} employs CNN both for cone detection and key points estimation for localization purposes.
The active detection of other dynamic agents in a racing domain is also detailed in \cite{betz5}, using an approach that is similar to that adopted in urban settings, where LiDAR, RADAR, and camera data are typically fused.

Global planning in the racing domain is approached by solving an optimization problem to retrieve the lowest lap time trajectory. Solutions based on minimum curvature and minimum time-based optimization considering the vehicle dynamics can be found in \cite{massaro}\cite{heilmeier}.

For local planning, sampling-based methods are a popular option due to their effectiveness in various kinds of robotics problems related to obstacle avoidance. Different versions of the Rapidly-exploring Random Tree (RRT) method have been presented, especially for FSD and small-scale platforms, combined with local controllers, predictions using a vehicle model, and curve refinement \cite{feraco}\cite{arslan}\cite{bulsara}. For full-scale vehicles and high-speed conditions, different local planners have been proposed, generating a graph of possible trajectories and choosing the one that minimizes a cost function defined on different criteria \cite{stahl2019}\cite{raji}. Other methods proposed optimization-based controllers considering obstacles and the free driveable area \cite{buyval}\cite{liniger}.

Regarding the controller module that tracks a certain reference path and speed profile, Model Predictive Control (MPC)-based methods had a great impact due to their advantages in considering complex systems with their inputs, outputs, and constraints, despite the burden on the algorithm design, modelling, and optimization for real-time usage. Some works focused on representing non-linear models of the vehicles \cite{novi}\cite{vazquez}, while others use simpler models considering uncertainties and constraining the control on some physical parameters\cite{wischnewski}. More classical controllers based on slip angle or feed-forward steering have reached similar performances in sport-cars \cite{laurense}\cite{kapania}. 

For what concerns the whole autonomous software stack, the literature presents several works related to the FSD competitions in which a brief description of each module is given \cite{chen}\cite{nekkah}\cite{tian}\cite{culley}.
\cite{kabzan} can be considered the most complete system paper presenting implementation details for the common modules needed to succeed on the FSD events, describing the adopted testing framework and the weak points for future improvements.
For full-scale racecars, \cite{caporale} and \cite{betz3} present their architecture for the Roborace competition giving particular attention to the motion planning and control modules, with very limited details on the object detection problem. Similar works not associated to any competition have been published. In \cite{funke}, a system architecture is presented focusing on localization, path planning, and control of a commercial sportcar at its limits, including the design of a safety module. \cite{funk} described the design of the hardware and software of an electric racecar autonomously driven on a challenging Swiss mountain road.

All the mentioned works do not consider multi-agent scenarios or they assume the information on other vehicles to be given.
\cite{betz5} presented the software architecture and methodology of the TUM Autonomous Motorsport team for the IAC. The authors described each module of the stack adopted during the first two events, including the head-to-head race.
A framework report can be found in \cite{urmson}, detailing the architecture of the vehicle that won the 2007 DARPA Urban Challenge. In this work, we present a complete autonomous stack for multi agent scenario, including additional details that we consider fundamental for high speed racing.
\section{Software Stack}
\label{sec:sw_stack}

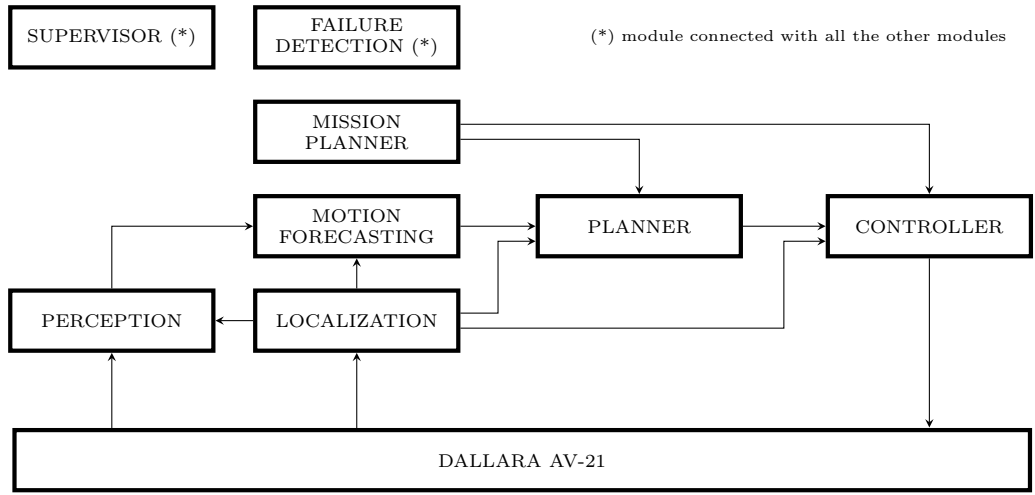
\begin{figure}[htb!]
  \centering

\begin{tikzpicture}[auto]
  \tikzset{
    block/.style = {
      draw,
      rectangle,
      minimum width=2.7cm, 
      minimum height=0.8cm,
      align=center,
      ultra thick,
      font=\scriptsize,}
  }
  \tikzset{
    contributed/.style = {
    }
  }
  \tikzset{
    thesis/.style = {
    }
  }

  \node[block,
        minimum width=13.5cm] (dallara) {DALLARA AV-21};
  \node[block,
        above left=1cm and -2.7cm of dallara] (perception) {PERCEPTION};
  \node[block,
        right=0.5cm of perception] (localization) {LOCALIZATION};
  \node[block,
        thesis,
        above=0.4cm of localization] (forecasting) {MOTION\\FORECASTING};
  \node[block,
        thesis,
        right=1.0cm of forecasting] (planner) {PLANNER};
  \node[block,
        contributed,
        right=1.1cm of planner] (controller) {CONTROLLER};
  \node[block,
        contributed,
        above=0.4cm of forecasting] (mission) {MISSION\\PLANNER};
  \node[block,
        above=0.4cm of mission] (failure) {FAILURE\\DETECTION (*)};
  \node[block,
        contributed,
        left=0.5cm of failure] (supervisor) {SUPERVISOR (*)};
  
  \node[right=1.6cm of failure, align=left] (note) {\tiny{(*) module connected with all the other modules}};
        	
  \draw[-stealth] (perception.north) |- (forecasting.west);
  \draw[-stealth] (forecasting.east) -- (planner.west);
  \draw[-stealth] (planner.east) -- (controller.west);
  \draw[-stealth] (localization.west) -- (perception.east);
  \draw[-stealth] (localization.north) -- (forecasting.south);
  \draw[-stealth] ($(localization.east) + (0, 0.1)$) -- ($(localization.east) + (0.5, 0.1)$) |- ($(planner.west) - (0, 0.2)$);
  \draw[-stealth] ($(localization.east) + (0, -0.1)$) -- ($(localization.east) + (4.3, -0.1)$) |- ($(controller.west) - (0, 0.2)$);
  \draw[-stealth] ($(mission.east) + (0, 0.1)$) -| (controller.north);
  \draw[-stealth] ($(mission.east) + (0, -0.1)$) -| (planner.north);
  \draw[-stealth] (controller.south) -- ($(controller.south) - (0, 2.25)$);
  \draw[-stealth] ($(perception.south) - (0, 1.0)$) -- (perception.south);
  \draw[-stealth] ($(localization.south) - (0, 1.0)$) -- (localization.south);
\end{tikzpicture}
  \caption{Diagram block with the software modules of \texttt{er.autopilot}. The modules marked with an asterisk are connected with all the other modules. }
  \label{fig:software_modules}
\end{figure}

The \texttt{er.autopilot 1.0} software stack consists of multiple modules following the Perceive-Plan-Act paradigm.
In \figref{fig:software_modules}, a block diagram with a high-level overview of the modules is shown.

A Localization module produces the state estimate of the ego vehicle used by all the other modules. By getting the data from the sensors of the AV-21, the Perception stack is able to detect the other vehicles and objects in the environment. This information is used mainly by the Motion Forecasting and Planner modules to predict the opponent's movement and generate a local path to avoid the collision or perform an overtake. A Mission Planner is integrated into the software stack as a behavioral planner able to get the signals from Race Control and from an internal state machine in order to give high-level decisions to the Planner and Controller, such as entering or exiting the pitlane and starting to overtake the opponent. Lastly, the Controller module is the one responsible to generate the correct actuation commands to be sent to the vehicle.

The modules, represented as nodes, communicate with each other using the ROS2 framework and Eclipse Cyclone DDS as middleware. Nodes are compiled into a shared library loaded at runtime which makes it possible to run multiple nodes in separate processes or as a single process. A base class has been defined for some nodes. For the control methods, the base class contains the callbacks needed to receive the vehicle state from the localization node and the actuation commands feedback, as well as some common methods. Each implementation extends the base class with the additional callbacks, methods, and interfaces to the libraries needed. Indeed, we decided to use the ROS2 nodes as wrappers for the communication, leaving the pure algorithmic parts as standalone software to be interfaced with.
The only logics implemented on the control base node are the safety checks on the lateral and heading errors of the path tracking performance. Three thresholds have been defined:
\begin{itemize}[noitemsep]
\item \texttt{max\_error}: Above this value, the target speed of the vehicle is linearly decreased.
\item \texttt{max\_error\_soft}: After passing this threshold the target speed is set to zero and the controller performs a soft stop. 
\item \texttt{max\_error\_hard}: Reaching this value, a hard brake is actuated over the command of the controller.
\end{itemize}

All the code executed on the car is written directly in C++ or generated from Matlab Simulink using the C-code generation tool. Python and Julia languages have been used for offline scripts related to data analysis, trajectory optimization and refinement, and visualization. A Docker container has been created for easy deployment on the vehicle and on the developers' machines, as well as on our online GitLab pipeline for basic testing.

\subsection{Localization of the ego vehicle}
\label{sec:localization}
The purpose of the localization module is to provide an estimate of the ego vehicle state using the available information from the sensor data or other software components.
Several architectures were investigated before converging to the final one, which is hereafter presented.

The vehicle is equipped with two GNSS modules. As explained in \autoref{sec:dallaraav21}, each of these provides the RTK-corrected position of the primary antenna, together with other information like the estimated speed and heading, and additional information on the quality of the position solution also called fix. 
Each receiver module is connected to an IMU that provides linear acceleration and angular rates, and is used to provide some pre-filtered signals useful for the user (such as estimated roll, pitch, and yaw angles). We decided to ignore these pre-filtered signals and only use sensor raw data, to retain a finer control on the localization pipeline. For a detailed description of the antenna's capabilities please refer to the producer website\footnote{\url{https://novatel.com/products/receivers/enclosures/pwrpak7d}}.

In the design, we had to consider several vehicle and operational domain requirements:
\begin{itemize}
    \item The two receivers called \textit{top} and \textit{bottom} receiver, had the antennas mounted on the main longitudinal axis and lateral axis of the car, respectively. We found the relative positioning of the antennas to slightly affect the quality of the fix, hence different weighting was considered for the two.
    \item The ego vehicle estimation has to be consumed by other modules, the fastest ones being the planning module (running at 50 Hz) and the control module (running at 100 Hz).
    \item The RTK correction was not always reliable, and the same holds for the GNSS signal, which is a common problem for these kinds of systems (for a more in-depth discussion please refer to \cite{massa}).
    \item The vehicle can reach a maximum velocity of around $300$ km/h, hence latency should be reduced to a minimum to guarantee a tight correspondence between the car's position on track and its latest available estimate.
\end{itemize}

For all the above reasons, we chose to equip our car with a robust localization filtering scheme based on an Extended Kalman Filter and to develop a LiDAR based localization system to further enhance the system's robustness in case the GNSS signal is lost.




\subsubsection{GNSS Localization}


The model used for the estimation is a simple kinematic unicycle in global coordinates \autoref{eq:ekf_model}. Given the absence of a side-slip angle sensor on the car, it is difficult to obtain a reliable estimate for this important quantity. Despite this simplification, which has been taken into account in the motion controller, the state estimate was accurate enough for the localization even at high speed.
\begin{align}
\begin{split}
    \dot x(t) &= v_x(t)\cos\left(\theta(t)\right) + \nu_x(t) \\
    \dot y(t) &= v_x(t)\sin\left(\theta(t)\right) + \nu_y(t) \\
    \dot \theta(t) &= \omega_z(t) + \nu_{\theta}(t)\\
    \dot v_x(t) &= a_x(t) + \nu_{v_x}(t)\\
    \dot a_x(t) &= \nu_{a_x}(t)
\end{split}
\label{eq:ekf_model}
\end{align}
The filter is implemented in such a way that it can manage the asynchronous data sources at the fastest possible rate, which is 250 Hz as per the IMU inputs. Model predictions are computed at the same frequency, while corrections are applied whenever new inputs are available. Measurements can come from different sources, such as GNSS or LiDAR based localization for the position and/or heading, wheel speed or GNSS speed for the vehicle velocity, IMU for the yaw rate. The quality of the incoming signals is evaluated partially at the sensor level (e.g. before sending the readings to the EKF) and partially at the filter level when receiving the data, before using it to compute a correction (e.g. by checking the consistency of the GNSS fix with the last estimated vehicle state, or by discarding measurements that are too old due to a lack of real-time processing capability on the on-board computer).

Despite its simplicity, this model allowed us to obtain a precise estimate of the car's position, heading, and longitudinal acceleration.

\subsubsection{LiDAR Localization}\label{sec:lidar_loc}


For the LiDAR vehicle localization, point clouds are first synchronized and merged together. Additionally, each cloud is individually motion-compensated using IMU data. 

\begin{figure}[!ht]
    \centering
    \includegraphics[width=0.9\textwidth]{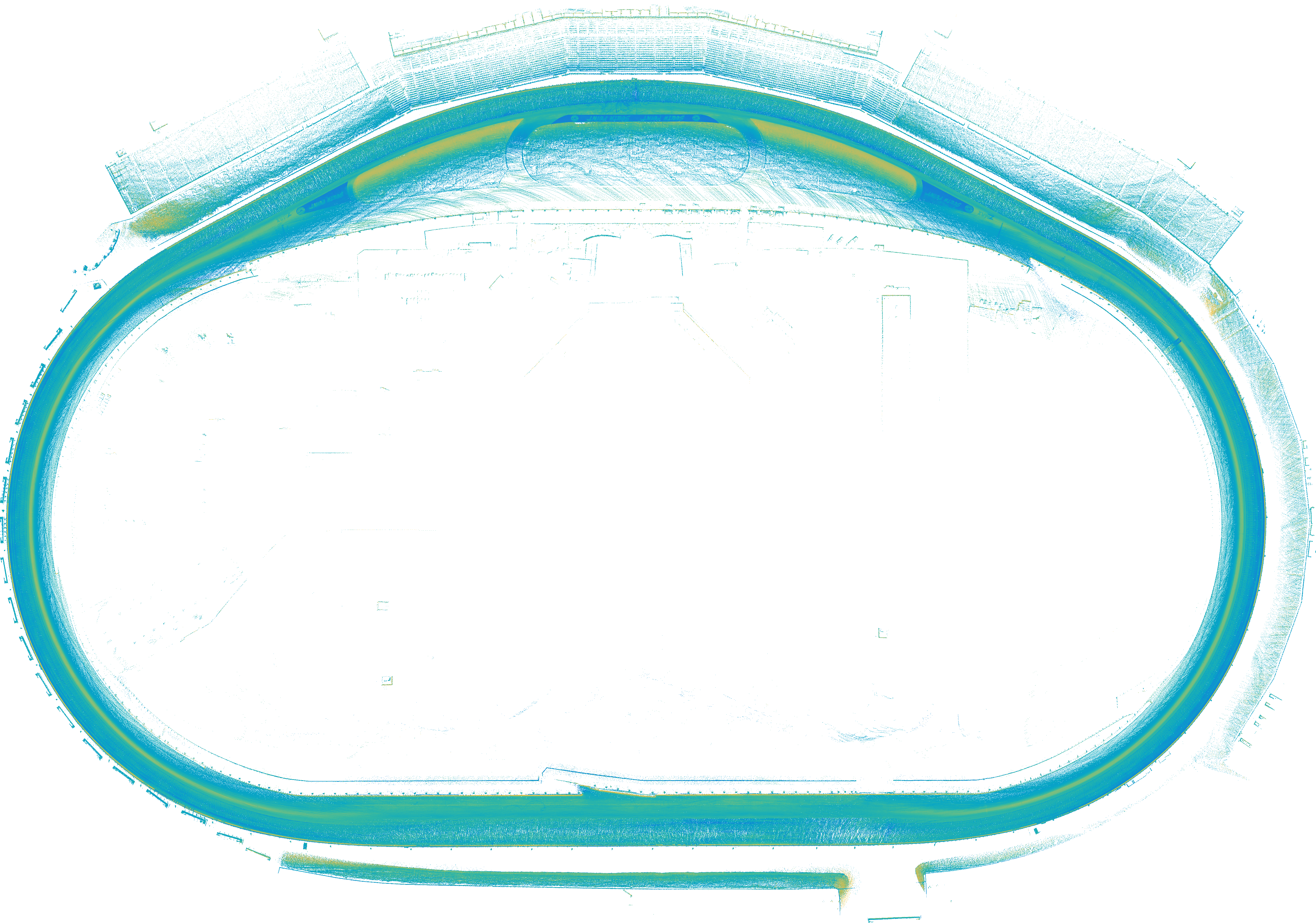}
    \caption{Top view of the LiDAR map obtained for the LVMS circuit. The color used for the points in the cloud is determined by the intensity value of each point.}
    \label{fig:lidar_map}
\end{figure}

Mapping is done offline, on a log covering the whole track at a slow speed, to enhance the map quality.
LiDAR clouds are aligned using a LiDAR Odometry and Mapping (LOAM) method aided by vehicle odometry and GPS. 
The obtained map is later globally optimized using GTSAM~\cite{dellaert2012factor}\footnote{https://github.com/borglab/gtsam} to maintain the shape and minimize the distance to the GPS trajectory.
The mapping process produces a georeferenced point cloud, that is then used by the LiDAR localization method. A top view of the resulting map is shown in \Cref{fig:lidar_map}. 

From the LiDAR depth map, vertical objects are extracted filtering the image by the pixel normal value. The filtered point cloud is used to localize the car on a 2D top-down map of the circuit.
The 2D map is a likelihood field of the filtered point clouds.
A particle filter approach is used to localize the car on the 2D map.
The particle filter is parallelized on GPU evaluating the probability of each point of each particle.
The extrapolated LiDAR localization can be used by the EKF as an alternative to (or together with) the GPS position.
\subsection{Perception}
\label{sec:perception}
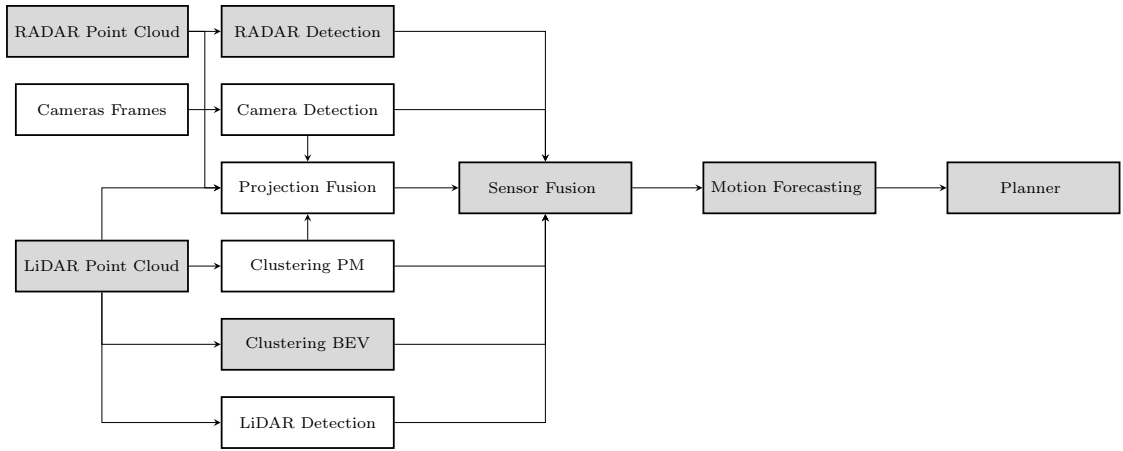
\begin{figure}[htb!]
  \centering
  \resizebox{\textwidth}{!}{

\begin{tikzpicture}[auto]
  \tikzset{
    block/.style = {
      draw,
      rectangle,
      minimum width=2.7cm, 
      minimum height=0.8cm,
      align=center,
      thick,
      font=\scriptsize,}
  }
  \tikzset{
    used/.style = {
       fill=gray!30!white,
    }
  }
  \tikzset{
    thesis/.style = {
    }
  }
\tikzset{
  -|-/.style={
    to path={
      (\tikztostart) -| ($(\tikztostart)!#1!(\tikztotarget)$) |- (\tikztotarget)
      \tikztonodes
    }
  },
  -|-/.default=0.5,
  |-|/.style={
    to path={
      (\tikztostart) |- ($(\tikztostart)!#1!(\tikztotarget)$) -| (\tikztotarget)
      \tikztonodes
    }
  },
  |-|/.default=0.5,
}

  \node[block, used, 
        above left=1cm] (clustering_bev) {Clustering BEV};
  \node[block,  
        below=0.4cm of clustering_bev] (lidar_det) {LiDAR Detection};
  \node[block,
        above=0.4cm of clustering_bev] (clustering_pm) {Clustering PM};
  \node[block, used,
        left=0.5cm of clustering_pm ] (lidar) {LiDAR Point Cloud};
  \node[block,
        above=0.4cm of clustering_pm] (fusion_proj) {Projection Fusion};
    \node[block, used,
        right=1.0cm of fusion_proj] (fusion) { Sensor Fusion };
    \node[block, used,
        right=1.1cm of fusion] (motion_fore) { Motion Forecasting };
    \node[block, used,
        right=1.1cm of motion_fore] (planner) { Planner };
  \node[block, 
        above=0.4cm of fusion_proj] (camera_det) {Camera Detection};
    \node[block,
        left=0.5cm of camera_det ] (camera) {Cameras Frames};
  \node[block, used,
        above=0.4cm of camera_det] (radar_det) {RADAR Detection};
    \node[block, used,
        left=0.5cm of radar_det ] (radar) {RADAR Point Cloud};

  \draw[-stealth] (radar.east) -- (radar_det.west);
  \draw[-stealth] (radar.east) to[-|-=.5] (fusion_proj.west);
  \draw[-stealth] (camera.east) -- (camera_det.west);
  \draw[-stealth] (camera_det.south) -- (fusion_proj.north);
  \draw[-stealth] (clustering_pm.north) -- (fusion_proj.south);
  \draw[-stealth] (lidar.east) -- (clustering_pm.west);
  \draw[-stealth] (lidar.north) |- (fusion_proj.west);
  \draw[-stealth] (lidar.south) |- (clustering_bev.west);
  \draw[-stealth] (lidar.south) |- (lidar_det.west);
  
  \draw[-stealth] (fusion_proj.east) -- (fusion.west);
  \draw[-stealth] (clustering_pm.east) -| (fusion.south);
  \draw[-stealth] (clustering_bev.east) -| (fusion.south);
  \draw[-stealth] (lidar_det.east) -| (fusion.south);
  \draw[-stealth] (camera_det.east) -| (fusion.north);
  \draw[-stealth] (radar_det.east) -| (fusion.north);
  
  \draw[-stealth] (fusion.east) -- (motion_fore.west);
  
  \draw[-stealth] (motion_fore.east) -- (planner.west);
  
  
\end{tikzpicture}
  }
  \caption{Perception scheme of \texttt{er.autopilot}.}
  \label{fig:percpeption_modules}
\end{figure}

The complete perception scheme of our stack is depicted in \Cref{fig:percpeption_modules}. All the blocks are hereafter discussed, but only the gray ones have been used for the races at IMS and LVMS.

The white blocks are implemented and are working, but we excluded these blocks for the following two main reasons. The first is the unsuitable accuracy, in fact, we did not trust some pipelines (e.g. LiDAR Detection) due to the high false positive rate. The second, related to the camera pipelines, is a bandwidth issue we experienced while reading the six cameras with all the other sensors. It occurred several times that reading the cameras' streams led to a higher drop of LiDAR packets or even to LiDAR failure that was irreversible and needed a system reboot. Considering that the LiDAR is the sensor with the most reliable detections, we sacrificed the cameras to not have such a problem during racing days. We will investigate how to solve this issue to exploit those high potential sensors as well.

\subsubsection{Drivers, Settings and Calibration}

Before introducing the algorithms that run on the sensors' data, the acquisition, configuration, and calibration of the sensors need to be discussed. 

The raw data coming from the LiDARs, cameras, and RADAR (and also GNSS) are collected using our own-made drivers instead of the official ROS2-based ones provided by the sensors' manufacturers. 
This has been done to manage all the low level data we consider useful that are not contemplated in the ROS2-based drivers and to reduce the delay between sensor data reading and usage, especially for big data such as frames or point clouds, on our perception algorithms.

We decided to use all six Mako cameras, with a resolution of $1032\times772$ pixels and a frequency of $10$ FPS.
We utilized a limited resolution and frequency in our setup to prevent band saturation. 
We used all three Luminar LiDARs, setting a FOV of $15\deg$, the Gaussian pattern, the center at $0\deg$ at IMS ($1\deg$ at LVMS due to banking), and a frequency of $20$ Hz. 
We employed a Gaussian pattern to increase the point density on the horizon, while the field of view and layer number were optimized to maintain a 20Hz frequency while preserving the point density. 
For the RADAR, we decided to employ only the frontal one, using the default settings and a frequency of $10$ Hz, the only available option for the given sensor.

Thereafter, we took care of the sensors' calibration. Firstly we performed intrinsic camera calibration exploiting a checkerboard pattern $8\times6$ printed on a rigid panel and the Kalibr tool~\cite{oth2013rolling}. Then we implemented and performed camera-LiDAR extrinsic calibration with the same pattern, matching the checkerboard detected from the camera frames with the ones recognized in the LiDAR depth images. At the end of these procedures, we knew both the intrinsic parameters of the cameras and the transformation with respect to all the involved sensors. 
\subsubsection{LiDAR Clustering - Bird's-eye-view approach}
\label{perpipe:lidar_bev}

The LiDAR Clustering  Bird's-eye-view (BEV) pipeline takes the LiDAR point cloud as input, removes the ground, and gives as output tracked objects clusters. It only executes on the CPU, which makes it robust to GPU failures. 
To remove the ground, the normals of the point cloud have been exploited. For each point, its $x$, $y$, and $z$ normals have been computed and the points have been filtered on the norm value on the vertical axis (z). Additionally, all the points higher than a certain threshold (i.e. 3m) and the points belonging to the ego vehicle have been removed. 

Once the cloud is processed, a BEV image of the remaining points is built. 
On that image we run a Connected Component algorithm, to group the points into objects. 
That computes the clusters that we can reproject on the point cloud. 

For a more stable detection, we also inserted a tracker in the pipeline. 
The tracker tracks the position, and in particular the center of the cluster, using an EKF, and matches the objects in different iterations using a nearest neighbors technique. 



\begin{figure}[h!]
    \begin{subfigure}[b]{\textwidth}
        \includegraphics[width=\textwidth]{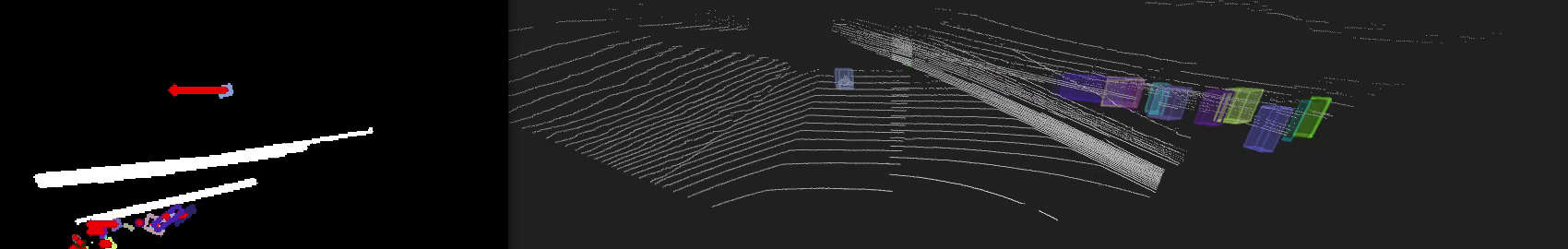}
    \end{subfigure}
    
    \begin{subfigure}[b]{\textwidth} 
        \includegraphics[width=\textwidth]{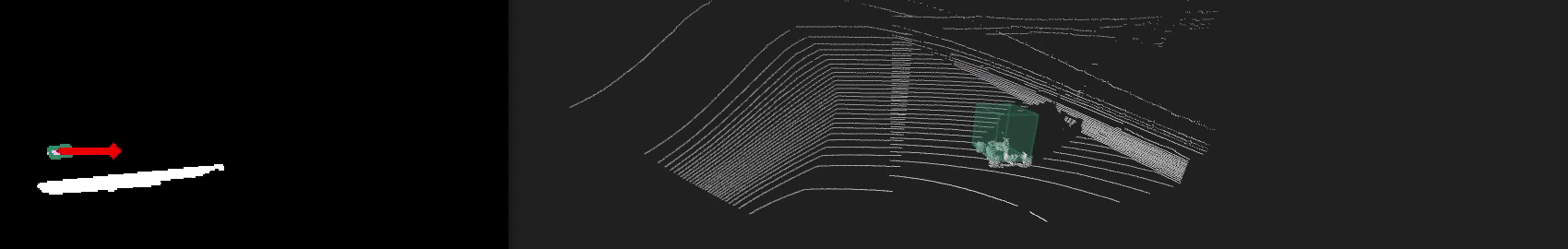}
    \end{subfigure}
    \caption{Results of the clustering BEV pipeline. The algorithm employs a low-resolution representation of the BEV, depicted on the left, where the white lines indicate the track walls, and their respective clusters are ignored. Instead, the rectangles represent the clusters, color-coded based on their local tracker, and the red arrows indicate their speed vectors. On the right side, the clusters are projected back onto the cloud.} 
    \label{fig:clus_bev_res}
\end{figure}

Some visual results are depicted in \Cref{fig:clus_bev_res}, while overtaking a vehicle (top), where the vector representing the velocity (red arrow in the BEV) is negative, and while overtaken (bottom) where the velocity of the opponent is positive. 

\subsubsection{LiDAR Clustering - Point Map approach} 
\label{perpipe:lidar_pm}

The LiDAR clustering Point Map (PM) pipeline takes the LiDAR point cloud as input, removes the ground, and gives as output objects clusters. It executes both in GPU and CPU and it is an alternative to the other clustering algorithm.

At first, the point cloud is converted into a Point Map, an image that contains at least 3 channels that, instead of representing RGB values, are the position of the single point in the 3D environment $x,y,z$. Besides position, information such as intensity, time of flight, and ring index can also be included in the PM channel. The whole pipeline then uses this converted PM. 

\begin{figure}[!h]
    \centering
    \includegraphics[width=\textwidth]{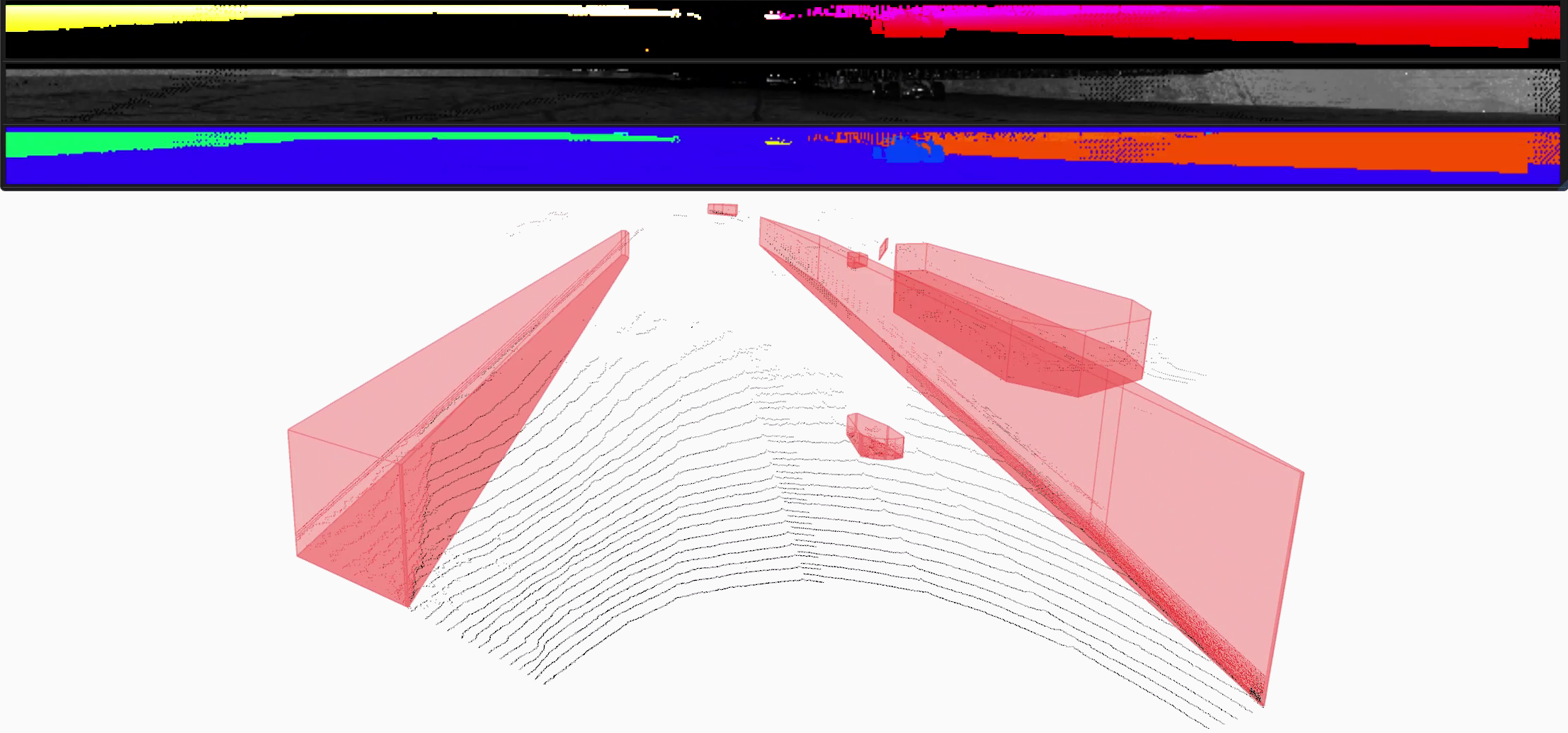}
    \caption{Results of the clustering PM pipeline. On the top, there are three LiDAR PM, in order without the ground, the plain PM, and the one with clusterized components. On the bottom, the clusters are projected on the cloud. }
    \label{fig:clus_pm_res}
\end{figure}

In this approach, introduced in \cite{costi2022realtime}, the ground is removed with an upgraded version of the Line Fit Ground Segmentation \cite{himmelsbach2010fast}. 
Then the clustering is computed using again a Connected Components approach, but in this case on the PM rather than on a BEV. The clustering algorithm is subdivided into two main steps. The first one works directly on the Point Map exploiting a Connected Component algorithm to compute neighbors and label them with the same ID, executing entirely on the GPU. 
In the second step, running on CPU instead, the neighbors' data extracted from the PM are elaborated to aggregate neighbors in different clusters. The results are reported in \Cref{fig:clus_pm_res}. 

\subsubsection{Camera detection}
\label{perpipe:camera_det}

The camera detection pipeline takes the camera frames as input and gives as output tracked detected vehicles. It executes both in CPU and GPU, but most of the computation is performed on the latter.

The detection of the other AV-21 vehicles is performed using an Object Detection Convolutional Neural Network. We adopted YOLOv4~\cite{bochkovskiy2020yolov4}, implemented via tkDNN~\cite{verucchi2020systematic}, a custom framework that optimizes its performance on Nvidia GPUs.

To correctly detect open-wheel racecars, we trained the model on an open source dataset, Deep Drive BDD100k~\cite{yu2020bdd100k} to learn road objects (cars, bikes, pedestrians, and so on). Then we collected data on different tracks (LOR, IMS, LVMS), down-sampled the different logs on the different cameras, and manually labeled almost $400$ images using the LabelImg tool\footnote{https://github.com/heartexlabs/labelImg}.
Using those labeled images, we fine-tuned the network only for the car class. Eventually, the network only detects open-wheel racecars. 

\begin{figure}[h!]
    \begin{subfigure}[b]{0.32\textwidth}
        \includegraphics[width=\textwidth]{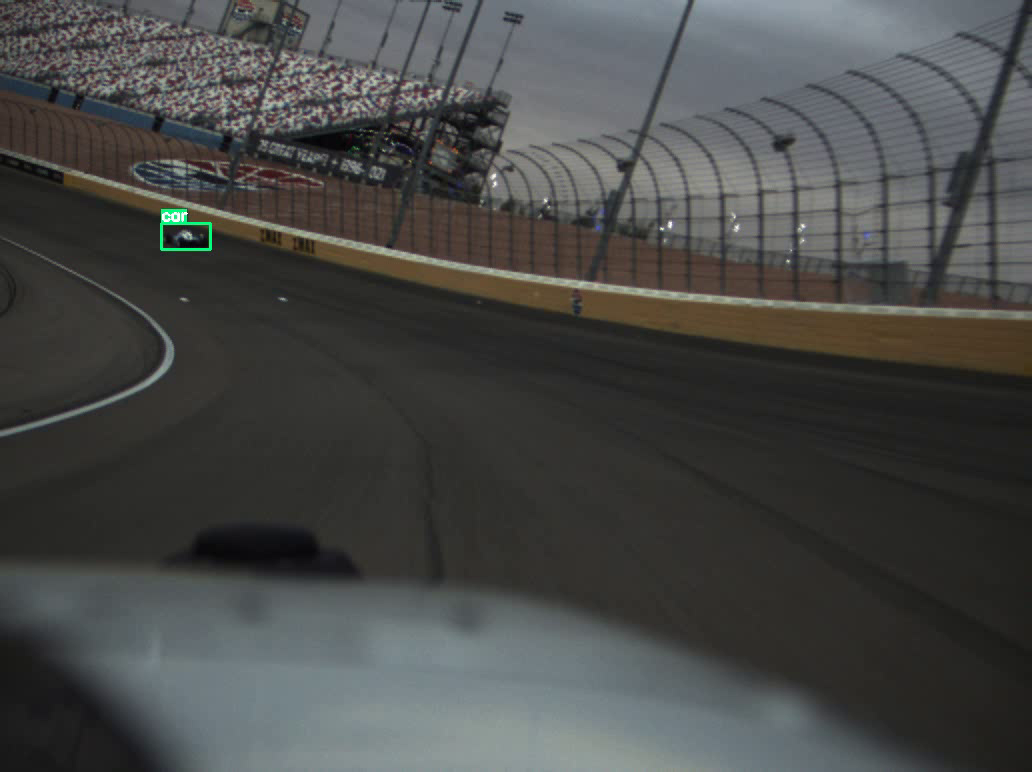}
    \end{subfigure}
    \hfill
    \begin{subfigure}[b]{0.32\textwidth} 
        \includegraphics[width=\textwidth]{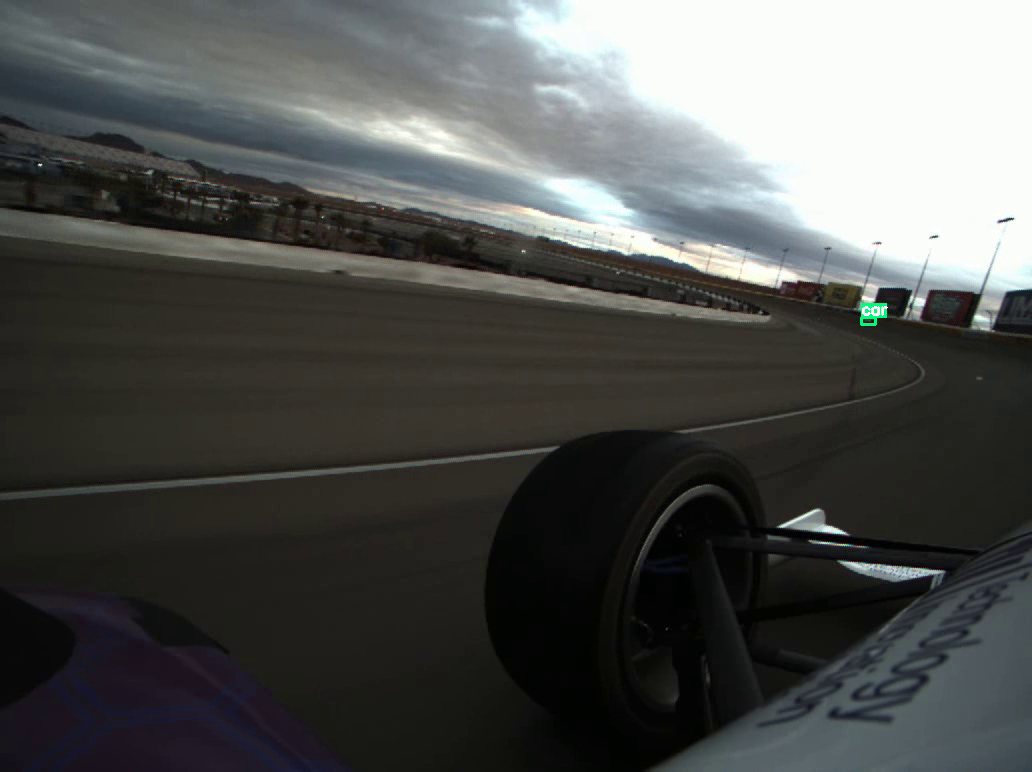}
    \end{subfigure}
    \hfill
    \begin{subfigure}[b]{0.32\textwidth} 
        \includegraphics[width=\textwidth]{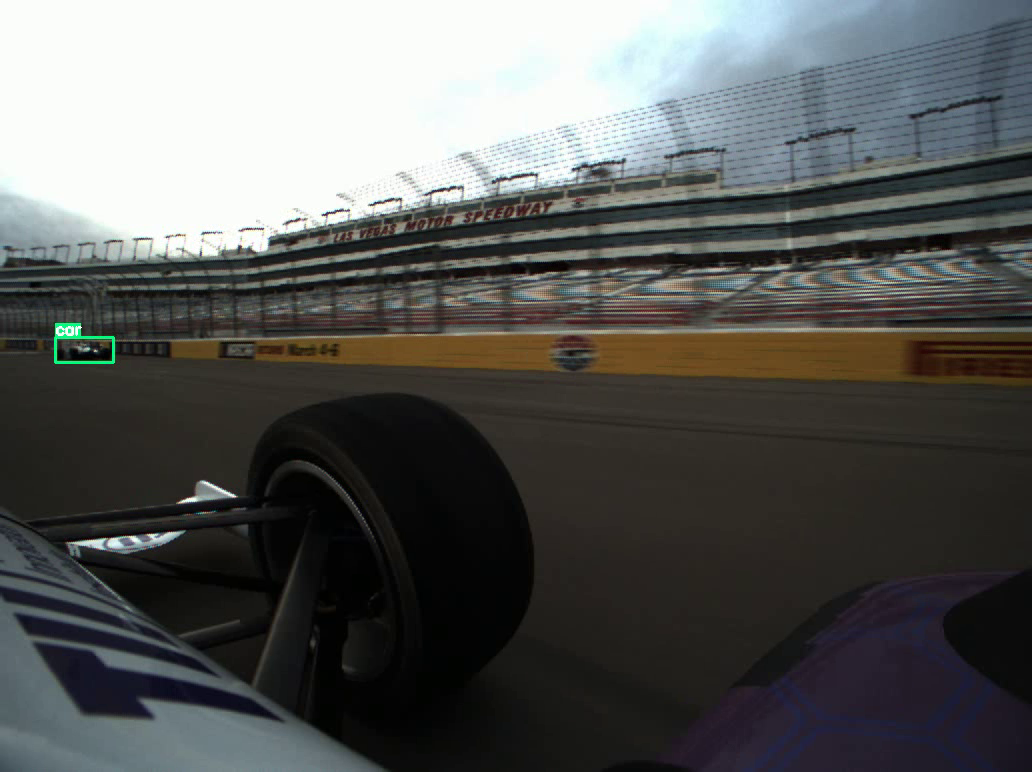}
    \end{subfigure}
    
    \begin{subfigure}[b]{0.32\textwidth}
        \includegraphics[width=\textwidth]{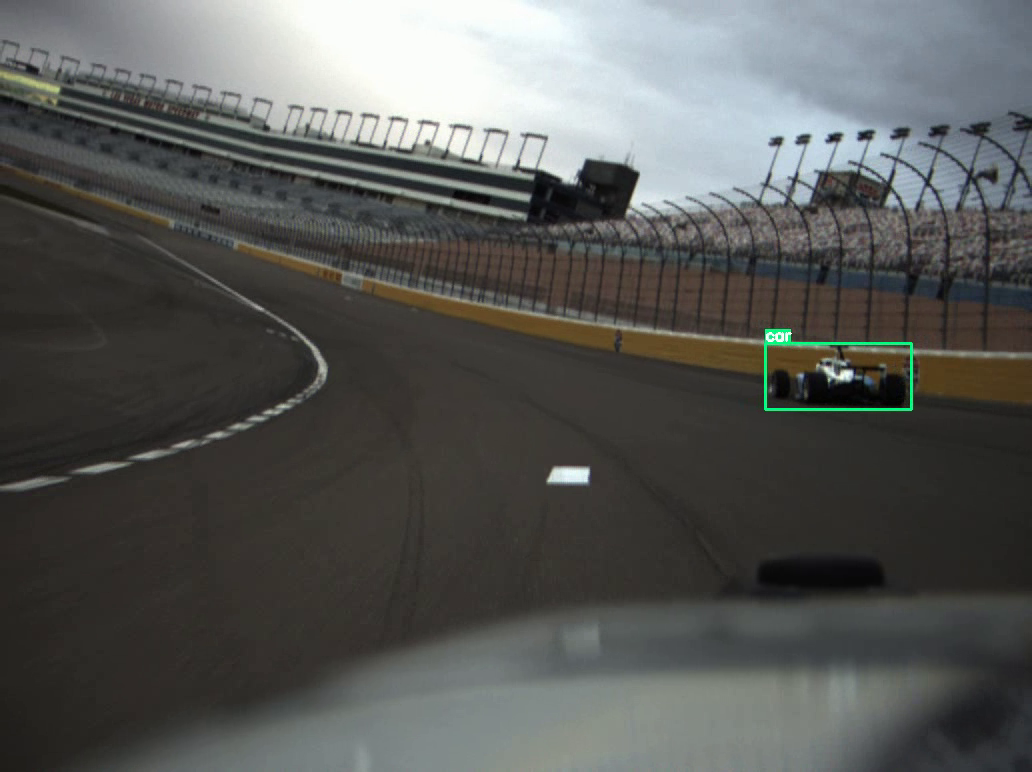}
    \end{subfigure}
    \hfill
    \begin{subfigure}[b]{0.32\textwidth} 
        \includegraphics[width=\textwidth]{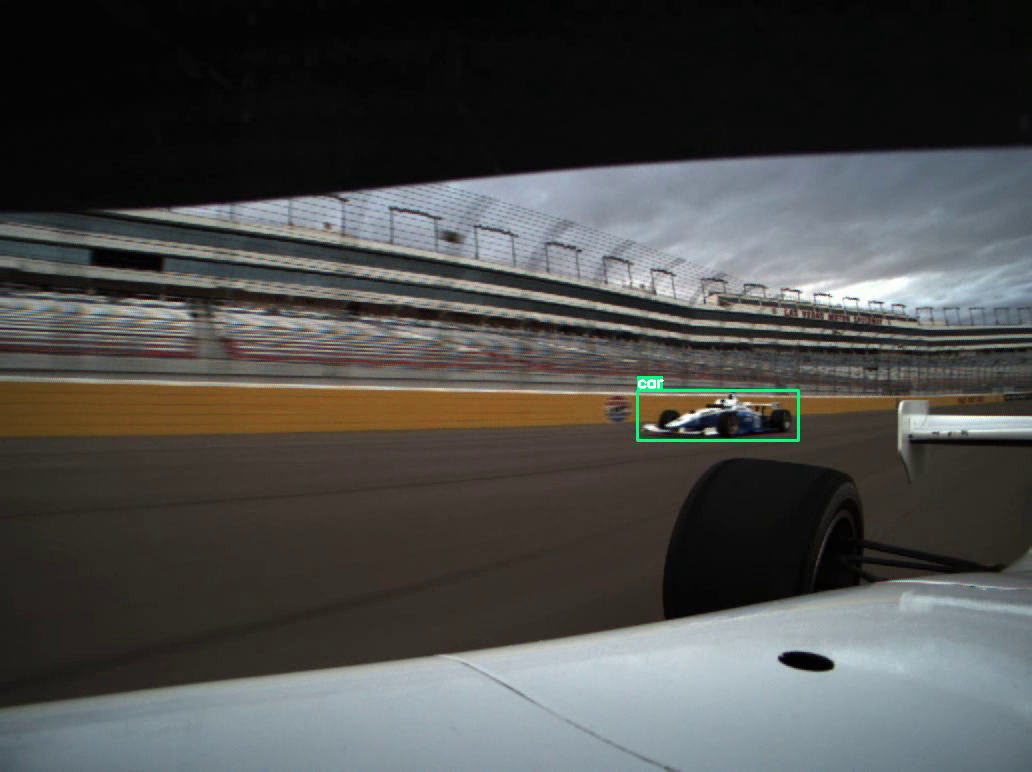}
    \end{subfigure}
    \hfill
    \begin{subfigure}[b]{0.32\textwidth} 
        \includegraphics[width=\textwidth]{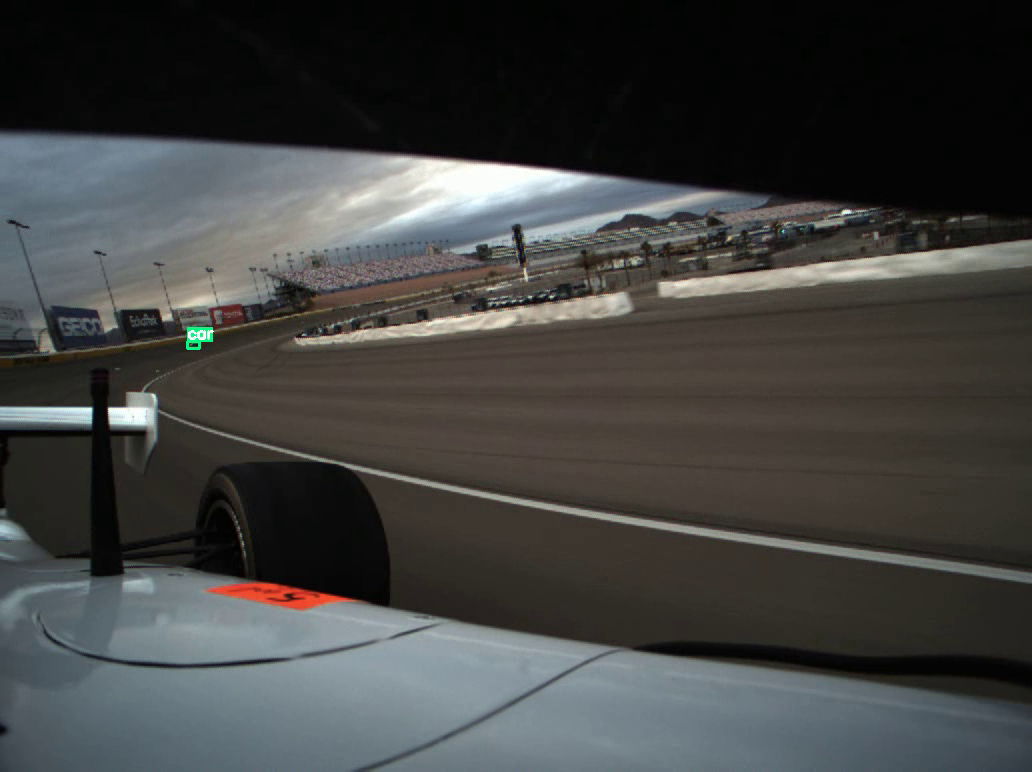}
    \end{subfigure}
    \caption{Results of vehicle detection on the six different camera views.}
    \label{fig:cam_det}
\end{figure}

Once trained, the network has been deployed on tkDNN\footnote{https://github.com/ceccocats/tkDNN}, which uses TensorRT\footnote{https://developer.nvidia.com/tensorrt} and CUDA kernels to optimize each network layer. The visual result of the detection trained network for all the cameras is reported in \Cref{fig:cam_det}. 

To estimate the distance of the object, a simple geometric approach reported in~\Cref{eq:geom_dist} has been adopted, based on the intrinsic calibration of the cameras (in particular their \texttt{focal\_length}), the actual height of the vehicle in mm \texttt{object\_h\_mm}, and the height of the detected vehicle in pixels \texttt{object\_h\_pixel}. 

\begin{equation}
    \texttt{dist} = \frac{\texttt{object\_h\_mm} \cdot \texttt{focal\_length}}{\texttt{object\_h\_pixel}} 
    \label{eq:geom_dist}
\end{equation}

We also implemented an NN-based method to estimate the distance, with a simple Encoded-Decoder approach, but the results were unsatisfactory.  

Finally, the same tracker used for the LiDAR clustering BEV pipeline has been used to track the vehicles in the frames.

\subsubsection{LiDAR detection}
\label{perpipe:lidar_det}

A very similar approach to the camera detection has also been applied to the LiDAR-based detection. From the point clouds we have constructed LiDAR images based on the intensity of the points. Using this format, we have collected a dataset of almost $600$ images and we have manually labelled them. 

\begin{figure}[h!]
    \begin{subfigure}[b]{\textwidth}
        \includegraphics[width=\textwidth]{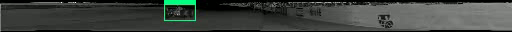}
    \end{subfigure}
    
    \begin{subfigure}[b]{\textwidth} 
        \includegraphics[width=\textwidth]{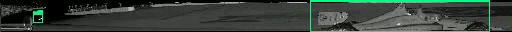}
    \end{subfigure}
    \caption{Results of vehicle detection on frontal (top) and lateral (bottom) LiDAR intensity images.} 
    \label{fig:lid_det}
\end{figure}

We modified Deep Drive BDD100k images to monochrome images, adapting the format to the LiDAR images one $(16:1)$. We then trained YOLOv4 on the modified BDD100k, to then fine-tune it on the $600$ labelled images. 
The results of vehicle detection are reported in \Cref{fig:lid_det}. 

In this case, the objects' distance is given by the LiDAR, so there is no need for estimation. For the tracking, yet again we adopted the EKF mentioned tracker.

\subsubsection{RADAR Detection }
\label{perpipe:radar_det}

The RADAR detection pipeline takes the RADAR point cloud as input and gives as output tracked moving objects, executing on CPU.

The point cloud given by the RADAR is already processed, and it is not possible to retrieve the raw data. Therefore we only applied filtering to the input data, considering only the stable moving objects lying inside the track boundaries.

\subsubsection{Projection Fusion}
\label{perpipe:prj_fusion}

The Projection Fusion pipeline takes as input the LiDAR point cloud, the RADAR point cloud, the detected vehicles from the camera, and the clusters of the LiDAR Clustering PM pipeline. It gives as output detected vehicles that have been recognized at least by two different sensors. It only executes on the CPU. 

The proposed approach exploits camera projection to properly fuse detected objects from camera images with 3D estimations. The algorithm tries to estimate the 3D location in world coordinates for each detected vehicle. The projection converts vertices from the world coordinate system to the camera pixel coordinates system with \Cref{eq:camera_projection}.

\begin{equation}
\centering
\left[\begin{array}{c} u\\ v\\ z \\ \end{array}\right] = K [R|T] \left[\begin{array}{c} X\\ Y\\ Z \\ \end{array}\right]
\label{eq:camera_projection}
\end{equation}

Where $K$ is the camera intrinsic matrix,  $[R|T]$ is the camera extrinsic calibration matrix, $[u, v,z]$ is the undistorted point in camera pixel coordinates, and finally $[X,Y,Z]$ are the real world coordinates. It is worth mentioning that accurate intrinsic and extrinsic calibrations are required to reach satisfactory results.

The LiDAR point cloud is fused with camera detections with the following steps. 
(i) \Cref{eq:camera_projection} is applied to all the points of a cloud, for each camera. 
(ii) All the LiDAR points projected outside the camera bounding box are filtered out, therefore a frustum of 3D points is considered for each object. 
(iii) A single point in the frustum is chosen as the Point-Of-Interest (POI), sorting all the points by distance and picking the nearest (the first element), the median (the element in the middle of the array), or another custom array position. 
(iv) Finally, the location of the object is estimated as the average between all the points in the neighborhood of the POI. 

The RADAR point cloud is fused differently with the camera detections. From the RADAR point clouds, we have a single point for each object, therefore a single point is fused with each camera bounding box with the following steps. 
(i) RADAR points are projected on-camera images. 
(ii) The matching cost is estimated as $\frac{| cx - x |}{(w/2)} $, 
where $cx$ is the horizontal coordinate of the camera box center, $x$ is the horizontal coordinate of the RADAR projected point, and $w$ is the width of the camera box.  
(iii) The Hungarian algorithm~\cite{kuhn1955hungarian} is applied to matchboxes and RADAR points using the cost previously computed.
(iv) Finally, incorrect matches are filtered out via a user-defined threshold on the cost.

LiDAR clusters are considered as already detected 3D objects from another source. Similarly to RADAR-camera fusion, a single cluster is fused with each camera-detected object, with the following steps.
(i) Each point of the cluster is projected on the camera and the bounding box of these points is calculated.
(ii) The matching cost is computed as the inverse of Intersection-over-Union (IoU) between the camera box and cluster projected box.
(iii) Hungarian matching is then in charge of the matching and, finally, (iv) bad matches are filtered out with a threshold.

At this point, there are multiple pairs of objects from the camera and another 3D source, and two steps are yet to be performed: fusion among all the cameras and fusion of all the object pairs.
There are two straightforward cases: (i) if the objects belong to the same camera, the objects are fused with the same bounding box, then it’s the same object, otherwise, it’s not; (ii) if detected vehicles from multiple cameras are fused with the same cluster, then it’s the same object.
For the other cases, aggregation is not as simple, and the proposed approach exploits 3D coordinates from fused 3D sources. A matrix cost is computed considering 3D box reprojection and euclidean distance, combined with a weighted sum. Matches with costs exceeding a threshold parameter are discarded.

At this stage, a list of aggregated objects containing all the camera boxes and all the LiDAR, RADAR, and cluster positions has been obtained. 
Every 3D pose of the aggregated object has a score associated, calculated proportionally to the focal length of the camera fused with it. Higher confidence is given to objects detected from cameras with higher focal lengths because the field of view is narrower and the boxes are bigger. This score is used as a weight for a weighted average that gives the final aggregated object 3D position in world coordinates.

\subsubsection{Sensor Fusion Module}
\label{perpipe:sensor_fusion}

All the presented pipelines flow into a Sensor Fusion module. This acts as an aggregator for all the different detection pipelines active on the machine. In particular, it deals with the transformation of the raw detections from local to global coordinates, the association between the new detections and the ones already tracked, and the prediction of their movement.

The node aggregates several pipelines of detections from various sensors located at different positions in the car, and each of them produces detections in its own reference system. 
In order to aggregate them together, we decided to transform every detection to global coordinates via~\Cref{eq:tf_trassform}, using the ego vehicle position computed by the localization node.

\begin{equation}
T_{\text{global\_obj}} = T_{\text{loc}}\,\,T_{\text{sensor}}\,\,T_{\text{local\_obj}}
\label{eq:tf_trassform}
\end{equation}

Each $T_{\text{i}}$ is a 4x4 transformation matrix. $T_{\text{global\_obj}}$ is the object pose in global coordinates, $T_{\text{loc}}$ is the car pose given by the localization node, $T_{\text{sensor}}$ is the sensor pose relative to the car CoG, and $T_{\text{local\_obj}}$ is the pose of the detected object locally to its sensor.

The estimated position of each detection is endowed with uncertainty, due to the method or the accuracy of the sensor itself, as well as the position estimated by the localization module.
Therefore, when applying ~\Cref{eq:tf_trassform} the localization error is propagated into the position estimation of the detections. For this reason, it is also necessary to propagate the localization covariance on the detection covariance. 

The object tracking can work in two coordinate systems: (i) Cartesian $(x,y)$ and (ii) Frenet $(s,d)$, in which the central trajectory of the track is used as a reference.
In each case, a Kalman filter with a material point model is used. 
For the Cartesian version the state is $ [x ~ y ~ V_x ~ V_y]^T$ and the correction $[x ~ y]^T$, while for the Frenet version the state is $[s ~ d ~ V_s ~ V_d ]^T$ and the correction $[s ~ d]^T$ 
The Kalman filter calculations are the same in both cases, a simple Cartesian to Frenet transformation is applied on the input and vice versa on the output.

To summarise, the algorithm is composed of the following steps:
\begin{enumerate}
    \item Kalman prediction.
    \item Filtering of detections if (i) outside the track, (ii) inside the area of the ego vehicle, (iii) they have high covariance. 
    \item Detection association, with Hungarian matching and Mahalanobis distance.
    \item Kalman correction.
    \item Creation of new tracklets, i.e. tracked objects, if far from existing tracklets.
    \item Removal of tracklets with too large covariance (not corrected for too long).
\end{enumerate}

All the tracklets that are active are unique detections that are then passed to the motion forecasting module and finally to the planner. 

\subsection{Planning}
\label{sec:planning}
To be able to safely avoid static obstacles and perform overtakes at high speeds, as requested by the IAC competitions at IMS and LVMS, in addition to the global planner which produces offline the optimal racing line on each track, we implemented the modules needed to predict the other agent's movement and generate a local trajectory considering all the static and moving obstacles in the surroundings. We give here an overview of the proposed solution, while a more detailed report can be found in \cite{raji}.

Considering the race rules limiting the defender to keep the inner line, one strategy could have been to switch directly to a racing line positioned in the outer lane of the track as soon as the role of the ego car becomes the attacker. On one hand, this approach can be considered safer since the two vehicles keep separate lines for most of the time except during the line switching performed in safe moments. On the other hand, staying on the outer lane where usually there is more dirt could result in less grip at high speeds and longer distances with respect to the defender's inner line. Considering our research interest in creating solutions suitable for unconstrained racing scenarios with more than two vehicles on track, we decided to perform the overtakes once in the proximity of the opponent keeping the same racing line followed while defending.

\subsubsection{Global Planner}
\label{global_planner}
A minimum-time optimization problem is solved for the global planning, formulating the nonlinear problem in JuMP and solving it using IPOPT. The dynamics of the vehicle, presented in Section \ref{sec:control}, are transformed in the spatial domain discretizing the continuous space model with a discretization distance.
The cost function is defined as
\begin{equation}
\label{eq:global:cost}
   J_{opt}(x_k, u_k) = -\dot{s}_k + u^T Ru + B(x_k) \,,
\end{equation}
where $x$ is the state vector, $u$ is the inputs vector, $\dot{s}$ is the progress rate, $B(x_k) = q_B \alpha_r^2$ is a regularization term which penalizes the rear slip angle $\alpha_r$, and $u^T Ru$ regularizes the inputs rates.
The overall problem is defined as
\begin{align*}
    \begin{split}
        \min_{X, U} &\ \ \sum_{k=0}^{N} = J_{opt}(x_k, u_k) \\
        s.t. &\ \ x_{k+1} = f_s^d(x_k, u_k)\,, \\
        &\ \ f_s^d(x_N, u_N) = x_0\,, \\
        &\ \ x_k \in X_{track} \quad x_k \in X_{ellipse}\,,\\
        &\ \ a_k \in \bm A, \, u_k \in \bm U,  \ k = 0,\dots,N, 
    \end{split}
\end{align*}
where $X=[x_0, ..., x_N]$, and $U=[u_0, ..., u_N]$ are the state and input sequences respectively. $X_{ellipse}$ is a constraint on the friction ellipse, and $X_{track}$ represents a track constraint. $\bm{A}$ and $\bm{U}$ are respectively box constraints on the physical inputs $a$ and their rate of change $u$.

\subsubsection{Motion Forecasting}
The goal of motion forecasting is to estimate the future trajectory of the vehicles detected by the perception module. The estimated trajectories are then used by the motion planning algorithm to avoid collisions.

For each obstacle, the perception module provides a unique identifier and its position in a Cartesian frame.
Given the sequence of the position of an obstacle, the goal is to predict its future trajectory. We employed a Kalman filter with a model defined in a Frenet frame.

Given the position of the $i$-th obstacle in a Cartesian frame $x_i(k), y_i(k)$, the position of the obstacle in the Frenet frame $s_i(k), n_i(k)$ is computed.
The model of the obstacle is defined as:
\begin{align}
  \label{eq:forecasting:model_dot_s}
  \dot s_i(k + 1) & = \dot s_i(k) \\
  \label{eq:forecasting:model_d}
  n_i(k + 1)      & = n_i(k)
\end{align}
Equation\;\eqref{eq:forecasting:model_dot_s} states that the longitudinal speed of the obstacle is constant, whereas Equation\;\eqref{eq:forecasting:model_d} indicates that the lateral displacement from the reference path is constant.

This simple model exploits the fact that the only objects of interest on the track are other cars that will follow a racing line similar to the one that the ego car is following. For this reason, we decided to define the model in a Frenet frame that uses the race line as the reference path.
Moreover, we can assume that the cars will run all the time at an almost constant speed because the oval shape of the tracks involved does not require as many decelerations and accelerations as in a course road track. From Equation\;\eqref{eq:forecasting:model_dot_s}, \eqref{eq:forecasting:model_d} the state space model used in the Kalman filter can be derived.

To summarize, at each step, for every obstacle, the following steps are applied to predict its future trajectory:
\begin{enumerate}[noitemsep]
  \item The new measurement $\hat s_i(k), \hat n_i(k)$ is computed from $\hat x_i(k), \hat y_i(k)$.
  \item Using the new measurement, the Kalman filter is updated with a prediction step, followed by a correction step.
  \item The future trajectory of the obstacle is predicted by applying $m$ consecutive prediction steps to the Kalman filter.
  \item The trajectory is converted back into the Cartesian frame.
\end{enumerate}

\subsubsection{Local Planner}

The local planner is an extension of \cite{werling}, as it computes the trajectory generation in a Frenet coordinate frame, where the following adjustments have been implemented to satisfy the needs of our racing scenario:
\begin{itemize}[noitemsep]
  \item The main reference used for the Frenet frame is the optimal racing line generated by the global planner.
  \item The time interval $T$ between each node of the trajectories is kept constant since the controller requires a trajectory with a fixed length in time.
  \item The collision check of the trajectories set is performed in the Frenet frame to avoid converting the trajectories into a Cartesian frame. Rather than doing the checks on the polynomials, we sampled each trajectory in a finite number of points by a time interval $\Delta_T$.
  \item Improved collision check method adding a soft constraint to avoid the edge cases when only hard constraints are considered and to be safer in case of noises in the localization and the control loop.
  For each trajectory $\tau_i$, a collision coefficient $\gamma_i \in [0, 1]$ is computed, where $\gamma_i = 0$ indicates that the trajectory is not colliding with any obstacle, whereas $\gamma_i = 1$ indicates that the trajectory is violating the safety margins (hard constraint). Then the total cost computed is:
    \begin{equation}
      C_{tot,i} = k_{lat} C_{lat,i} + k_{lon} C_{log,i} + k_{soft} \gamma_i^2
    \end{equation}
    with $k_{lat}, k_{lon}, k_{soft} > 0$.
    
    To compute $\gamma_i$ we decided to exploit the Euclidean distance from the safety margin.
    For every trajectory $\tau_i$ the minimum distance $d_i$ from the safety margin is computed.
    Then $\gamma_i$ is defined as
    \begin{equation}\label{eq:collision_coef}
      \gamma_i = \max \left \{1 - \frac{d_i}{\Delta_{soft}}, 0 \right \}
    \end{equation} 
    where $\Delta_{soft} > 0$ is a parameter to enlarge or reduce the effect of the soft constraint.
    \item The initial conditions required to generate the set of trajectories are calculated by projecting the car's position on the best trajectory at the previous step. At the very first step of the planner instead, the initial conditions are calculated purely on the car's position.
    \item A different distance keeping mode. The desired speed used in the generation of the longitudinal movements is calculated by a simple proportional controller which considers the opponent's speed, the desired distance to keep, and the current distance. This mode is used when the rules or the race control are not permitting us to perform an overtake.
\end{itemize}
Along with the main planner, a simple emergency planner is run, whose solution is used when the Supervisor module detects a failure in the main planner or in the modules it depends on.
The emergency planner continuously extends the last feasible planned path with a smooth polynomial in the Frenet coordinate frame, being able to make the car move to the inner side of the track or enter the pit lane, if requested by the race control.

\subsubsection{Mission Planner}
\label{sec:mission_planner}
The mission planner is responsible for generating the reference signals and instructions used by the Local Planner based on the position of the car on the track, the phase of the race, and the flags received from Race Control.
In particular, it controls when the car can enter or exit the pit lane, if the car has to perform the warm-up lap, which is the maximum allowed speed, whether the car is allowed to overtake or not, and which is the minimum distance that the car has to maintain from the opponent if it is not allowed to overtake.

\begin{figure}[htb!]
  \centering
  \begin{tikzpicture}[auto]
  \tikzset{
    block/.style = {
      draw,
      rectangle,
      minimum width=2.7cm, 
      minimum height=0.8cm,
      align=center,
      ultra thick,
      font=\scriptsize,}
  }
  
  \node[block] (init) {INITIALIZATION};
  \node[block,
        below=0.5cm of init] (pitexit) {PIT EXIT};
  \node[block,
        below=0.5cm of pitexit] (run) {RACE};
  \node[block,
        right=1cm of pitexit] (pitentry) {PIT ENTRY};
  \node[above=0.7cm of init] (start) {start};

  \draw[-stealth] (start.south) -- (init.north);
  \draw[-stealth] (init.south) -- (pitexit.north);
  \draw[-stealth] (pitexit.south) -- (run.north);
  \draw[-stealth] (run.east) -| (pitentry.south);
  \draw[-stealth] (pitentry.north) |- (init.east);
\end{tikzpicture}
  \caption{Overview of the mission planner state machine. Each of the shown states groups more specific states and transitions which depend on the car's position, actuators, and engine status, and external signals sent by Race Control.}
  \label{fig:mission:sm_overview}
\end{figure}
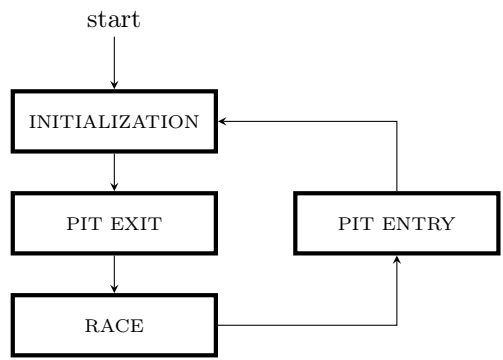

A fundamental requirement for the mission planner is to be easy to change because it needs to rapidly adapt to changes in race rules.
For this reason, we used a Finite-State Machine (FSM) to define the logic. A high overview of the FSM is shown in Figure \ref{fig:mission:sm_overview}. 

For the implementation, we relied on \texttt{scxmlcc} \footnote{scxmlcc: https://github.com/jp-embedded/scxmlcc}, an open-source tool that autogenerates the C++ code of the state machine from an XML file that defines the states, the events and the transitions.
\subsection{Modelling and Control}
\label{sec:control}

Given the current state of the car, the controller computes the actuation commands to track the reference trajectory produced by the local planner illustrated in Section~\ref{sec:planning}.
At high speed, a Model Predictive Controller (MPC) is used to control simultaneously the steering, the throttle, and the brake.
The vehicle model used in the MPC is a dynamic single-track model identified from a high fidelity multi-body simulation. A kinematic model was also developed to work accurately at low speed. Due to the limited testing time, the integration between the kinematic and the dynamic models wasn't implemented, and the control at low speed (below 100 kph) has been delegated to a Pure Pursuit algorithm and a PID controller, respectively for the steering and the pedals.
A hysteresis and a consistency check on the steering wheel commands of the two controllers are applied to switch safely between the two solutions based on the operational conditions.
The gearbox is controlled via a state machine that, based on the RPM of the engine, selects the appropriate gear.

\subsubsection{Modelling}

\begin{figure}[htb!]
  \centering
  
  \begin{subfigure}[t]{0.56\textwidth}
    \includegraphics[
        width=0.99\textwidth,
    ]{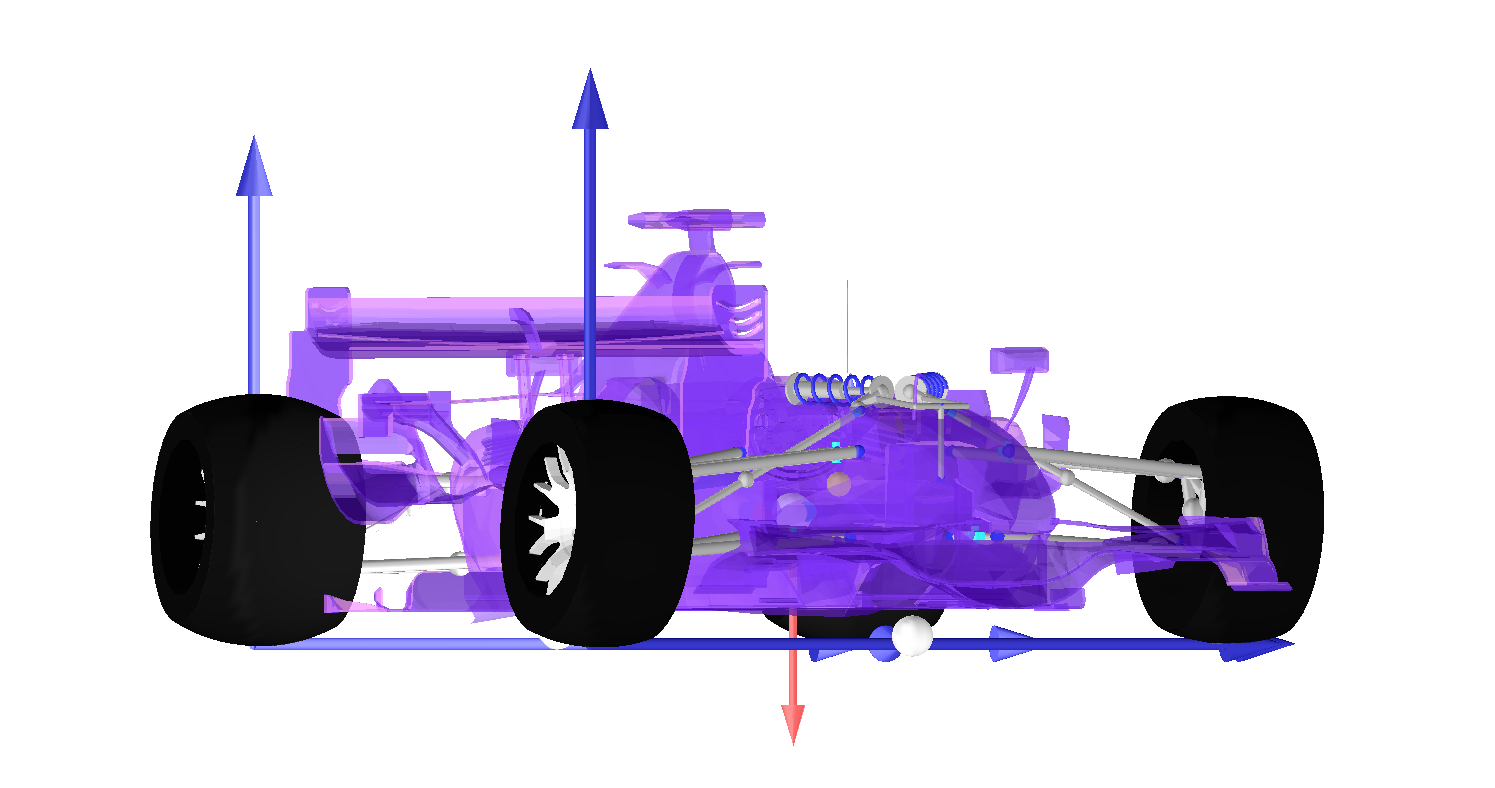}
    \caption{Vehicle animation}
  \end{subfigure}
  \begin{subfigure}[t]{0.43\textwidth}
    \includegraphics[
        width=0.99\textwidth,
        trim={3cm 13cm 3cm 3cm},
    ]{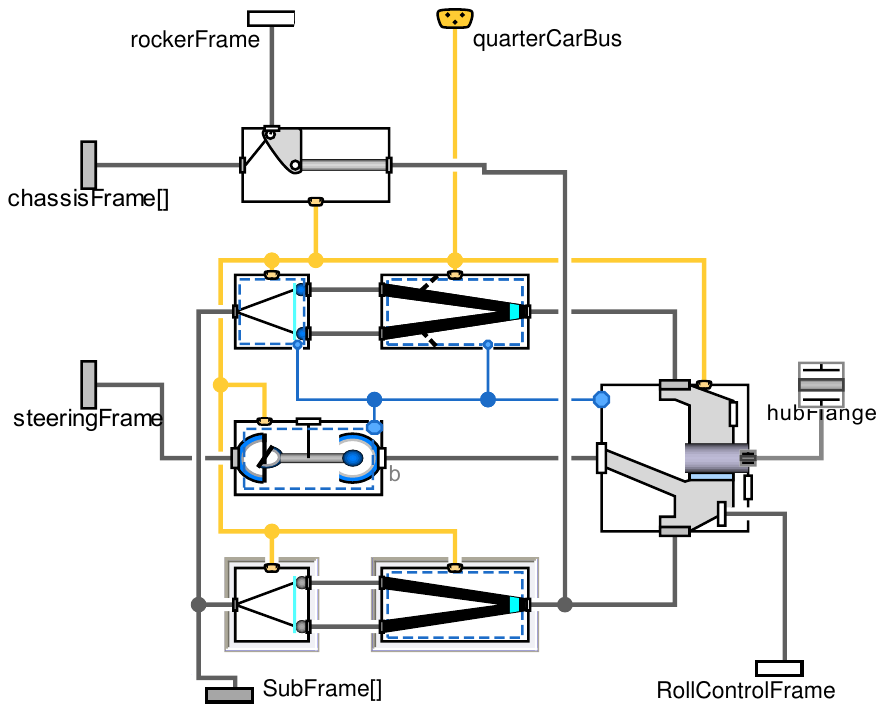}
    \caption{Blocks of the suspensions subsystem}
  \end{subfigure}

  \caption{Multi-body model built on Dymola using the Claytex VeSyMA Motorsports libraries.}
  \label{fig:Dymola_Model}
\end{figure}

Prior to the testing time and physical access to the car, we developed a multi-body model of the AV-21 on Dymola \cite{dempsey} using the VeSyMA - Motorsports libraries provided by Claytex{\footnote{\href{https://www.claytex.com/}{https://www.claytex.com/}}}. The first set of parameters derives from data provided by the IAC organizers and the vehicle's components manufacturers. The remaining unknown details have been estimated from available information on similar vehicles and commercial racing-game simulators like RFactor 2{\footnote{\href{https://www.studio-397.com/}{https://www.studio-397.com/}}}. The model has been refined after gathering experimental data on the track. In \autoref{fig:Dymola_Model}, an overview of the model on Dymola is shown, in which the front suspensions subsystem blocks scheme is highlighted.

\begin{figure}[htb!]
  \centering
  
  \begin{subfigure}{0.44\textwidth}
    \includegraphics[
        width=0.99\textwidth,
        trim={2.8cm 9cm 3cm 9cm},
        clip
    ]{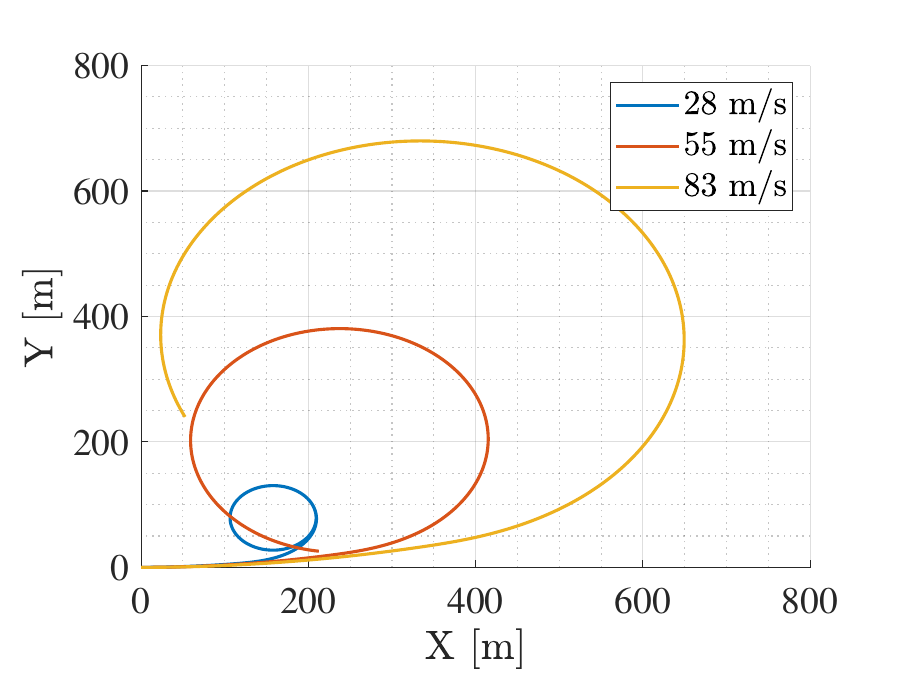}
    \caption{Trajectory}
  \end{subfigure}
  \begin{subfigure}{0.44\textwidth}
    \includegraphics[
        width=0.99\textwidth,
        trim={3cm 9cm 3cm 9cm},
        clip
    ]{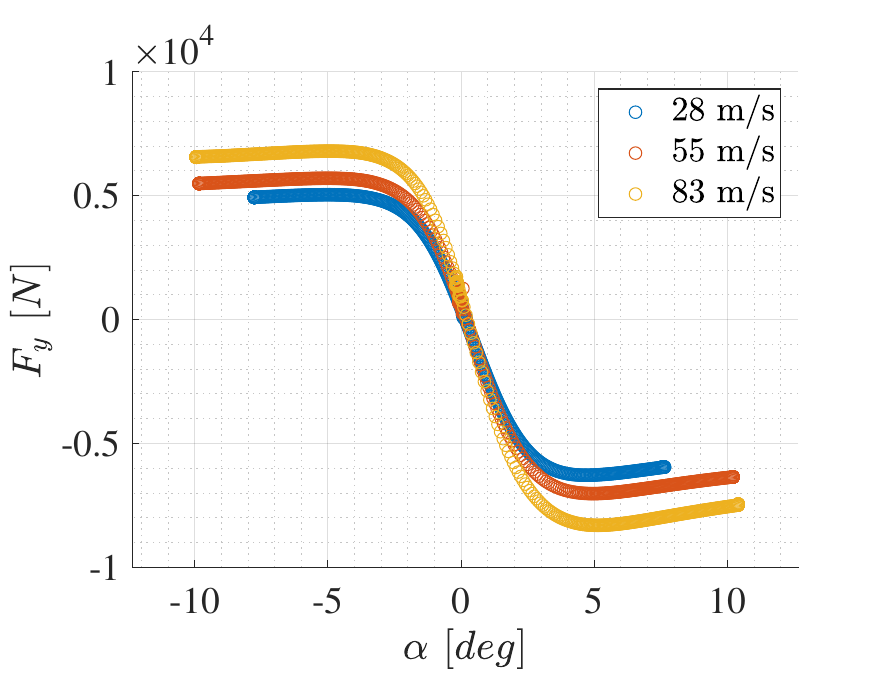}
    \caption{Front axle characteristic}
  \end{subfigure}

  \caption{Virtual test results of the ramp steer manoeuvre at three different speeds.}
  \label{fig:Ramp_test}
\end{figure}

Particular attention has been given to the following components:
\begin{itemize}[noitemsep]
  \item Tires: the force-slip model is based on a Pacejka Magic Formula 6.2 \cite{Pacejka1991THEMF} with Kelvin-Voigt spring-damper vertical load, combined slip parameters and neglecting the relaxation length. Inflation pressure and camber angle are considered asymmetrical for the right and left sides due to the setup for oval track. 
  \item Powertrain: a 3D diagram with engine torque, RPM, and the throttle position has been implemented starting from the engine test bench data. The gear ratio, final drive, and shift time have been defined from telemetry and manufacturers' data.
  \item Aerodynamics: a simple model with drag and downforce coefficients is used. The centre of pressure is positioned between the front and rear axle to define the correct aero balance.
  \item Suspensions: the modelled components include the double wishbone geometry with a vertical anti-roll bar, the rocker, and the shock absorber. Stiffness and damping are defined for all components.
  \item Body: the sprung and unsprung masses are modelled considering the centre of gravity and cross-weight to validate the experimental data of static load balance. The inertia matrix is defined as well.
\end{itemize}
The multi-body model has been used to produce manoeuvres that cannot be easily replicated on the real vehicle due to the lack of suitable space and limited testing time. In \autoref{fig:Ramp_test} we report the ramp steer manoeuvres produced at different speeds. The wheels' steering angle is set to vary from zero to the maximum value at the rate of 1°/s, while the vehicle speed remains constant.

\begin{figure}[htb!]
	\centering
	\includegraphics[width=0.70\columnwidth,
	trim={2cm 19cm 2cm 1.5cm},
    clip
	]{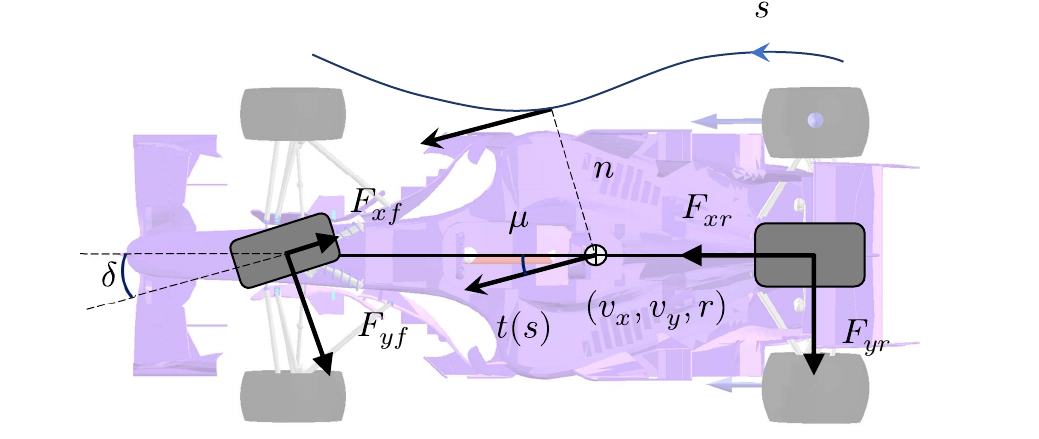}
    \caption{Dynamic single-track model on curvilinear coordinates, where $s$, $n$, and $\mu$ are the progress along the path, the orthogonal deviation from the path, and the local heading. $Fx_{f,r}$ and $Fy_{f,r}$ are the longitudinal forces and lateral forces arising from the interaction between tires and the ground. ($v_x$,$v_y$,$r$) identify the motion field of the centre of gravity and $\delta$ is the steering angle.}
	\label{fig:ST_Model}
\end{figure}

The high fidelity model has been ported to a single-track model on curvilinear coordinates, shown in \autoref{fig:ST_Model}, including the forces due to the road bank angle, the aerodynamic effects, the longitudinal forces on the rear axle generated by the turbo-charged engine and the tire forces represented with a simplified Pacejka Magic Formula considering the vertical load and camber angle as well as the combined slip effects. The equations of motion and the explanation of the modelled forces are presented in \cite{raji}.

\subsubsection{Pure Pursuit Controller}
\label{sec:control:pp}

An extension of \cite{Coulter-1992-13338} has been developed. The target point is chosen at a curvilinear distance \texttt{lookahead} from the projection of the car's position on the local path, hence the reference curvature is obtained as
\[
k_{\text{pp}} = 2 \psi_{\text{target}} / \texttt{lookahead},
\]
where $\psi_{\text{target}}$ is the angle of the target point position with respect to the \textit{x-axis} of the local reference frame. The curvature is then converted to a steering angle at the wheels using the classical kinematic steering model:
\[
\delta_{\text{wheel}} = \arctan{(k_{\text{pp}} \cdot \texttt{wheelbase})}.
\]
The \texttt{lookahead} is updated at each step depending on the current speed and lateral error, in both cases with a contribution proportional to a reference value.

\subsubsection{Warm-up manoeuvre}

Without tire warmers and considering the low ambient temperatures during the race events, which were a maximum of 12.2$^\circ C$ (54$^\circ F$) on October 21, 2021, at IMS and 17.2$^\circ C$ (63$^\circ F$) on January 7, 2022, at LVMS, it was extremely important to heat the tires as much as possible. One approach, followed by the majority of the teams, consisted in incrementally increasing the speed of the vehicle during the first couple of warm-up laps following a normal raceline and running at least one lap at a speed higher than 50 m/s (180 km/h) where it has been demonstrated that the tires' temperatures increase rapidly. However, this approach produces higher energy and therefore higher temperature on the rear axle with respect to the front axle. 

For this reason, during the first couple of laps, we performed an open-loop warm-up manoeuvre at 25 m/s (90 km/h) which consists of a series of $\pm$ 80deg steering wheel angle commands on top of the Pure Pursuit algorithm. 

The manoeuvre has been produced considering the following parameters:
\begin{itemize}[noitemsep]
  \item \texttt{steer\_val}: the steering wheel angle that should be commanded during the manoeuvre;
  \item \texttt{step\_duration}: the amount of time during which \texttt{steer\_val} is kept;
  \item \texttt{step\_gap}: the amount of time between two \texttt{step\_duration}, during which the steering controller is not overridden;
  \item \texttt{curvature\_threshold}: a value for checking whether to reduce \texttt{steer\_val} based on the curvature of the path in front of the car.
\end{itemize}
This solution aims to increase the temperature of the front tires, which is important to reduce the probability to occur in an understeering condition, before setting a higher speed and following the global trajectory to increase more homogeneously the temperature on all the tires; some results of this manoeuvre will be later discussed in \autoref{subsec:planning_control}. 

\begin{figure}[htb!]
  \centering
  
  \begin{subfigure}{0.49\textwidth}
    \includegraphics[
        width=0.99\textwidth,
        trim={2.5cm 9cm 3cm 9cm},
        clip
    ]{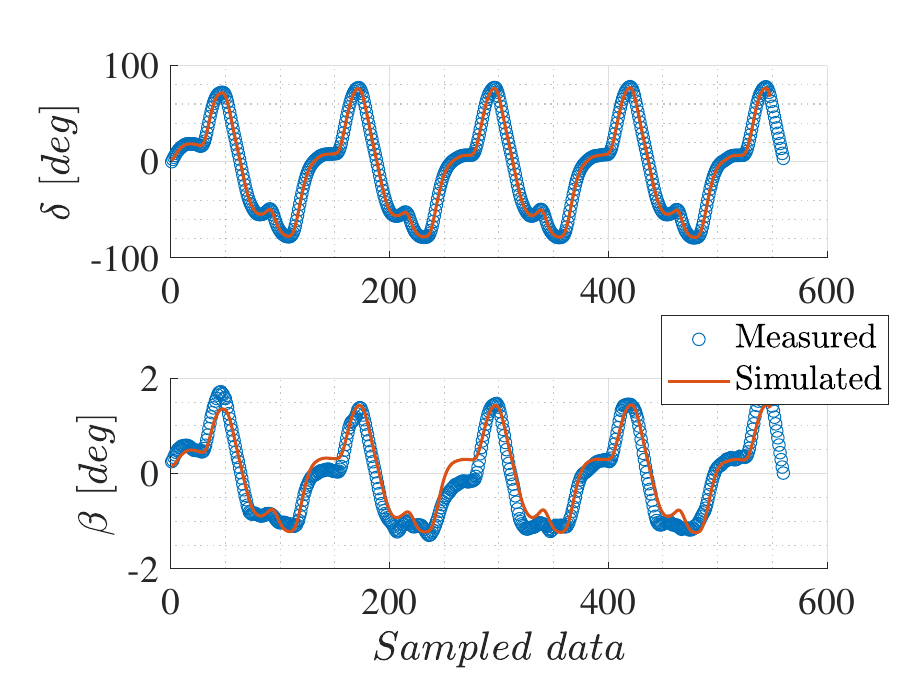}
    \caption{The steering angle $\delta$ (a model input) and the sideslip angle $\beta$ (an output signal).}
  \end{subfigure}
  \begin{subfigure}{0.49\textwidth}
    \includegraphics[
        width=0.99\textwidth,
        trim={2.8cm 9cm 3cm 10cm},
        clip
    ]{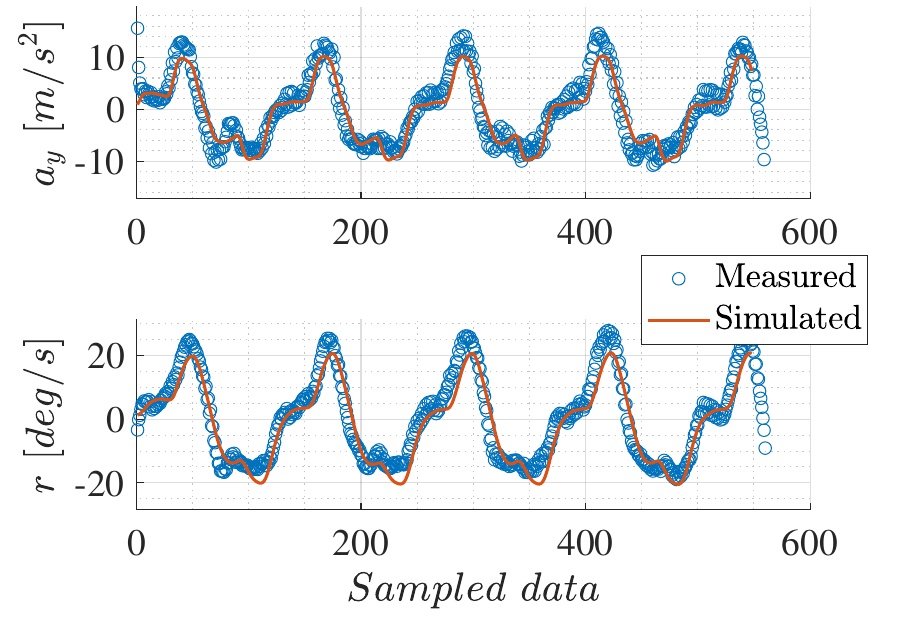}
    \caption{Lateral acceleration and yaw rate.}
  \end{subfigure}

  \caption{Virtual test results on warm-up manoeuvre compared with the experimental data.}
  \label{fig:Dymola_Warm-up}
\end{figure}

The stability of the vehicle during the warm-up manoeuvre has been evaluated on the multi-body model developed on Dymola. In \autoref{fig:Dymola_Warm-up} simulation data are compared with measurements acquired during tests with an additional optical speed sensor.

\subsubsection{Model Predictive Controller}
\label{sec:control:mpc}

The MPC used at high speed is an extension of \cite{vazquez}, where the optimization problem is formulated as in \ref{global_planner} using the model discretized in time $f_t^d(x_t, u_t)$. The main differences from the original work are:

\begin{itemize}[noitemsep]
  \item A more complex model
  \item The path and velocity produced by the Frenet-based planner are used as a reference to be tracked, considering the cost function
  \begin{equation*}
   J_{MPC}(x_t, u_t) = -\dot{s}_t + q_n n_t^2 + q_\mu \mu_t^2 + q_v | s_{v,t} | + u^T Ru + B(x_t)\,,
    \end{equation*}
    where additionally to the terms used in \eqref{eq:global:cost}, it includes the path following weights $q_n$ and $q_\mu$, and a velocity tracking weight $q_v$ on the slack variable $s_{v,t}$.
    \item The optimization problem is solved using HPIPM \cite{frison}. 
    \item An automatic differentiation library, CppADCodeGen \footnote{\href{https://github.com/joaoleal/CppADCodeGen}{https://github.com/joaoleal/CppADCodeGen}}, is exploited to obtain the derivatives of the non-linear differential equations of the model producing the source code which is statically compiled offline and linked dynamically at runtime. This led to a speedup of the MPC keeping the computational execution below the 10ms on high-end Intel processors such as E-2278GE, i7-10750H, and similar.
\end{itemize}

Considering the uncertainties of the dynamics of the actuator at speeds never tested before and to cope with a potential model mismatch related to the force offset of the asymmetrical setup of the AV-21, we decided to set a high value to the costs on the physical inputs $a$ and their rate of change $u$. This has been done expecting to avoid critical oscillations and to keep a smooth movement in exchange for a slower system and a potentially higher path tracking error.

\subsubsection{Controller Mux}
\label{sec:controller-mux}
The Pure Pursuit controller (Section \ref{sec:control:pp}) and the Model Predictive controller (Section \ref{sec:control:mpc}) run in parallel, each producing a control command.
Both these commands are sent to a Controller Mux node, which selects one of the two sources based on their priority and availability, and routes its message to the hardware.

In \texttt{er.autopilot 1.0} the Model Predictive controller has the highest priority, followed by the Pure Pursuit controller.
In the rare event in which the Model Predictive controller fails to find a solution or it doesn't provide a control message at the required rate, the Controller Mux switches to the Pure Pursuit controller. As mentioned in (Section \ref{sec:control:mpc}), the MPC is not used at low speed, therefore the Controller Mux uses the Pure Pursuit for this case as well. The decision is made by checking a flag sent by the controllers indicating whether their commands should be applied.

During a switch, the commands are interpolated to smoothly match the ones of the new command source. This is done to avoid sudden changes in the control commands that could lead to undesired behavior.

\subsection{Supervisor and Safety Layer}
\label{sec:supervisor}

\subsubsection{Supervisor}

The supervisor module coordinates all the software modules. In particular, it takes part in the start-up sequence of the car and commands an emergency stop if an anomaly is detected by the failure detection module. It listens to the Mission Planner presented in \autoref{sec:mission_planner}, the Race Control, and to the joystick used by the pit crew to trigger a manual emergency stop.
Besides this main Supervisor module, \texttt{er.autopilot 1.0} uses a concept we called MicroSupervision. Each ROS2 node of the software stack has some checks on the liveliness of the most important data such as the vehicle state (position and velocity), the commands feedback, and the status of other modules. In particular, the controller base node has the possibility to directly stop the car. This can be seen as a redundancy in the general safety system of the architecture.

\subsubsection{Failure detection}

The failure detection module is responsible for detecting anomalies in the system. One of its tasks is to monitor the signal of all the car’s sensors to check if the values are in the nominal ranges or if the sensors give the correct outputs, excluding for example \textit{NaN} or values with a wrong scale/range. Some sensible parameters related to the engine, transmission, fuel, and battery have additional checkups related to the optimal operating range in order to guarantee peak performance. When some of these sensors are out of their optimal values, but in acceptable ranges for a certain amount of time, the failure detection module sends a warning to the supervisor. On the other hand, if the sensors reach critical values, the emergency signal is triggered. 

Another task of this module is to monitor the status of all the software stack in order to notify the supervisor if some of them trigger an error state or stop working. In the latter, the failure detection module monitors the timestamps of the messages sent for communication purposes in order to check for timeout conditions and as a redundancy takes advantage of the QoS (Quality of Service) API exposed by the ROS2 middleware interface (namely \texttt{DDS} \footnote{Data Distribution Service: \url{https://www.omg.org/spec/DDS/} })  in order to have confirmation for a potential crash of the modules. 

Furthermore, it checks that the connection between the car and the base station is alive. If an anomaly is detected, an error is sent to the supervisor module which reacts accordingly.

\section{Simulation}
\label{sec:simulation}
Two simulation environments have been used to test the software stack, considering the following criteria:

\begin{itemize}
    \item Vehicle Dynamics fidelity: the simulated vehicle handling should behave similarly to the real one and should be easy to test different road friction coefficients, tire temperatures, and tracks.
    \item Simulation to Reality gap: should be limited the differences from the reality for what concerns the steps followed on the real car for pit entry and exit, and the signals sent between the race control and the pitcrew. The sensors' interfaces and communication protocols used should be replicated as well.
    \item Ease of use: each team's developer should be able to run the entire software stack and the simulator on the same machine, and easily restart and change the simulation scenario.
\end{itemize}
A single simulator that satisfies all these conditions is still in development, where the aim is to include the multibody model developed on Dymola into the simulator described in \ref{unity-sim}.

\subsection{AssettoCorsa}
\label{assettocorsa}
AssettoCorsa\footnote{\href{https://www.assettocorsa.it/}{https://www.assettocorsa.it/}} is a racing-game simulator developed by Kunos Simulazioni. It is popular for its realistic dynamics and for the possibility to be easily extended with custom vehicles, tracks, and plugins. Furthermore, the simulator exposes an interface in Python to retrieve in real-time detailed data related to the running vehicles such as position, velocities, accelerations, tires, and aerodynamics. 

We developed additional interfaces to send the actuation commands, as well as created the ROS2 wrappers to use the same messages of the real system.
We started with custom mods of the Dallara IL-15 and the oval tracks available online. The car model has been adjusted to replicate the engine map, setup, and tire model of the Dallara AV-21. Our Motion Planning and Control algorithms have been heavily tested in this simulation environment in which it is possible to easily produce challenging scenarios changing several parameters such as the road friction coefficient, car setup and stability, wind, and slipstream effects, as well as running against multiple AIs or human-driven agents. A Windows machine is dedicated to running the racing game and publishing the ROS2 messages whereas a separate machine with Linux runs a version of the \texttt{er.autopilot 1.0} software disabling some nodes related to the Perception, and adapting the parameters related to Race Control and other communications that are not replicated on the simulator.

Further details on the interfaces, customization and potential contribution of this simulation platform will be presented in a separate work.

\subsection{Unity-based semi-HiL Simulator}
\label{unity-sim}

Besides AssettoCorsa, we decided to implement a lightweight Unity-based simulator to test all the software stack onto, with the same interface as the real car, easy to install and use. 

It is semi Hardware In the Loop (HIL) approach, given that the communication is the same as the car at the lowest level possible, in particular: 
\begin{itemize}
    \item The Raptor and MyLaps communicate via a virtual CAN interface as the real car;
    \item The GPS is simulated and sent via TCP, using messages formatted as for the real Novatel GNSS modules;
    \item The LiDAR is simulated and sent via UDP, using messages formatted as the real Luminar. 
\end{itemize}
Moreover, the race track in the simulator is georeferenced as well, therefore the GPS positions coincide with reality.

The car dynamics are provided by the NWH Vehicle Physics 2\footnote{http://nwhvehiclephysics.com}, and then they have been tuned to match as much as possible our vehicle settings. Despite the ease of adjusting the vehicle model, it has not been possible to reach fidelity on the lateral dynamics as accurately as on AssettoCorsa or Dymola. 

The simulator was designed to be used on the real hardware in a Hardware In the Loop (HIL) fashion, through CAN and Ethernet connections. Nonetheless, it can run in a Software In The Loop (SIL) fashion on any high-end laptop like those used by the team for development. In particular, this simulator has been used by the developers to implement and validate the correctness of modules related to the system integration such as Mission Planner (\autoref{sec:mission_planner}), Controller Mux (\autoref{sec:controller-mux}), Supervisor and Safety (\autoref{sec:supervisor}).
Finally, there is also the possibility to run the simulator automatically, with a predefined mission, and headless, without visualization. We exploited this feature to include a simulation test in our GitLab pipelines.  

\section{Telemetry and Visualization}
\label{sec:telemetry}
\begin{figure}[htb!]
  \centering
  \includegraphics[
    width=0.60\textwidth,
    trim={4cm 22cm 3cm 1.5cm},
    clip
  ]{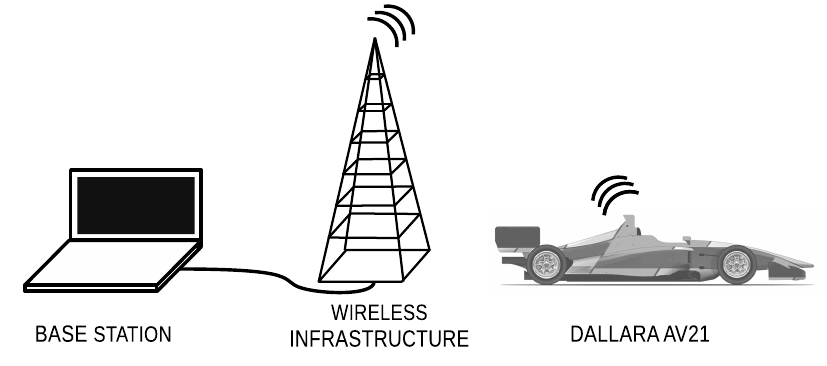}
  \caption{The Dallara AV-21 communicates with the base station via a wireless infrastructure made up of multiple antennas placed around the track. The base station is connected to the infrastructure using an Ethernet cable.}
  \label{fig:infrastructure}
\end{figure}

The Dallara AV21 car constantly communicates with an on-ground computer called base station. The base station and the car communicate via a wireless infrastructure that is made up of multiple antennas placed around the track (\figref{fig:infrastructure}).

From the base station, it is possible to send commands to the car and monitor all the signals that are relevant to evaluate the performance of the car. From a joystick connected to the base station, it is possible to reduce the speed of the car and command an emergency stop.
If communication with the base station is lost, the car performs a graceful stop.

\subsection{Telemetry Data}

To overcome the issue of the limited bandwidth of the wireless infrastructure, the TII Euroracing team has implemented a proprietary protocol based on UDP that drastically reduced the amount of data going through the infrastructure.
All the signals coming from the car are downsampled to 5Hz and compressed before sending them to the base station.

\subsection{Visualization}

\begin{figure}[htb!]
  \centering
  \fbox{
  \includegraphics[
    width=0.75\textwidth
  ]{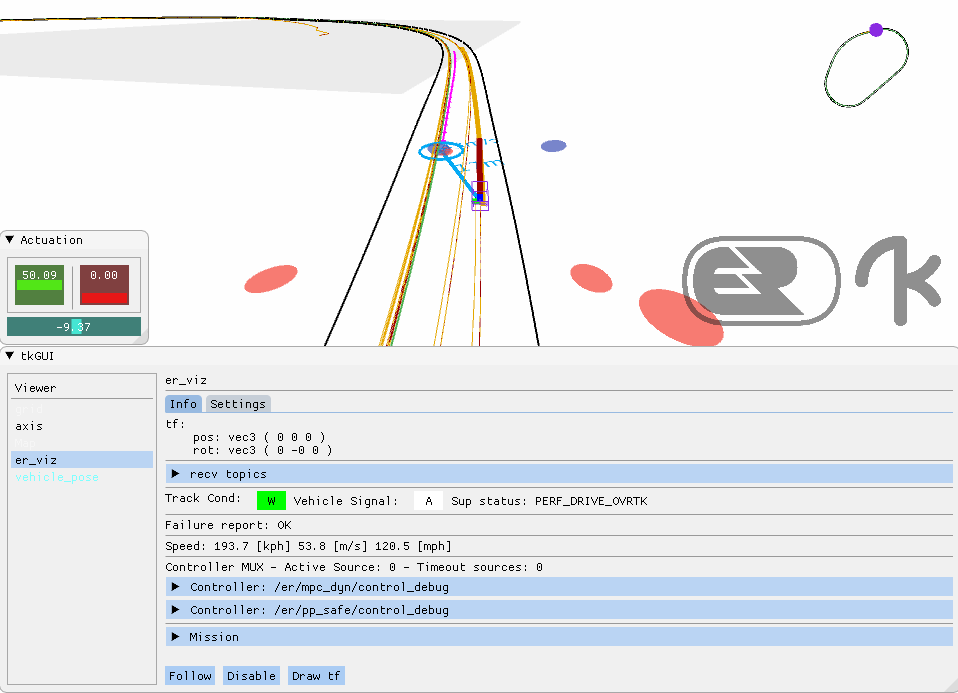}}
  \caption{\texttt{er.viz} is the visualization tool used on the base station to monitor the performance of the car. The car position is shown in the centre of the window; the detected car is marked with a blue circle; the desired trajectory is drawn with a thick yellow line, while the motion forecasting of the detected car is drawn in purple.}
  \label{fig:er_viz}
\end{figure}

While the car is running, on the base station the proprietary software er.viz (\figref{fig:er_viz}) allows the team to visualize the car position, the planned trajectory, and the detected obstacles, along with the car speed.  

Along with er.viz, on the base station the open-source software \emph{PlotJuggler}\footnote{PlotJuggler: https://www.plotjuggler.io/} is used to plot the signals in real-time. The most relevant signals monitored by the team during a run are the lateral error from the desired trajectory, the steering and throttle commands, the tires' temperatures, and the covariances of the localization.
\section{Results}
\label{sec:results}
In this section, we report the main results for each of the presented software modules.
\subsection{Localization}
\begin{figure}[htb!]
    \centering
    \includegraphics[
    width=\textwidth,
    trim={0cm 7cm 0cm 8cm},
    clip
    ]{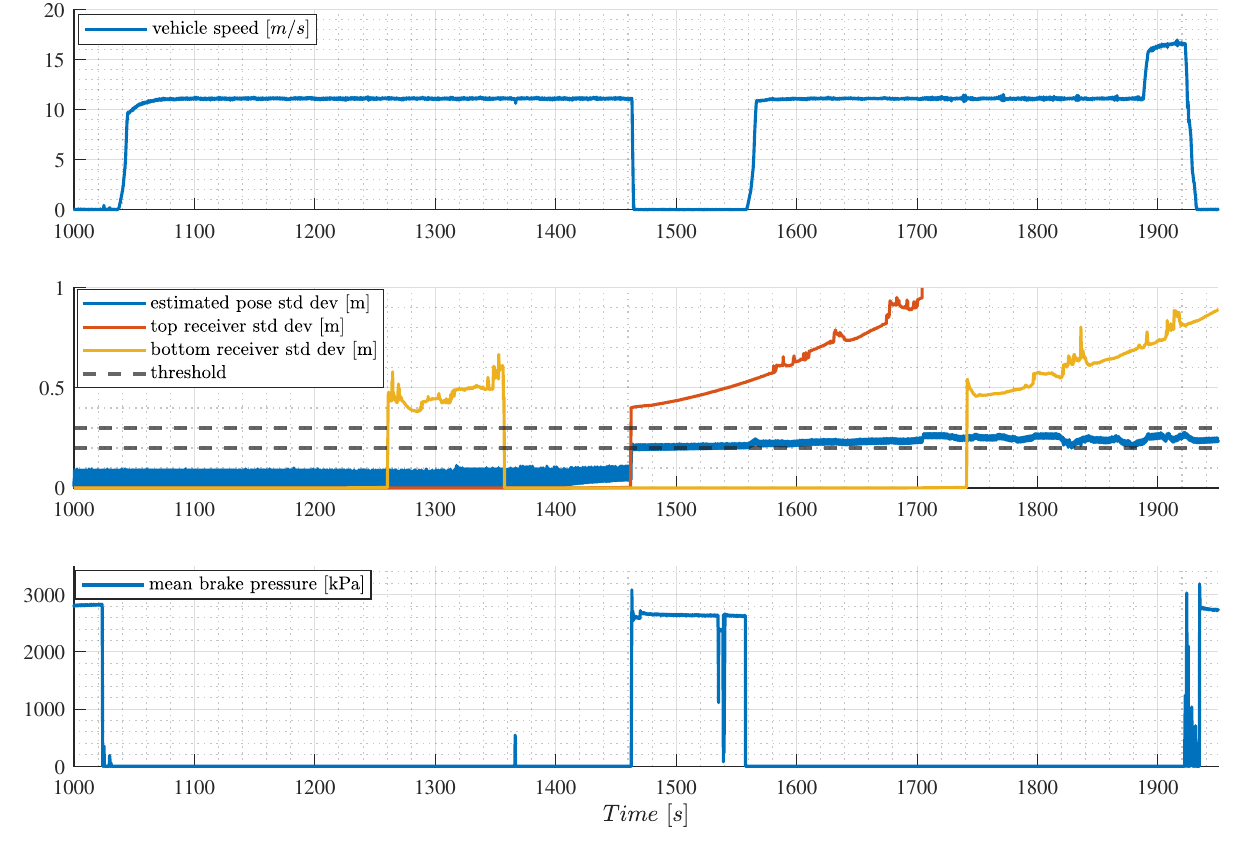}
    \caption{Ego-vehicle localization in absence of RTK correction. i) In the first test the RTK correction is disabled for the bottom receiver around $1250 s$. Given the tuning used in the test, this does not affect the accuracy of the estimation. The correction is then reapplied. At around time $1450 s$ the top receiver RTK correction is disabled, which immediately brings the pose estimate standard deviation (average over latitude and longitude) over the first threshold of $0.2 m$ triggering a safe braking of the car. ii) The safety threshold is increased to $0.3 m$ and the test continues. It is visible how the top receiver accuracy quickly degrades. Once also the bottom receiver RTK correction is disabled again at around $1750 s$, the accuracy is still within range and the car keeps moving. The car is then stopped manually and the test is considered successful.}
    \label{fig:RTK_drops}
\end{figure}

In evaluating the quality of the localization system, we could not rely on a ground truth system for comparison. We proceeded having in mind the following objectives:
\begin{enumerate}[noitemsep]
    \item Empirically compare the ego-vehicle estimate with the GNSS raw inputs, also considering their covariance.
    \item Evaluate the performance of the estimator in case the RTK signal was lost on one or both receivers.
    \item Understand the practical performance of the localization system and derive safety thresholds based on its accuracy and tolerance to sensor malfunctions.
\end{enumerate}
In \autoref{fig:RTK_drops} we show the result of some of these tests conducted at LOR, where each sensor RTK correction is disabled/enabled in different combinations.

The importance of these tests lies in the fact that they gave us confidence in the accuracy of the car pose estimates, which could then be used to calibrate the safety thresholds for the safe-stop signals in the failure detection module and to confidently increase the operating speed during the test sessions.
The LiDAR localization system described in \autoref{sec:lidar_loc} is treated as an additional, virtual, sensor in the EKF, providing position, heading, and velocity estimates with their confidence at around $25$Hz. Given the limited amount of track time we had to test this critical software component, we decided not to use it in the final races, to advance its development and present more results in future works.

\begin{figure}[ht!]
    \begin{subfigure}[b]{0.45\textwidth}
        \includegraphics[width=0.95\textwidth]{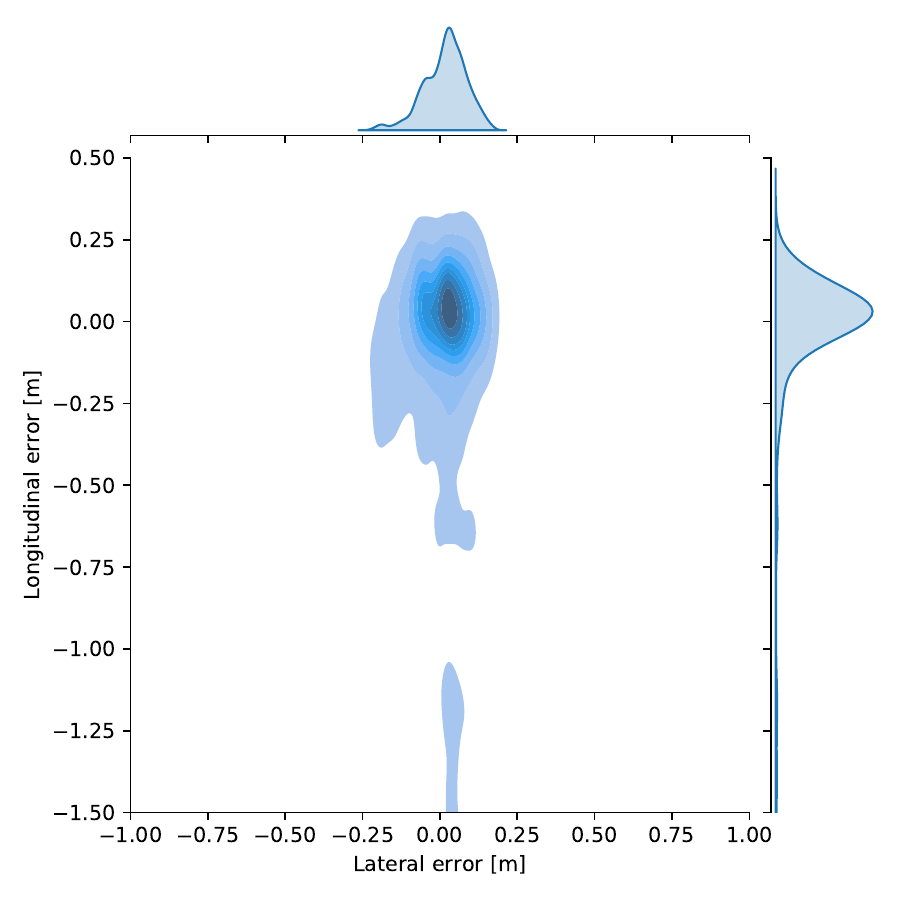}
    \end{subfigure}
    \hfill
    \begin{subfigure}[b]{0.55\textwidth} 
        \includegraphics[width=\textwidth]{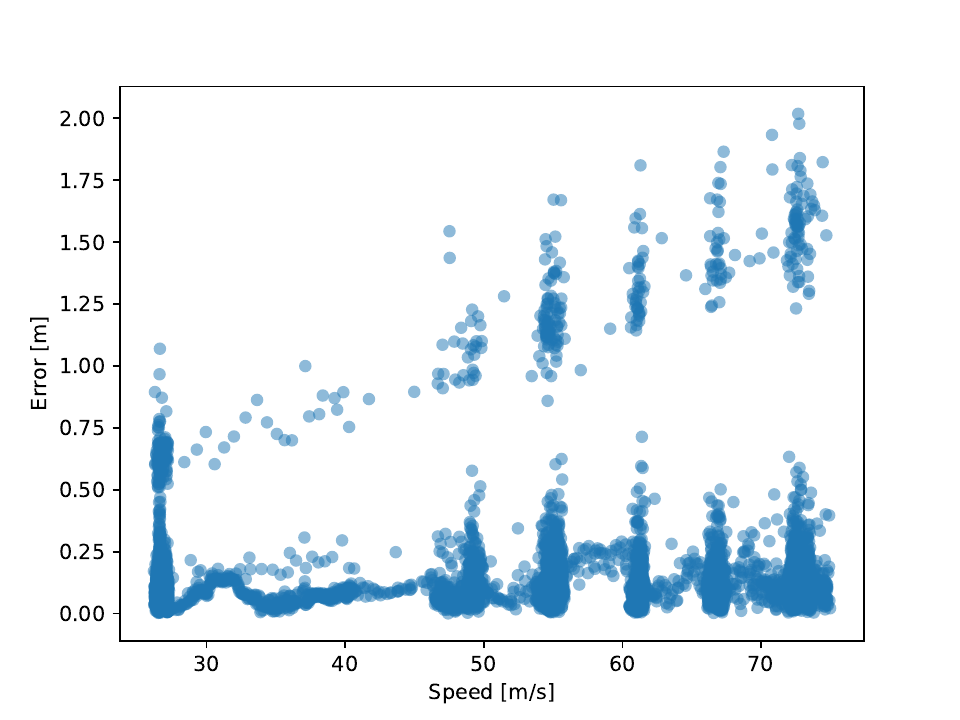}
    \end{subfigure}
    \caption{Results of the LiDAR localization at LVMS. On the left, is the jointplot of the lateral error (x-axis) and the longitudinal one (y-axis). On the right is the plot of the error module (y-axis) at different speeds. }
    \label{fig:lid_loc}
\end{figure}

However, we report our current results in the charts of \Cref{fig:lid_loc}. On the left, it is depicted a jointplot with the distribution of the lateral and longitudinal error in meters. The error has been computed considering the RTK-corrected GPS position as ground truth, on the qualification log for the LVMS race where the car exits the pit, increments the speed up to $75$ m/s, and goes back to the pit. The origin in the plot represents the car's Centre of Mass.
The results are very promising: the maximum lateral error is $~25$ cm, while the longitudinal one is on average around $30$ cm, with few peaks close to $2$ m. 

The chart on the right shows the error in m (y-axis) at the different speeds in m/s (x-axis). From this second plot, we can notice that the error peaks increase while increasing the speed, as one could expect. We can then notice that the $2m$ errors are experienced only over $70$ m/s, while at lower speeds the maximum error is always lower. Moreover, the chart shows two distinct patterns of errors: frequent low errors at the bottom of the chart, and higher errors in the upper part. These patterns appear to be correlated with straight sections and curved exits, respectively.

\subsection{Object Detection and Tracking}
To correctly evaluate the perception pipeline, we have to consider both their execution time and their accuracy.

Given that there is not an official dataset to evaluate our methods on, we will report the accuracy results on the data we have manually labeled. For the same reason, we sometimes report both boxplot graphs and tables to serve as reference in future works.

For the Object Detection NN of the camera detection pipeline (\Cref{perpipe:camera_det}) the accuracy results have been reported in \Cref{tab:det_map}. The dataset statistics reported refer to the manually labeled images used to finetune the YOLOv4 pretrained on BDD100K. The Average Precision (AP) obtained for open-wheel racecars of our solution is 92\% when using a confidence threshold of $0.5$. 

\begin{table}[!ht]
    \centering
    \begin{tabular}{|c|c|c|c|c|c|}
    \hline
    Train-set Size & Val-set Size    &  Classes & AP 0.5 & AP 0.75    & AP 0.5:0.95 \\ \hline
    350 & 50              & 1       & 0.92    & 0.67        & 0.60          \\ \hline 
    \end{tabular}
    \caption{Accuracy results of the Object Detection of the camera detection pipeline (\Cref{perpipe:camera_det}).}
    \label{tab:det_map}
\end{table}

\begin{figure}[!ht]
    \centering
            \includegraphics[clip, trim=0cm 0.5cm 0.5cm 0.5cm, width=0.7\textwidth]{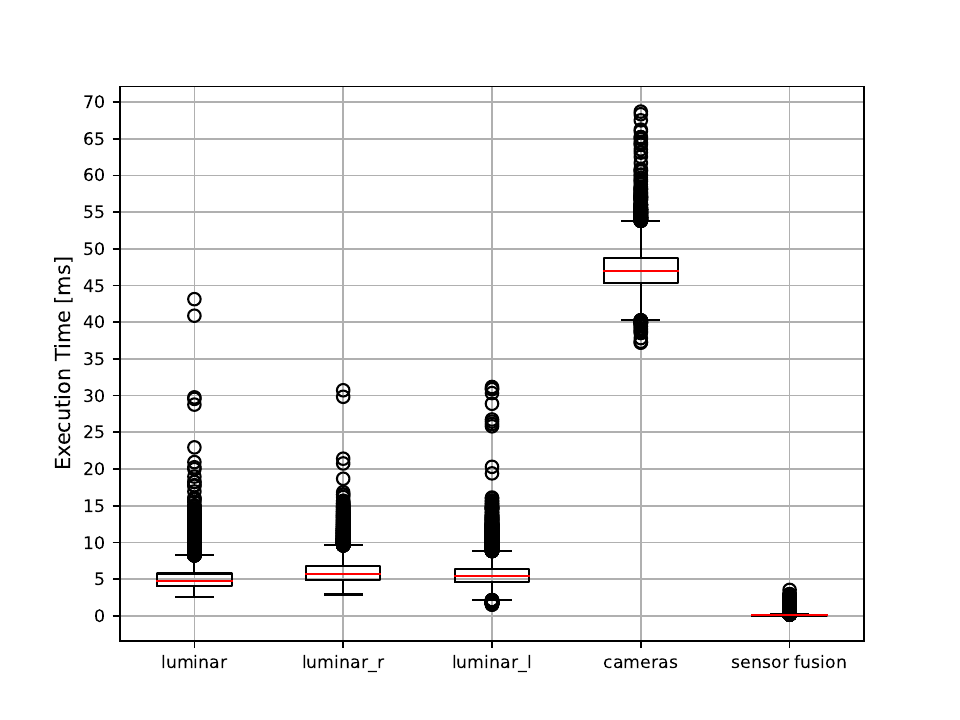}

    \caption{Boxplots of the execution times of the LiDAR clustering BEV(\Cref{perpipe:lidar_bev}), camera detection (\Cref{perpipe:camera_det}) and sensor fusion (\Cref{perpipe:sensor_fusion}) pipelines, over 16600 iterations. }
    \label{fig:perception_times}
\end{figure}

Regarding the execution time, \Cref{fig:perception_times} shows the boxplots of the LiDAR clustering BEV (\Cref{perpipe:lidar_bev}), camera detection (\Cref{perpipe:camera_det}) and sensor fusion (\Cref{perpipe:sensor_fusion}) for $16600$ iteration of one complete log run where these pipelines where active. The execution times refer to the vehicle's workstation. The three Luminar LiDAR are handled by three separated nodes, whose boxplots are \texttt{luminar}, \texttt{luminar\_l} and \texttt{luminar\_r}, while all the cameras are elaborated by a single node to have a single batched inference (therefore there is only one boxplot). 
Please keep in mind that the execution times provided were calculated while the computer was operating at full capacity, running the entire stack, rather than in isolation. As a result, the execution times for the same algorithm (such as \texttt{luminar}, \texttt{luminar\_l}, and \texttt{luminar\_r}) may fluctuate slightly.

Moreover, these pipelines' statistics are also reported in \Cref{tab:perception_times}. On average the BEV clustering with its internal tracking is performed in $5.5ms$, the vehicle detection and tracking on 6 camera frames in $47 ms$, and all the data are fused in $0.14ms$. 
Unfortunately, we have not recorded accurate profile data for the other pipelines. 

\begin{table}[!ht]
    \centering
    \begin{tabular}{|c|c|c|c|}
        \hline
         & \textbf{avg [ms]} & \textbf{max [ms]}  & \textbf{min [ms]} \\ \hline
         luminar          &  5.08 & 43.14 &  2.51 \\ \hline
         luminar\_r        &  5.93 & 30.73 &  2.91 \\ \hline
         luminar\_l        &  5.66 & 31.18 &  1.53 \\ \hline
         cameras          & 47.05 & 68.69 & 37.21 \\ \hline
         sensor fusion    &  0.14 &  3.54 &  0.04 \\
         \hline
    \end{tabular}
    \caption{Execution time statistics of the LiDAR clustering BEV(\Cref{perpipe:lidar_bev}), camera detection (\Cref{perpipe:camera_det}) and sensor fusion (\Cref{perpipe:sensor_fusion}) pipelines, over 16600 iterations.}
    \label{tab:perception_times}
\end{table}

Finally, we evaluated the result of the Sensor Fusion pipeline (\Cref{perpipe:camera_det}), which is the module in charge to give the input to the Planner, when merging detections from the LiDAR clustering BEV (\Cref{perpipe:lidar_bev}) and the Radar (\Cref{perpipe:radar_det}) pipelines on the head-to-head run against TUM, during the LVMS event. 

To compute that, we compared the GPS position of TUM's car, using the output of the Novatel top (in front of the car), and the GPS position of our vehicle, using the output of the Novatel top (in front of the car), and the detection computed by the Sensor Fusion module, that is the centre of the detected objects, aligning our data with the oppenents' ones with the Novatel's timestamp. We then converted the GPS positions into local coordinates, to compare them with the detections, and we only considered the case in which the opponent's car is not behind ours. 

It is important to notice that the considered ground truth (TUM's position) is a point in the front of the car, while the detection is usually a point in the back of the car, and that the car is almost 5 meters long and 2 meters wide.  
Keeping that in mind, to compute the goodness of the pipeline, we considered various scenarios in which the two cars' euclidean distance was in a certain range. 


This having been said, we evaluated the ranges $[0-10]$m, $[10-25]$m, $[25-50]$m, $[50-100]$m, $[100-150]$m, and $[0-150]$m.
\Cref{tab:fusion_res} reports the complete evaluation for those ranges, including True Positives (TP), False Negatives (FN), False Positives (FP), precision (p), recall (r), and longitudinal and lateral errors statistics, such as minimum (min), maximum (max), average (avg), and median (med).
It is worth saying that the recall is $1.0$ within $50$ meters, and the worse value is $0.65$ in the $[100-150]$m range, on the other hand, the precision is around $1$ when the distance is greater than $50$m. For smaller distances, the precision is worse due to the error of the detection, e.g. considering the range $[0-10]$m if the ground truth position is around 11.0 m (out of the considered range), while its detection is around $9$m and that is considered FP. 

\begin{table}[!ht]
    \centering
    \resizebox{\textwidth}{!}{
    \begin{tabular}{|c|c|c|c|c|c|c|c|c|c|c|c|c|c|c|}

        \hline
        \textbf{Min dist} &	\textbf{Max dist} & \textbf{TP}	    & \textbf{FN}	    &\textbf{FP}	    & \textbf{p} & \textbf{r} & \textbf{$min_x$} & \textbf{$max_x$} & \textbf{$avg_x$} & \textbf{$med_x$}	& \textbf{$min_y$} & \textbf{$max_y$}	& \textbf{$avg_y$} & \textbf{$med_y$} \\ \hline
        0.00	& 10.00  &	141.00	    & 0.00	    & 57.00	& 0.71	& 1.00 & 	0.48	& 7.13	    & -2.71	    & -2.86	& 0.05	& 2.10	& -1.06	& -1.10 \\
        10.00	& 25.00  &	298.00	    & 0.00	    & 69.00	& 0.81	& 1.00 & 	1.01	& 5.39	    & -3.40	    & -3.65	& 0.04	& 3.85	& -1.48	& -1.38 \\
        25.00	& 50.00  &	609.00	    & 0.00	    & 80.00	& 0.88	& 1.00 & 	0.71	& 5.74	    & -1.85	    & -2.42	& 0.09	& 5.30	& -0.88	& -1.18 \\
        50.00	& 100.00 &	2,814.00	& 101.00	& 34.00	& 0.99	& 0.97 & 	0.00	& 11.30	    & 0.54	    & 0.50	& 0.00	& 7.80	& 0.72	& 1.85  \\
        100.00	& 150.00 &	598.00	    & 327.00	& 0.00	& 1.00	& 0.65 & 	0.01	& 98.28	    & 10.40	    & 5.36	& 0.12	& 97.20	& 4.40	& -2.26 \\
        0.00	& 150.00 &	4,460.00	& 428.00	& 0.00	& 1.00	& 0.91 & 	0.00	& 98.28	    & 0.41	    & -0.66	& 0.00	& 97.20	& 0.49	& -0.10 \\ 
        \hline
    \end{tabular}
    }
    \caption{Results of the Sensor Fusion (\Cref{perpipe:sensor_fusion}) output (centre of detection) compared with the TUM position (Novatel top). The distance between TUM's car and ours is in the ranges [Min dist-Max dist] m, and all the error statistics (minimum, maximum, average, and median) are in meters.}
    \label{tab:fusion_res}
\end{table}

\begin{figure}[ht!]
    \begin{subfigure}[b]{0.24\textwidth}
        \includegraphics[clip, trim=0cm 0cm 1cm 2cm,width=\textwidth]{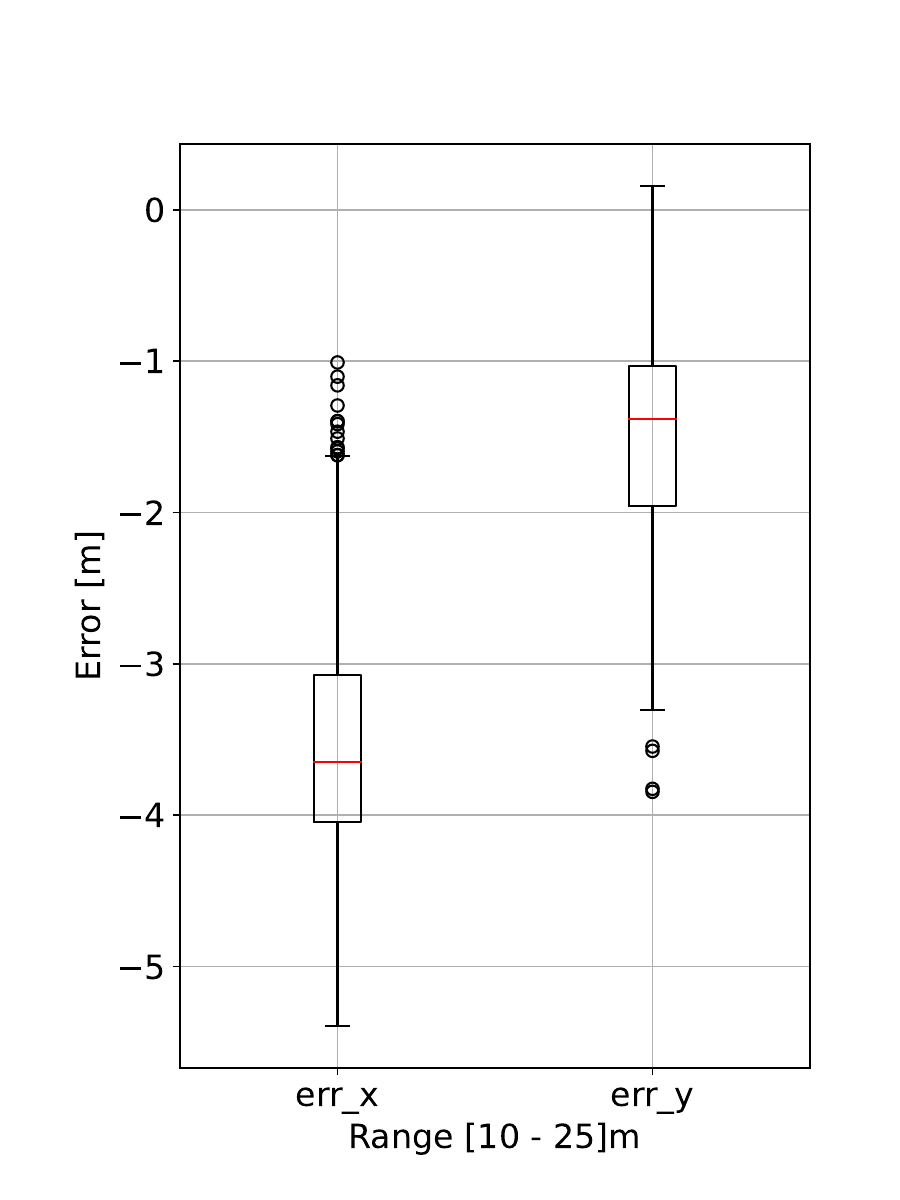}
    \end{subfigure}
    \hfill
    \begin{subfigure}[b]{0.24\textwidth} 
        \includegraphics[clip, trim=0cm 0cm 1cm 2cm,width=\textwidth]{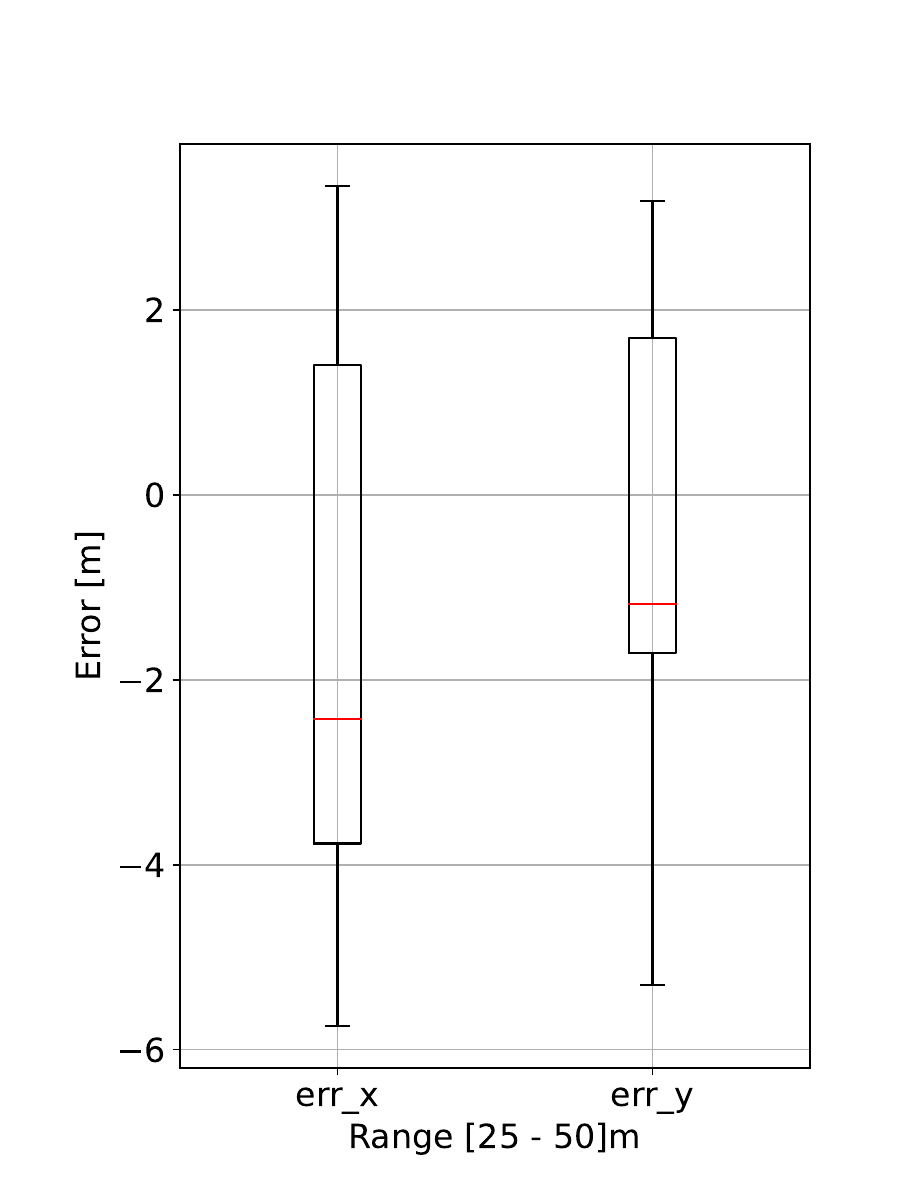}
    \end{subfigure}
    \hfill
    \begin{subfigure}[b]{0.24\textwidth} 
        \includegraphics[clip, trim=0cm 0cm 1cm 2cm,width=\textwidth]{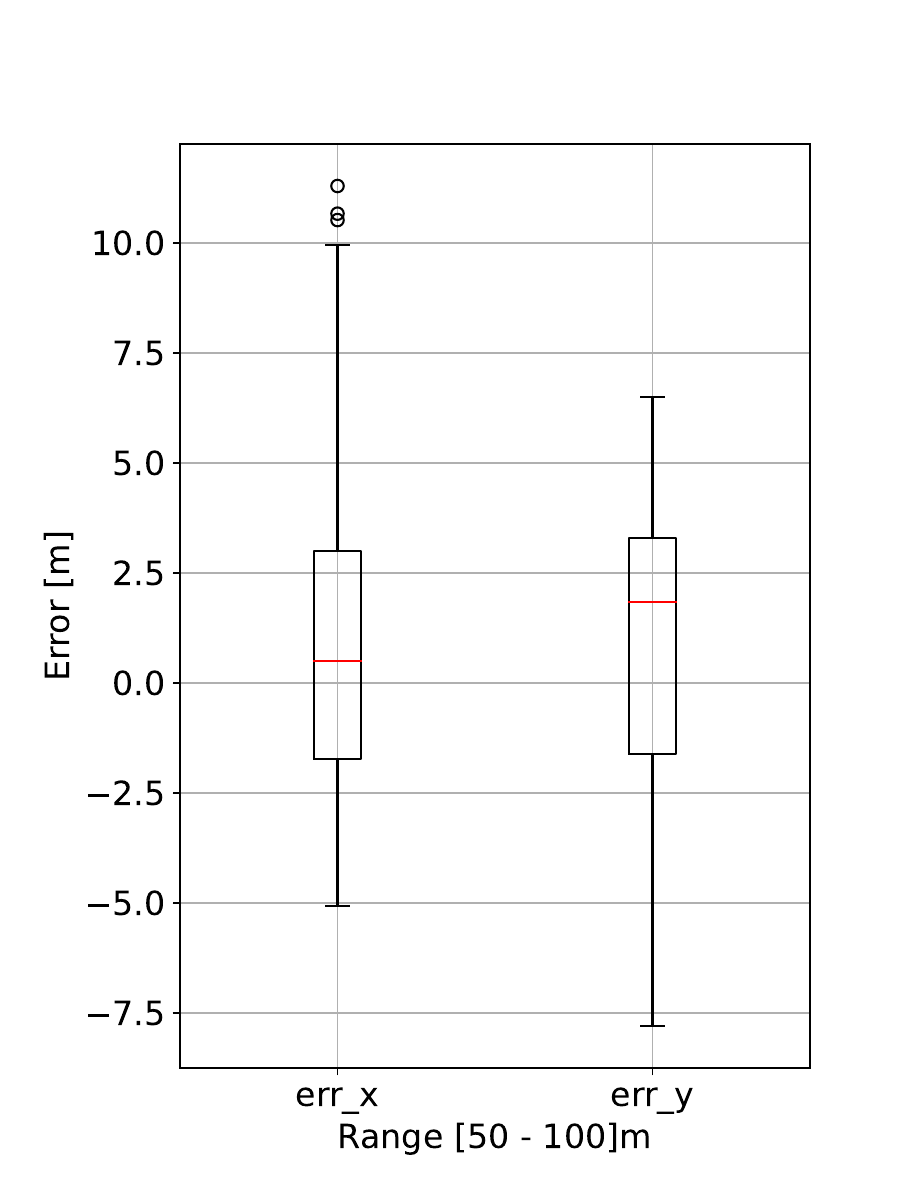}
    \end{subfigure}
    \hfill
    \begin{subfigure}[b]{0.24\textwidth} 
        \includegraphics[clip, trim=0cm 0cm 1cm 2cm,width=\textwidth]{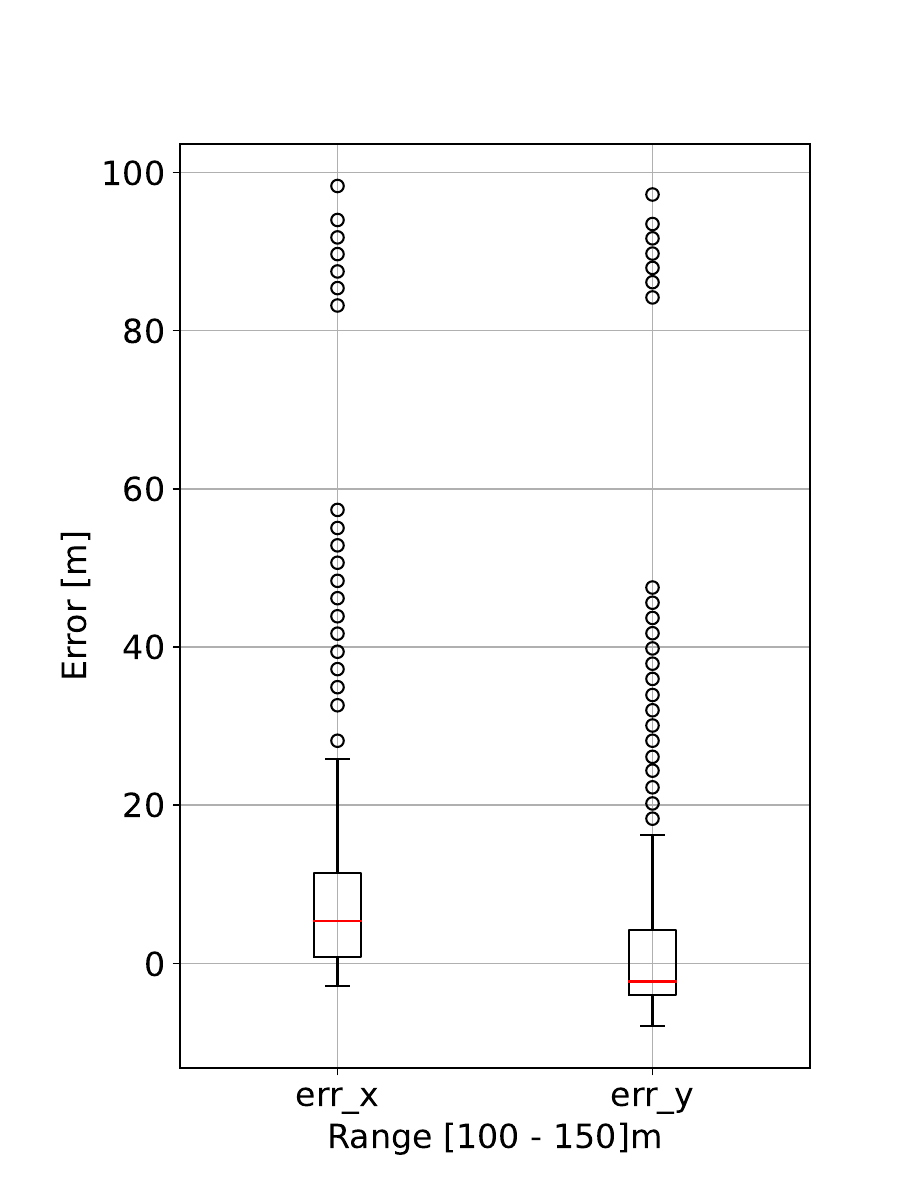}
    \end{subfigure}
    \caption{Statistics of the longitudinal and lateral error of the Sensor Fusion's output (\Cref{perpipe:sensor_fusion}) compared with the TUM position (Novatel top). The distance between TUM's car and ours is, from left to right, in the ranges $[0-25]$m, $[25-50]$m, $[50-100]$m, and $[100-150]$m.}
    \label{fig:fus_plot}
\end{figure}

Besides the table, the lateral and longitudinal errors of the detections are depicted in \Cref{fig:fus_plot}, in particular for the ranges $[10-25]$m, $[25-50]$m, $[50-100]$m, and $[100-150]$m. From the range $[10-25]$m, the average longitudinal error is $-3.40$m, which is almost the distance between the top position (ground truth) and the rear one (centre of detection seen from behind). More in general, the maximum longitudinal error is around $10.4$m and the lateral is around $4.4$m (without considering the actual size of the cars) within $100$m of distance. It has greater outliers, up to $98.28$m of longitudinal error and $97.2$m of lateral error when the distance between the two vehicles is over 100m; the cause can be found in the only presence of radar detection, less reliable than LiDAR, at that range and the divergence of the Kalman Filter on the banking.     

\subsubsection{Motion Forecasting}
A complete analysis of the performance of the motion forecasting module would not have
been possible without the help of the TUM team, which provided us with the GPS log
of their car. The GPS position is extremely accurate thanks to the RTK correction, so
it has been used as the ground truth to evaluate the output of the motion forecasting
module. Without using the log of another car, the only way to analyze the performance
would have been the usage of the perception module output as ground truth. However,
the estimation error of the motion forecasting module would have been affected by the
error of the perception module.

\begin{figure}[htb!]
  \centering
  \includegraphics[
    width=0.98\textwidth,
    trim={1cm 11cm 1cm 11cm},
    clip
  ]{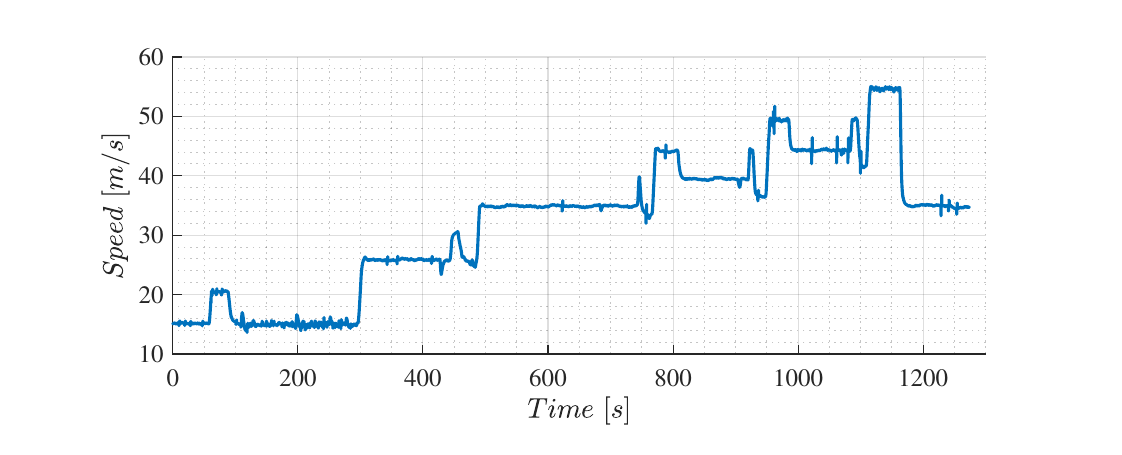}
  \caption{Speed profile in the log provided by the TUM team. The dataset contains multiple laps at different speeds.}
  \label{fig:result:tum_speed}
\end{figure}

In the dataset used, the TUM car did several laps at different speeds, ranging from 10 m/s up to 55 m/s (Figure\;\ref{fig:result:tum_speed}).
The position of the car at each step has been used directly as input to the motion forecasting module.

\begin{figure}[htb!]
  \centering
  \includegraphics[
    width=0.98\textwidth,
    trim={1cm 9cm 1cm 9cm},
    clip
  ]{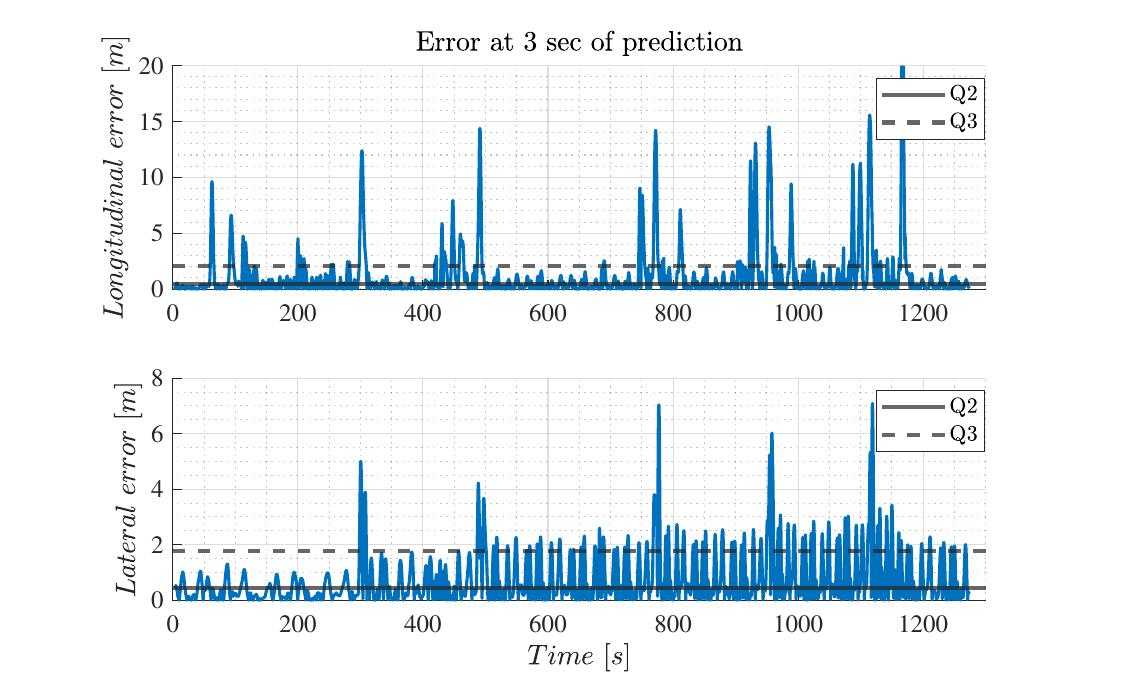}
  \caption{On the top graph, the longitudinal error of the motion forecasting is illustrated at 3 seconds of prediction, whereas the bottom one contains the lateral error. The output of the motion forecasting module is 3 seconds long. For convenience, the statistical quantities are also reported in Table\;\ref{table:result:motion_forecasting_stats}.}
  \label{fig:result:motion_forecasting_3sec}
\end{figure}

The predicted trajectory of the motion forecasting module has a length of 3 seconds\footnotemark, so, to evaluate the accuracy of the prediction, the prediction error is computed at different prediction lengths. Moreover, the error is split along the longitudinal and lateral axes.
\footnotetext{Length in time: the last point of the predicted trajectory is the predicted position of the car in 3 seconds.}
In \figref{fig:result:motion_forecasting_3sec}, the error at 3 seconds of prediction is shown, along with the median (Q2) and the 75th percentile (Q3).
Table\;\ref{table:result:motion_forecasting_stats}  summarizes the statistical characteristic of the error at 1 second, 2 seconds, and 3 seconds of prediction.

\begin{table}[htb!]
  \centering
  \begin{tabular}[htb!]{|c|ccc|}
    \hline
    \multicolumn{4}{| c |}{\textbf{Longitudinal Err.} (m)}\\
    \hline
         & \textbf{1 sec} & \textbf{2 sec} & \textbf{3 sec}
    \begin{filecontents*}{results/motion_forecasting/error_s_stats_print_bis.csv}
,1sec,2sec,3sec
min,0.0,0.0,0.0
Q1,0.04,0.09,0.16
Q2,0.107,0.26,0.49
Q3,0.27,0.71,1.34
max,7.91,19.75,34.19
\end{filecontents*}
\csvreader[]{results/motion_forecasting/error_s_stats_print_bis.csv}{1=\stat,2=\one,3=\two,4=\three}{
      \\
      \hline
      \stat & \one & \two & \three
    }
    \\
    \hline
  \end{tabular}
  \begin{tabular}[htb!]{|c|ccc|}
    \hline
    \multicolumn{4}{| c |}{\textbf{Lateral Err.} (m)}\\
    \hline
         & \textbf{1 sec} & \textbf{2 sec} & \textbf{3 sec}
    \begin{filecontents*}{results/motion_forecasting/error_d_stats_print_bis.csv}
,1sec,2sec,3sec
min,0.0,0.0,0.0001
Q1,0.0659,0.1284,0.1939
Q2,0.1646,0.3136,0.454
Q3,0.4308,0.8345,1.2053
max,3.7161,6.0891,7.0894
\end{filecontents*}
\csvreader[]{results/motion_forecasting/error_d_stats_print_bis.csv}{1=\stat,2=\one,3=\two,4=\three}{
      \\
      \hline
      \stat & \one & \two & \three
    }
    \\
    \hline
  \end{tabular}

  \caption{Statistical quantities of the error of prediction at various prediction steps.}
  \label{table:result:motion_forecasting_stats}
\end{table}
By combining the speed profile (\figref{fig:result:tum_speed}) and the prediction error (\ref{fig:result:motion_forecasting_3sec}) we can see that the error increases with the speed.
Furthermore, the outliers are mostly due to the acceleration of the car.
This phenomenon is expected because the model Equation\;\eqref{eq:forecasting:model_dot_s} used to make the prediction assumes that the car is moving at a constant speed.
However the overall prediction error is promising: at 3 seconds of prediction the Q3 of the error is less than half of the car length on the longitudinal component and slightly more than half of the width of the car on the lateral component. 
This result lies in the same order of magnitude of the estimates performed relying only on the logs of our car, and helped us in tuning the safety thresholds in the planning stack.

\subsection{Planning and Control}
\label{subsec:planning_control}


Thanks to the optimization-based global planner it has been possible to generate a feasible raceline that does not take into consideration only the time minimization but also the rate of change of the steering wheel angle. Considering the uncertainties of the actuators, the model mismatch on the controller, and the need for a path feasible at all the possible velocities, we preferred to follow a smoother and safer trajectory than the potential minimum lap trajectory.

The Frenet-based planner and the Control module, have been tested thoroughly at various speeds, as presented in \cite{raji}.

\begin{figure}[htb!]
  \centering
  \includegraphics[
    width=0.9\textwidth
  ]{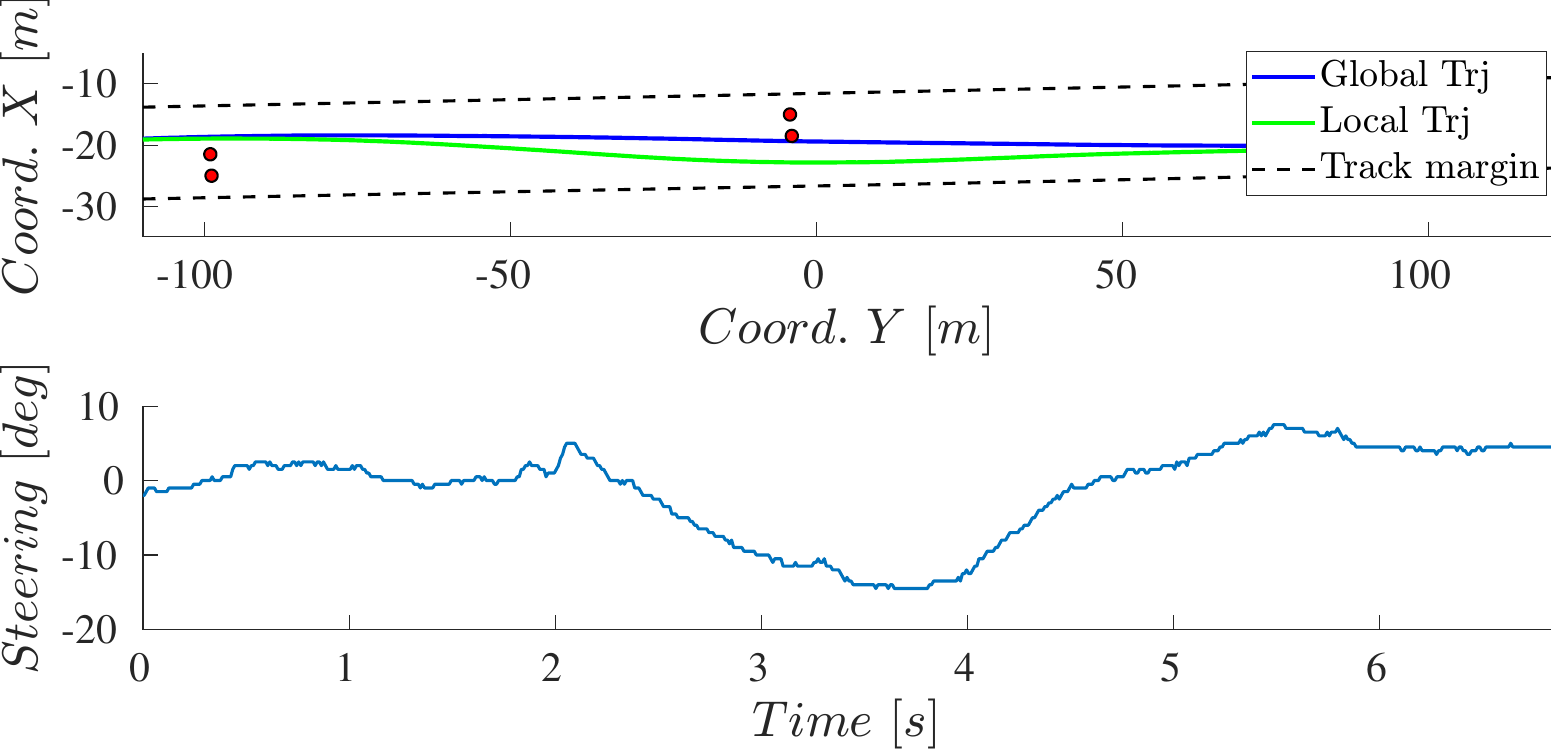}
  \caption{Avoiding pylons during the Semifinal at IMS. The location of four water-filled pylons is marked with red dots. The local planner is informed by the detection stack and reacts with a smooth trajectory to avoid them.}
  \label{fig:ims_piloni}
\end{figure}

\begin{figure}[htb!]
  \centering
  
  \begin{subfigure}[b]{0.49\textwidth}
    \includegraphics[
        width=0.99\textwidth,
    ]{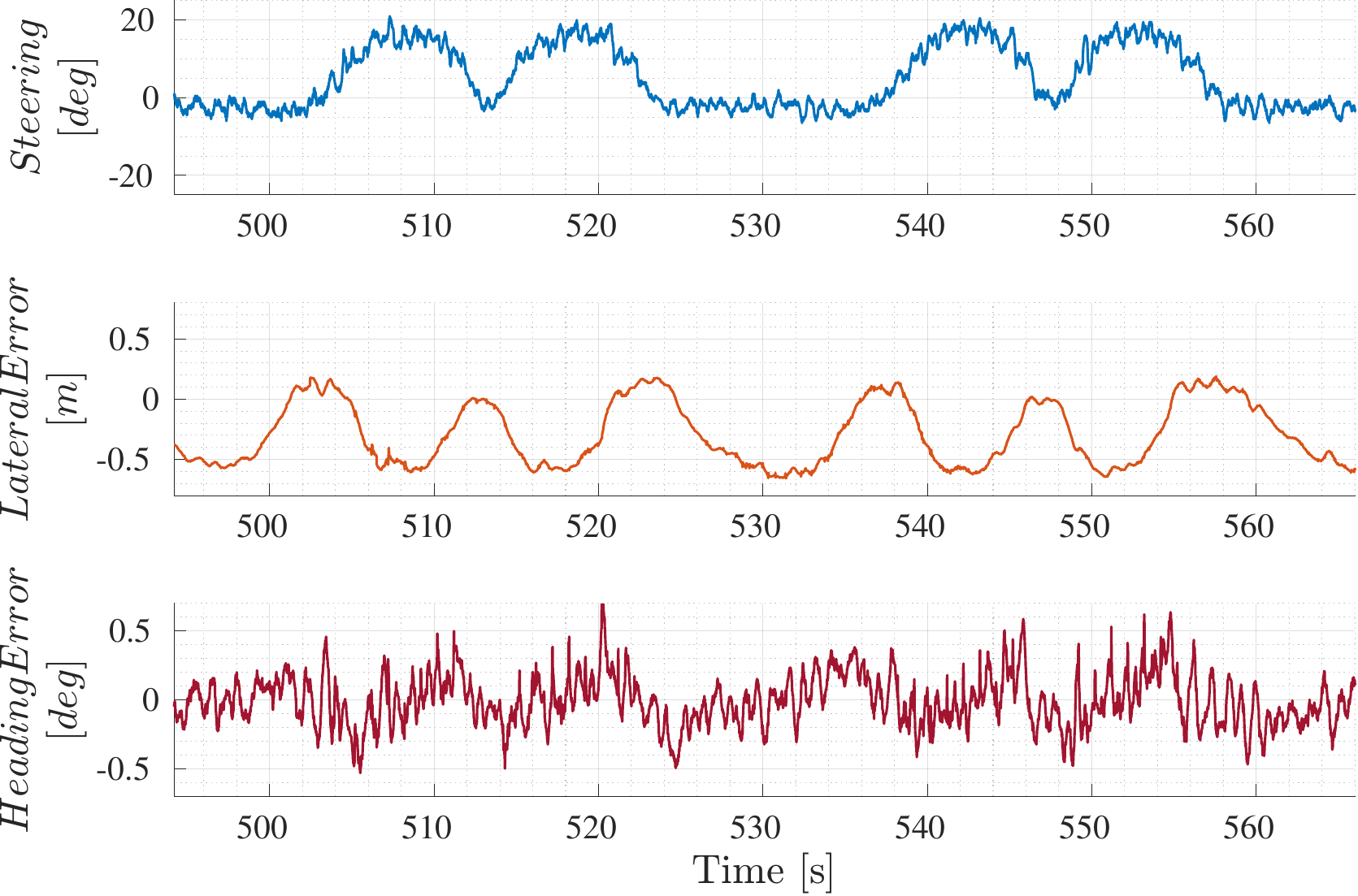}
    \caption{MPC Tracking Errors during a 215 kph lap at IMS.}
  \end{subfigure}
  \begin{subfigure}[b]{0.49\textwidth}
    \includegraphics[
        width=0.935\textwidth,
    ]{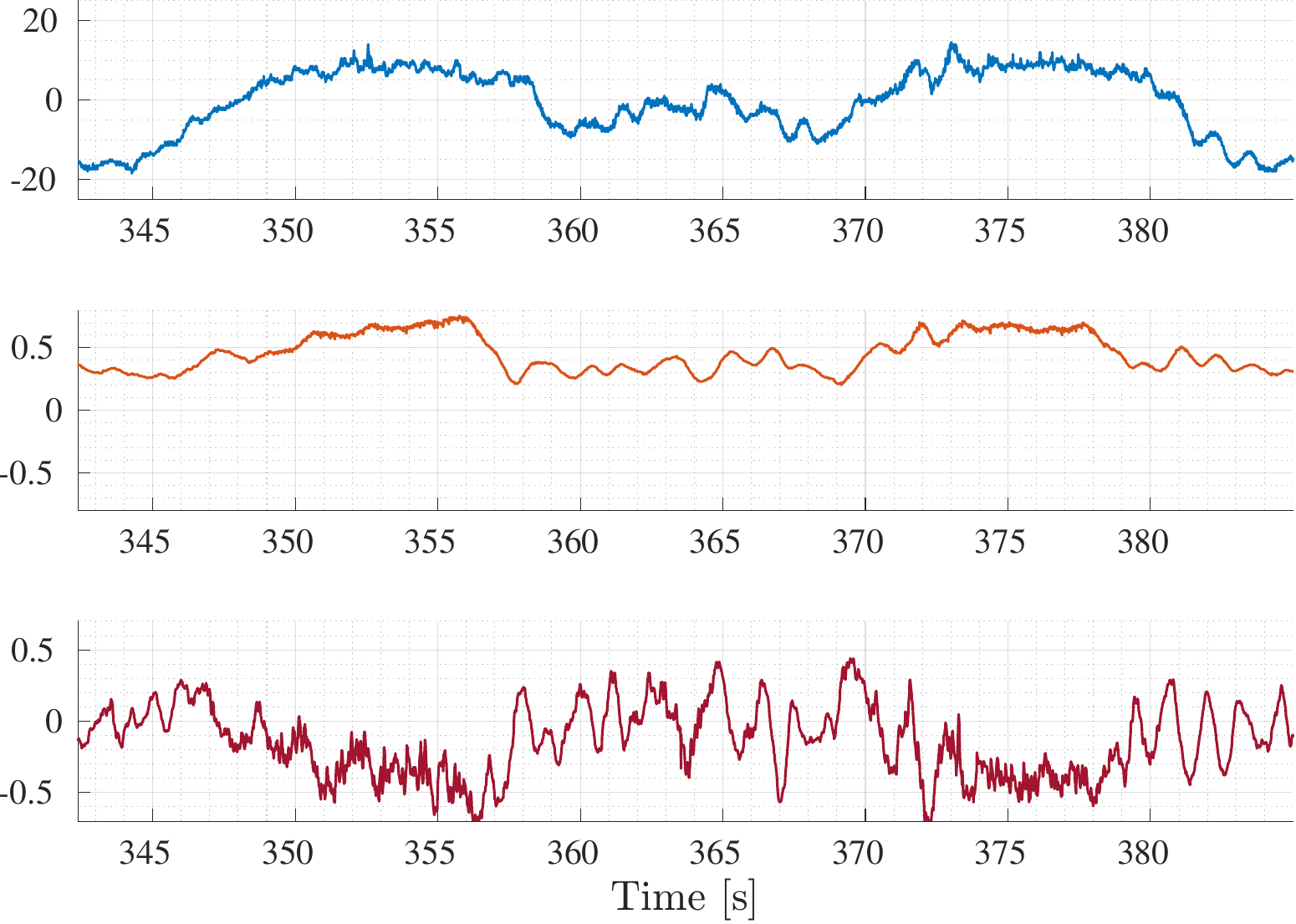}
    \caption{MPC Tracking Errors during a 220 kph lap at LVMS.}
  \end{subfigure}

  \caption{Comparing controller tracking performance at a similar speed in different circuits. Despite different tuning values being used, consistent behavior can be observed.}
  \label{fig:mpc_races_comparison}
\end{figure}

In particular, the planner has been capable of generating smooth overtaking trajectories with a speed that went from 22 m/s ($\sim$ 80 km/h) up to 64 m/s ($\sim$ 230 km/h), as well as performing safely static obstacles avoidance as can be noticed in \figref{fig:ims_piloni}. During the high-speed laps at LVMS, when the speed varied from 72 m/s to 75.5 m/s, the MPC reached a maximum lateral error of 1m and an RMS value of 0.5m, while the heading error was between 0.7deg and -1.0deg. On the other side, when the speed varied from 61.45 m/s to 63.16 m/s during the final at IMS, the maximum lateral error was -0.67m and the RMS value -0.29m, while the heading error was between 0.6deg and -0.5deg with a mean of -0.016. 
In \autoref{fig:mpc_races_comparison}, the tracking performance during the two events at similar speeds, over one lap, is shown. 
The error on the lateral tracking has been influenced by the choices on the regularization terms, and the mismatch in the tire model. 
The differences, mainly on the lateral error, are caused by changes in the weights' values in the MPC cost function we made, keeping unvaried the parameters of the Pacejka Magic Formula, from the first identification explained in \autoref{sec:control}, due to not sufficient time to safely validate on track a set of potentially more accurate coefficients.

\begin{figure}[htb!]
  \centering
  \includegraphics[
    width=\textwidth
  ]{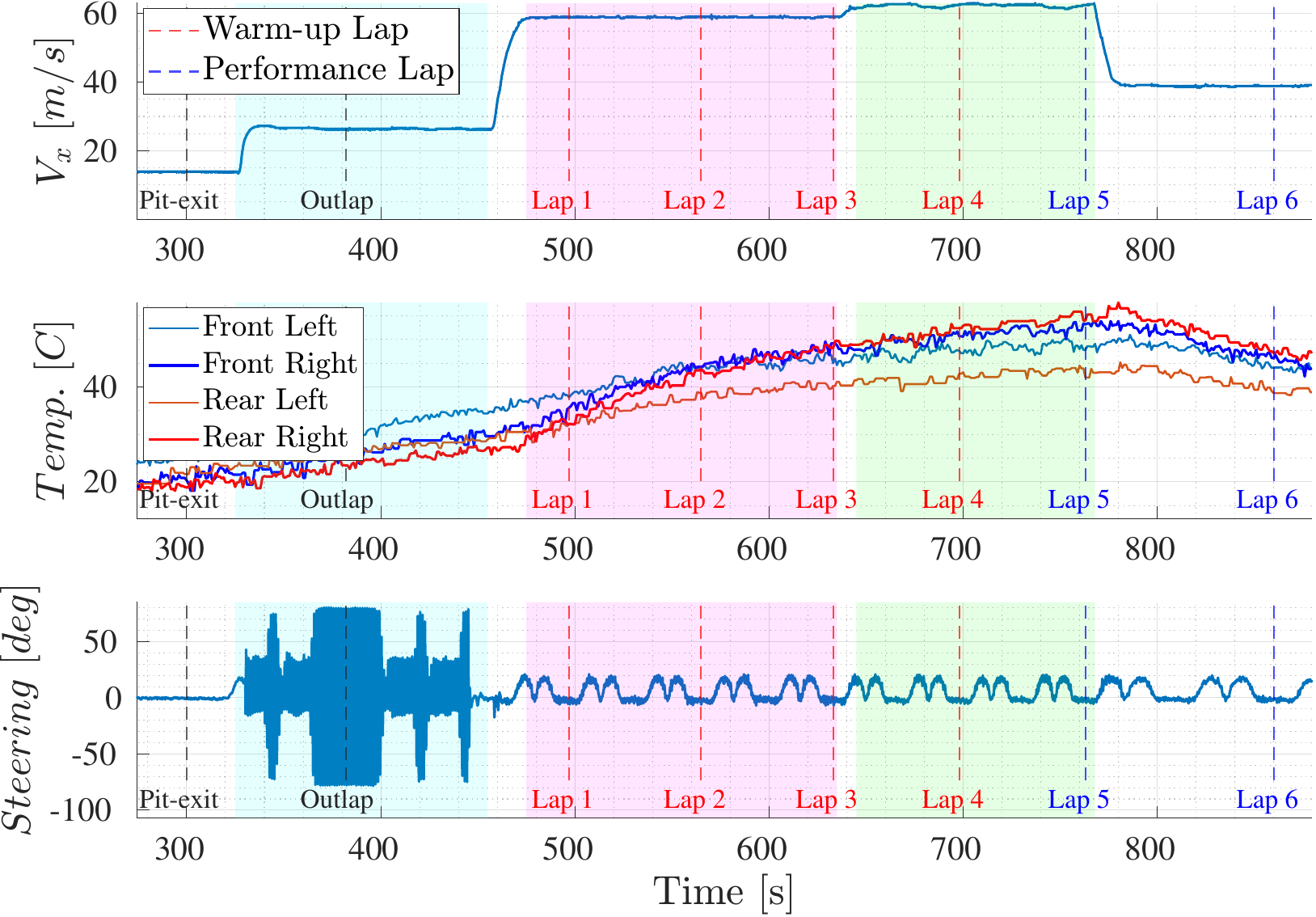}
  \caption{Tires temperatures during the final at IMS. From the pit-exit and during the outlap, the warmup manoeuvre is performed by alternating left and right steering. At higher speed, we relied purely on friction to increase the temperature. Given the low track temperature, in the few laps available for the race it was not possible to heat the tires to their nominal temperature (around $80^o$C).}
  \label{fig:tires_final_ims}
\end{figure}

The results of the Warm-up manoeuvre described in \autoref{sec:control} are presented in \autoref{table:tires_temp} and \figref{fig:tires_final_ims}, where data from the final event at IMS are used. It has been possible to increase the temperature on the front tires, before increasing the speed, as we were aiming for. As we saw in the simulations, this manoeuvre leads to a controlled front slip which can generate front temperature. It is noticeable that the rear tires' temperature increases when the car ran at 60 m/s while the temperature of the front tire began to flatten. The greater temperature on the right side is due to the banking effect and the high load transfer, which increases the vertical load on those wheels. This leads to deformation of the tire carcass which is converted into energy, therefore heat. This effect is greater on the rear right due to the car balance (both weight and aerodynamic) and the combined force.

\begin{table}[]
\centering
\begin{tabular}{c|c|ccllcll|c|}
\cline{2-10}
\textbf{}                                  &                                                                                                & \multicolumn{7}{c|}{\textbf{\begin{tabular}[c]{@{}c@{}}Temperature Gap {[}°C{]}\end{tabular}}}                                                                                                                                                                                                                                                                    &                                                                                               \\ \cline{3-9}
\multicolumn{1}{l|}{}                      & \multirow{-2}{*}{\textbf{\begin{tabular}[c]{@{}c@{}}Pit-exit\\ Temp. {[}°C{]}\end{tabular}}} & \multicolumn{1}{c|}{\cellcolor[HTML]{D3FFFD}\textbf{\begin{tabular}[c]{@{}c@{}}Warm-up\\ (1 lap)\end{tabular}}} & \multicolumn{3}{c|}{\cellcolor[HTML]{FFD7F6}\textbf{\begin{tabular}[c]{@{}c@{}}Target 60 m/s\\ (2.5 laps)\end{tabular}}} & \multicolumn{3}{c|}{\cellcolor[HTML]{D2FFD6}\textbf{\begin{tabular}[c]{@{}c@{}}Target 66 m/s\\ (2 laps)\end{tabular}}} & \multirow{-2}{*}{\textbf{\begin{tabular}[c]{@{}c@{}}Maximum\\ Temp. {[}°C{]}\end{tabular}}} \\ \hline
\multicolumn{1}{|c|}{\textbf{Front Left}}  & 25.5                                                                                           & \multicolumn{1}{c|}{\cellcolor[HTML]{D3FFFD}+ 9.9}                                                              & \multicolumn{3}{c|}{\cellcolor[HTML]{FFD7F6}+ 9.6}                                                                       & \multicolumn{3}{c|}{\cellcolor[HTML]{D2FFD6}+ 2.8}                                                                     & 51                                                                                            \\ \hline
\multicolumn{1}{|c|}{\textbf{Front Right}} & 20.3                                                                                           & \multicolumn{1}{c|}{\cellcolor[HTML]{D3FFFD}+ 10.5}                                                             & \multicolumn{3}{c|}{\cellcolor[HTML]{FFD7F6}+ 18.4}                                                                      & \multicolumn{3}{c|}{\cellcolor[HTML]{D2FFD6}+ 4}                                                                       & 53.9                                                                                          \\ \hline
\multicolumn{1}{|c|}{\textbf{Rear Left}}   & 21.6                                                                                           & \multicolumn{1}{c|}{\cellcolor[HTML]{D3FFFD}+ 7.3}                                                              & \multicolumn{3}{c|}{\cellcolor[HTML]{FFD7F6}+ 11.3}                                                                      & \multicolumn{3}{c|}{\cellcolor[HTML]{D2FFD6}+ 1.8}                                                                     & 45.2                                                                                          \\ \hline
\multicolumn{1}{|c|}{\textbf{Rear Right}}  & 18.8                                                                                           & \multicolumn{1}{c|}{\cellcolor[HTML]{D3FFFD}+ 8.4}                                                              & \multicolumn{3}{c|}{\cellcolor[HTML]{FFD7F6}+ 21}                                                                        & \multicolumn{3}{c|}{\cellcolor[HTML]{D2FFD6}+ 5.8}                                                                     & 57.8                                                                                          \\ \hline
\end{tabular}
\caption{Tires temperatures during the final at IMS. In the absence of tire warmers, and despite a strategy comprising warm-up manoeuvres and fast laps, the achieved temperatures after 6 laps are far from the nominal ones.}
\label{table:tires_temp}
\end{table}

\subsection{Overall Results in the Competition}
Using the presented software stack, the team achieved respectively the second and third position at the two main events of the competition. This section gives an overview of the overall results explaining the strategy we used and the failures which limited our final placements.
\FloatBarrier
\subsubsection{Indy Autonomous Challenge powered by Cisco}
In \Cref{table:ims_results} are summarized the results of the teams qualified for the Seminal and Final of the race at IMS and that they have been able to complete at least one of the rounds.

\begin{table}[]
\begin{tabular}{l|cc|cc|}
\cline{2-5}
                                                          & \multicolumn{2}{c|}{\textbf{Semifinal}}                                                                             & \multicolumn{2}{c|}{\textbf{Final}}                                                                                 \\ \cline{2-5} 
                                                          & \multicolumn{1}{c|}{\textbf{\begin{tabular}[c]{@{}c@{}}Average\\ Speed {[}m/s{]}\end{tabular}}} & \textbf{Position} & \multicolumn{1}{c|}{\textbf{\begin{tabular}[c]{@{}c@{}}Average\\ Speed {[}m/s{]}\end{tabular}}} & \textbf{Position} \\ \hline
\multicolumn{1}{|l|}{\textbf{TUM Autonomous Motorsport}}  & \multicolumn{1}{c|}{57.774}                                                                            & 2                 & \multicolumn{1}{c|}{\textbf{60.772}}                                                                           & 1                 \\ \hline
\multicolumn{1}{|l|}{\textbf{TII EuroRacing}}             & \multicolumn{1}{c|}{\textbf{58.628}}                                                                           & 1                 & \multicolumn{1}{c|}{51.83}                                                                           & 2                 \\ \hline
\multicolumn{1}{|l|}{\textbf{PoliMOVE}}                   & \multicolumn{1}{c|}{\textbf{55.634}}                                                                           & 3                 & \multicolumn{1}{c|}{DNF}                                                                           & 3                 \\ \hline
\multicolumn{1}{|l|}{\textbf{KAIST}}                      & \multicolumn{1}{c|}{37.71}                                                                           & 4                 & \multicolumn{1}{c|}{}                                                                           & DNQ               \\ \hline
\multicolumn{1}{|l|}{\textbf{Cavalier Autonomous Racing}} & \multicolumn{1}{c|}{53.592}                                                                           & DNF               & \multicolumn{1}{c|}{}                                                                           & DNQ               \\ \hline
\end{tabular}
\caption{Semi-final and Final results of teams qualified for the race at IMS. DNF stands for Did Not Finish, DNQ for Did Not Qualify. In bold, the best average speed of the top three teams is used to determine the final positions.}
\label{table:ims_results}
\end{table}

The order of the runs for the Semifinal has been set by draw. The PoliMove Autonomous Racing team was the first, followed by TUM, KAIST, Cavalier Autonomous Racing and finally us. Running after these teams gave us the advantage to know their results and setting a target speed high enough to conclude first without taking too many risks. 
Given the first position gained in the Semifinal, we started our run for the Final after TUM and PoliMove. The format consisted of four warm-up laps followed by two performance laps, in which the average speed was the criteria used to declare the winner, and one cool-down lap before coming back to the pit. 

\begin{figure}[htb!]
  \centering
  \includegraphics[
    width=0.70\textwidth
  ]{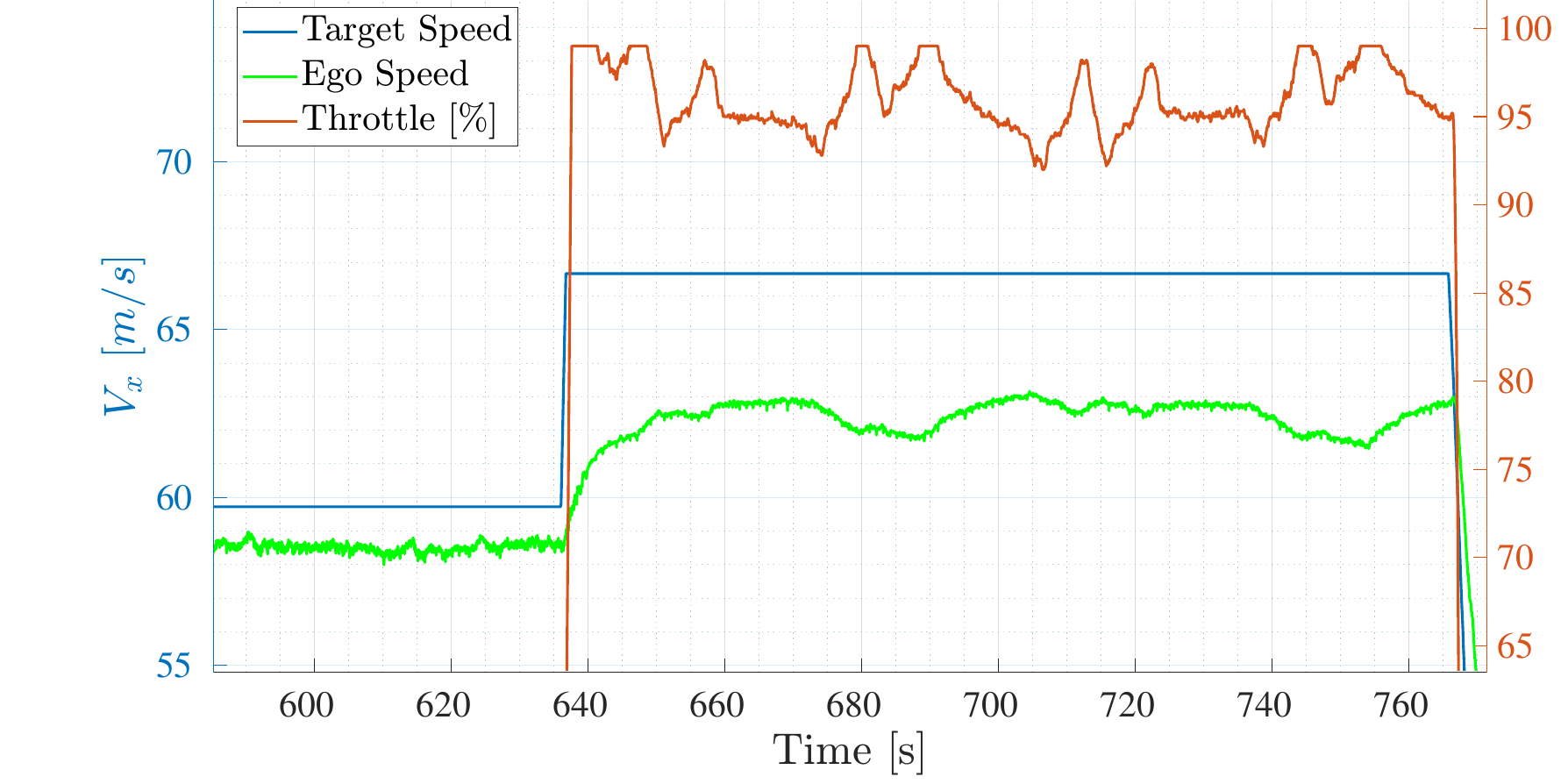}
  \caption{Fast laps in IMS Final. While the controller was requesting full throttle or a high value, the oscillations are due to a nonideal tuning of the turbocharger and a temporary malfunction of its mechanics.}
  \label{fig:ims_turbo_issue}
\end{figure}

Originally, our strategy, tested in simulation the night before the race, consisted in:
\begin{itemize}
    \item Four Warm-up laps: two laps using the Warm-up manoeuvre presented in \autoref{sec:control}, one lap at 200 km/h and the last one at 240 km/h.
    \item Two Performance laps at a speed equal to or higher than 240 km/h considering the other teams' result.
    \item One Cool-down lap at 150 km/h.
\end{itemize}
During the time available before our turn for the Final round, we considered performing the weaving manoeuvre to heat first the front tires at 95 km/h just on the first warm-up lap, followed by two laps at 215 km/h and a final warm-up lap at 240 km/h. This choice has been applied and tested in our SiL simulator a few minutes before the run without the attention properly needed to guarantee its correctness. This resulted in what we internally called the ``Million dollar bug'', considering the first place prize. 
Taking into account the TUM's result of an average speed of 218.8 km/h and a hardware failure that led PoliMove to keep their average speed of 200.03 km/h gained during the Semifinal, we decided to keep the target speed for the performance laps at a speed of 240 km/h around all the track. 
As can be seen in \figref{fig:ims_turbo_issue}, despite a throttle pedal command higher than 90\%, the speed didn't increase more than 227.4 km/h due to a momentary malfunction on the turbo-charger. 
After completing the first performance lap, the car reduced its speed to 150 km/h, as depicted in \figref{fig:tires_final_ims}, which was the target for the cool-down lap. During the last-minute change in the code, we erroneously set the cool-down lap speed at the end of the fifth lap instead of the sixth. Concluding the last lap at the same speed as the previous one would have guaranteed an average speed high enough to win but the error led us to an average speed lower than the one achieved in the Semifinal positioning us second.

\subsubsection{Autonomous Challenge @ CES}

At the second event, the order was set by draw as well and we ended up starting first at the time trial part in which the laptime was the criteria for determining the seeding in the bracket of the Passing Competition explained in \autoref{sec:introduction}. The run consisted of a maximum of ten laps where the teams were not constrained to follow any kind of format. We decided to perform the outlap and the first lap doing the weaving manoeuvre at 100 km/h, followed by a series of laps at incrementally higher speed. During the seventh lap, we set a target speed of 77.7 m/s (280 km/h) but the vehicle reached a maximum of 75.5 m/s (272 km/h), despite a full throttle command as presented in \cite{raji}, achieving a lap time of 33.99 seconds.
Once the racecar came back to the pitlane, the mechanics found that a cable related to the powertrain has been detached and caused the speed limitation.
The results of the time trial are summarized in \autoref{table:lvms_timetrial_results}.

\begin{table}[]
\centering
\begin{tabular}{l|cc|}
\cline{2-3}
                                                          & \multicolumn{2}{c|}{\textbf{Qualifying}} \\ \cline{2-3} 
                                                          & \multicolumn{1}{c|}{\textbf{\begin{tabular}[c]{@{}c@{}}Fastest Lap Time {[}s{]}\end{tabular}}} & \textbf{Position} \\ \hline
\multicolumn{1}{|l|}{\textbf{TII EuroRacing}}  & \multicolumn{1}{c|}{33.99}                                                                            & 2 \\ \hline
\multicolumn{1}{|l|}{\textbf{TUM Autonomous Motorsport}}             & \multicolumn{1}{c|}{35.3}                                                                           & 3 \\ \hline
\multicolumn{1}{|l|}{\textbf{PoliMOVE}}                   & \multicolumn{1}{c|}{32.54}                                                                           & 1 \\ \hline
\multicolumn{1}{|l|}{\textbf{Cavalier Autonomous Racing}} & \multicolumn{1}{c|}{}                                                                           & DNF \\ \hline
\end{tabular}
\caption{Qualifying results of the race at LVMS; the 4 fastest teams in testing were admitted.}
\label{table:lvms_timetrial_results}
\end{table}

In the Passing Competition, we faced TUM for the Semifinal of the event. We've been able to pass the rounds up to the defending speed level of 58 m/s overtaking at 63 m/s (226.8 km/h).
Table\;\ref{table:result:planner_speeds} compares the overtaking speed with the other teams.

\begin{figure}[htb!]
  \centering
  
  \begin{subfigure}[b]{0.49\textwidth}
    \includegraphics[
        width=0.99\textwidth,
    ]{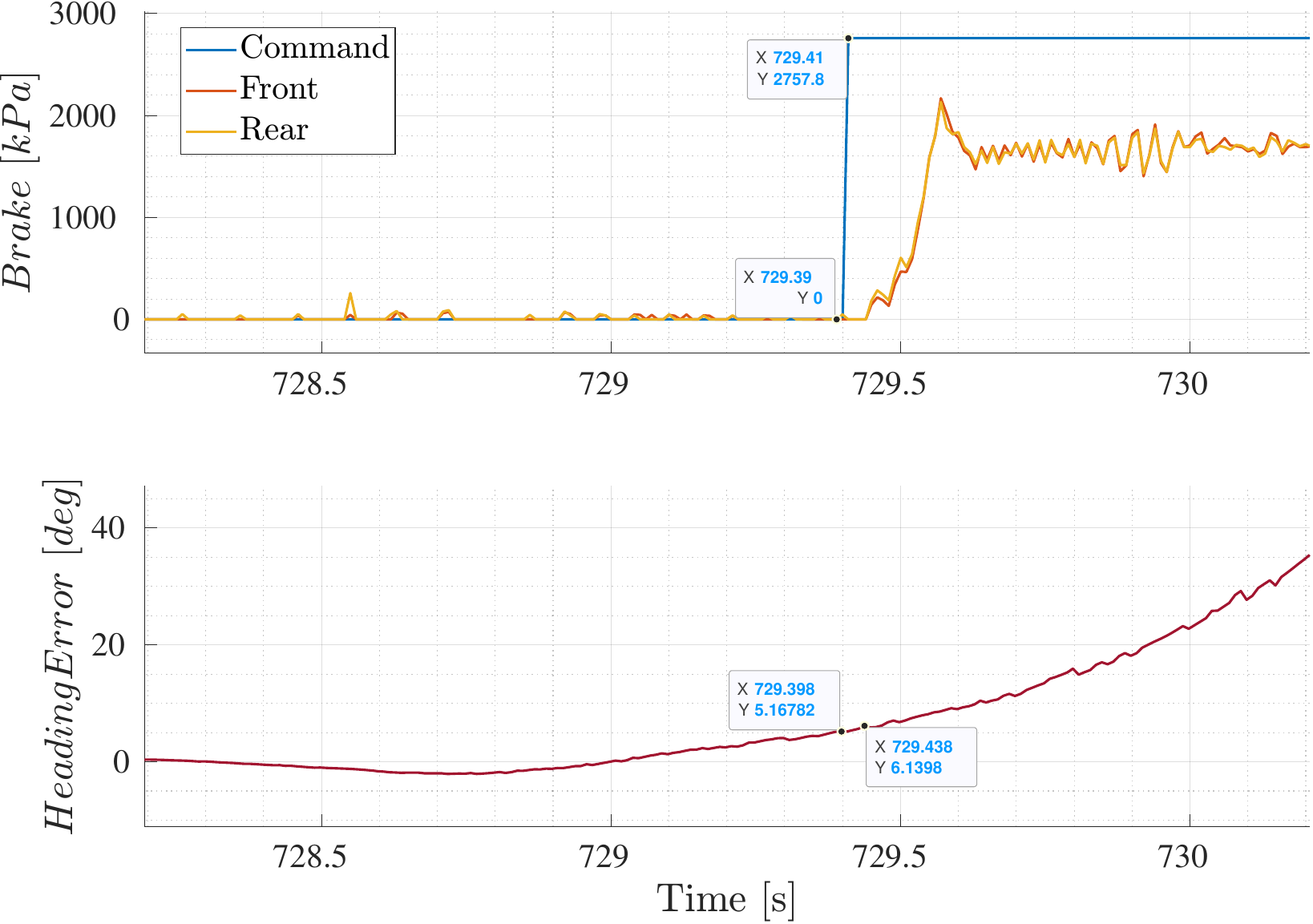}
    \caption{Heading Error triggering brake command.}
  \end{subfigure}
  \begin{subfigure}[b]{0.49\textwidth}
    \includegraphics[
        width=0.99\textwidth,
    ]{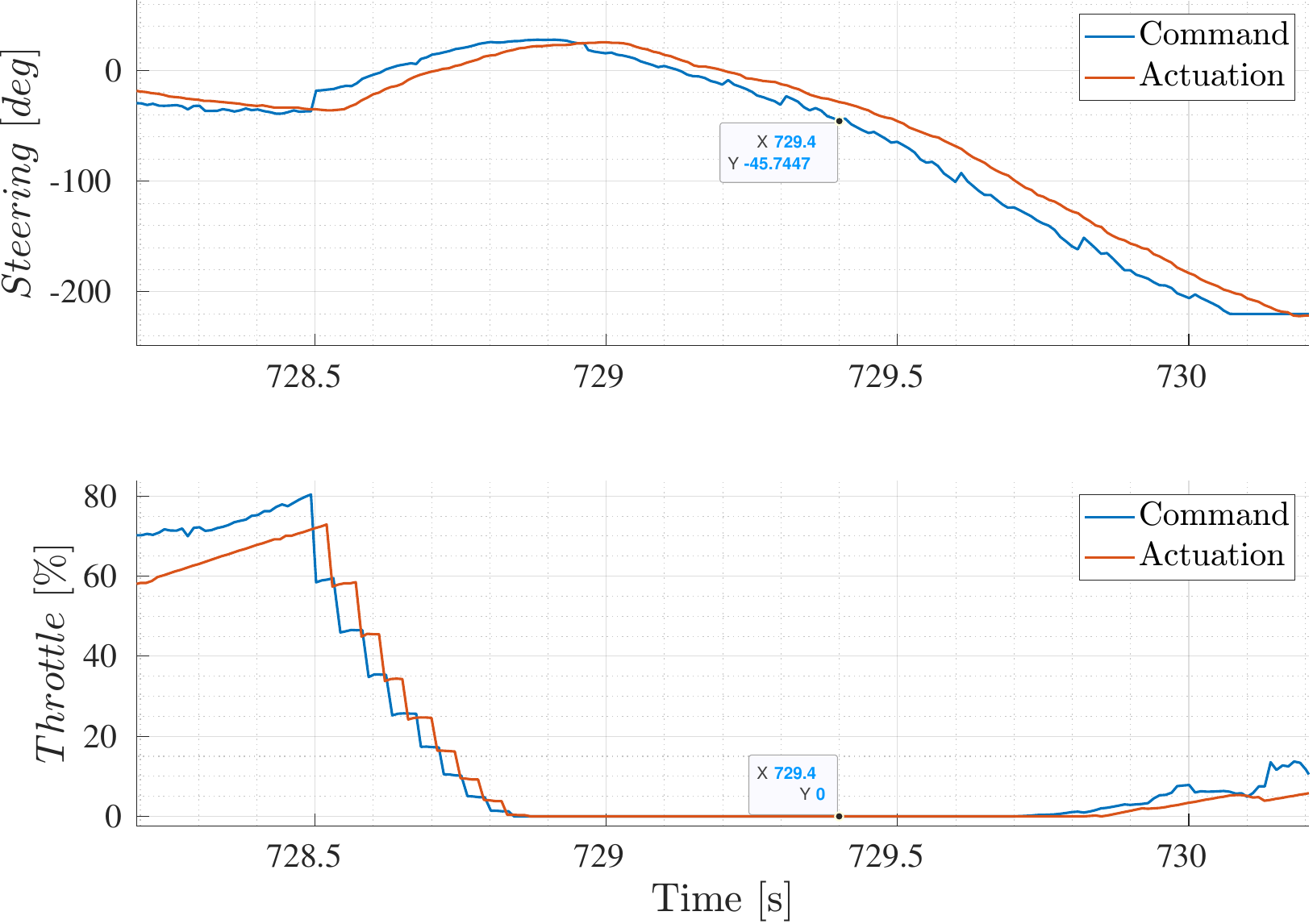}
    \caption{MPC and Raptor actuation signals.}
  \end{subfigure}

  \caption{LVMS semifinal crash analysis. A false detection of a ghost car in front of the car triggered a steering correction and reduction in throttle request (b). Shortly after, the heading error w.r.t. the planned path went over the safety threshold, and the supervisor triggered an emergency braking (a). At over $200$kph, this resulted in the loss of control of the vehicle.}
  \label{fig:lvms_crash}
\end{figure}

An edge case for the motion planning and control modules happened during the round at the defense speed of 60 m/s. A false detection of a standing obstacle by the Radar lets the planner generate a series of aggressive manoeuvres with a higher difference in terms of curvature and smoothness from each other. The MPC reacted with a much higher heading error than the one usually had during other tests on track at higher speeds, triggering a hard brake by the safety checks explained in \autoref{sec:sw_stack}. In \figref{fig:lvms_crash} can be seen that as soon as the heading error passed the value of 6deg, the brakes pressure started to increase bringing the car to a spin and hitting the track boundaries despite the controller was trying to steer to the right.
\FloatBarrier
\begin{table}[hbt!]
  \centering
  \begin{tabular}[hbt!]{|>{\centering\arraybackslash}p{2.5cm}|>{\centering\arraybackslash}p{2.2cm}|>{\centering\arraybackslash}p{2.2cm}|>{\centering\arraybackslash}p{2.2cm}|>{\centering\arraybackslash}p{2.2cm}|}
    \hline
    \textbf{Defence Speed (m/s)} & \textbf{TII EuroRacing} & \textbf{TUM} & \textbf{KAIST} & \textbf{PoliMOVE} \\ \hline
    36 m/s & $\checkmark$ & $\checkmark$ & $\checkmark$ & $\checkmark$ \\ \hline
    45 m/s & $\checkmark$ & $\checkmark$ & $\checkmark$ & $\checkmark$ \\ \hline
    51 m/s & $\checkmark$ & $\checkmark$ & $\checkmark$ & $\checkmark$ \\ \hline
    56 m/s & $\checkmark$ & $\checkmark$ & $\times$ & $\checkmark$ \\ \hline
    58 m/s & $\checkmark$ & $\checkmark$ & & $\checkmark$ \\ \hline
    60 m/s & $\times$ & $\checkmark$ & & $\checkmark$ \\ \hline
    62 m/s & & $\checkmark$ &  & $\checkmark$ \\ \hline
    65 m/s & & $\checkmark$ & & $\checkmark$ \\ \hline
    67 m/s & & $\times$ & & $\checkmark$ \\ \hline
  \end{tabular}
  \caption{Comparison of successful overtakes between all the qualified teams, during the Las Vegas Motor Speedway event. The defense speed is the speed of the defending car during the overtake.}
  \label{table:result:planner_speeds}
\end{table}

\vspace{-0.1cm}
\section{Lessons Learned}
\label{sec:lessons}
Robotics challenges are great instruments for pushing the integration of the latest research results in real-world applications and steer the efforts of research teams towards new results that can find application outside the lab. On top of this, the opportunity of competing against other teams is an important mean to build and maintain a research community active and exchange ideas at a fast rate.

We could easily argue that everyone who competes in a race wants to win. When this result is not achieved, it is very important to go back and make a rational analysis of the result. In our case, we summarize our analysis in the following points, hoping they could provide useful insights to the community.

\begin{itemize}
    \item Control solutions based on complex models of the vehicle bring difficulties related to the different conditions of the real world with respect to the estimated model. This mismatch could lead the researchers to move to different approaches based on Robust Control where, rather than modelling the non-linear dynamics of the system, the focus is on considering the uncertainty around a simpler model, or on bounding the control commands on a set of potential limits of the longitudinal and lateral accelerations. In our work, we demonstrated that a model identified from simulation has been used in an MPC in two different tracks with different weather conditions and different scenarios, with similar performances adapting only the weighting terms of the optimization problem and the cost function. Although not having updated the model after the validation on track due mainly to the limited testing time, we believe that the experience gathered on modelling, validating, and tuning the controller, will help on speeding up the process and being able to bring in time the needed refinements on the parameters for each new track and road condition. Further effort should be put into being able to automatically learn and refine the model parameters considering at runtime the stability and the tracking performance of the controller.
    \item Given the technical challenges that GNSS systems pose in practice \cite{massa}, we believe that investing effort in GNSS-denied localization systems will be key in future racing championships. We believe that to produce a step-change in autonomous technology, a racecar should be equipped with all its capabilities in the edge-vehicle, relying on external infrastructure only for interactions with race control and live telemetry streaming. A first step in this direction will be in reducing the importance of GNSS modules in ego-vehicle localization, up to the point where these could become non-necessary.
    \item The crash that happened in the head-to-head race has been caused technically by setting the threshold on the safety check related to the heading error without considering possible extreme scenarios, for the perception and planning modules, which are difficult to face in simulation. 6deg as the threshold on the heading error can be considered a strict value since in simulation the controller has been able to correctly react in similar conditions when the check was not enabled. Furthermore, it would have been trivial to understand the potential effects of a hard brake command at high speeds with tires below their nominal temperature. For this reason, the safety check would be reevaluated for future runs. Potentially, the hard brake would be limited to low speeds and to the occasions where the control commands are not properly applied to the vehicle.
    \item Differently from usual research, in our case the race and the competitive component of the challenge brought us to take important decisions in a stressful environment and in a short time. Besides technical errors, wrong or high-risky organizational and operational decisions could ruin the final result as well. In our experience, this has been proved in the case of the Million dollar bug where it has been neglected the high chance of a human error in applying and testing a last-minute change on the software.
\end{itemize}

\section{Conclusions}
\label{sec:conclusion}

In this work, we presented the complete software stack implemented by the TII EuroRacing team. Each module has been described including technical results as well as the overall achievements. Insights on the aspects we considered crucial for reaching speeds above 75 m/s (270 km/h), avoiding static obstacles, and racing in a head-to-head scenario, have been given. We explained the failures that brought us to not achieve the first position in the final events, with important considerations that could be beneficial to the other teams and researchers competing in challenges of different robots and fields.

With new Autonomous Racing challenges planned for the next years, the team is working on \texttt{er.autopilot 2.0}. The updated software should improve the robustness of the sensors fusion on the detection module to cope with potential false detection and disturbances on the racetrack. Despite the implementation of a light warm-up manoeuvre to increase the temperature of the front tires, we will consider more aggressive manoeuvres to be performed in closed-loop control to speed up the warm-up procedure and reach the ideal temperature. A key point to achieve this goal is to improve the path-tracking performance of the MPC. Thanks to the data gathered at high speeds, it has been possible to confirm the correctness of an updated version of the multibody model developed in Dymola, which has been used to identify the single-track model parameters for the controller.

Another important feature will be the capability of running on GNSS-denied solutions. In addition to the LiDAR-based localization, we will consider the integration of a pure local control method. In \cite{lee}, the authors presented a resilient navigation method based on following a distance from the wall of the track, using the LiDAR sensor and a variant of the RANSAC algorithm \cite{fischler}. Similarly, we will implement LiDAR-based and camera-based navigation for emergency situations.

Future applications will be on running the algorithms in free racing scenarios where more than two vehicles are allowed to drive without strict limitations on the possible trajectories to follow. For this purpose, it will be important the development of a Local Planner capable of generating aggressive, but feasible and safe, paths and velocity profiles. This would be possible by considering a dynamic model on the planning module and improving the integration between Planning and Control.


\subsubsection*{Acknowledgments}
We would like to thank all the students and researchers who contributed in small part to the development. In particular, André Jesus, Abdurrahman İşbitirici, Mankaran Singh, Andrea Serafini, Francesco Moretti, Maciej Dziubiński, and Vallabh Ansingkar.
Thanks to Claytex for the VeSyMA Motorsports libraries and MegaRide{\footnote{\href{https://www.megaride.eu/}{https://www.megaride.eu/}}} for the support in the tire model identification.
Thanks to SpinItalia{\footnote{\href{https://www.spinitalia.com/}{https://www.spinitalia.com/}}}, particularly Francesco La Gala, for the insight on the GNSS failure at LVMS.

We would also like to thank the IAC organization and all their partners for making this work possible.

Lastly, we would like to acknowledge the collaborative work done by all the teams during the first months at LOR and IMS. In particular, the work carried on by Alexander Wischnewski from TUM Autonomous Motorsport and Will Bryan from Autonomous Tiger Racing.

\bibliographystyle{apalike}
\bibliography{refs}

\end{document}